%% file: video_based_recons_arxiv.tex
\documentclass[10pt,twocolumn,letterpaper]{article}

\usepackage{iccv}
\usepackage{times}
\usepackage{epsfig}
\usepackage{graphicx}
\usepackage{amsmath,amsfonts,amssymb}
\usepackage{mathtools}
\usepackage{comment}

\usepackage{array}
\usepackage{multirow}
\usepackage{booktabs}
\usepackage{makecell}
\makeatletter
\@namedef{ver@everyshi.sty}{}
\makeatother
\usepackage{tikz}
\usepackage{subcaption}
\usepackage{caption}
\usepackage{xspace}
\usepackage{overpic}
\usepackage{wrapfig}
\usepackage[outline]{contour}
\contourlength{0.075em}
\usepackage[export]{adjustbox}

\usepackage[pagebackref=true,breaklinks=true,letterpaper=true,colorlinks,bookmarks=false]{hyperref}

\usepackage[capitalize]{cleveref}
\crefname{section}{Sec.}{Secs.}
\Crefname{section}{Section}{Sections}
\Crefname{table}{Table}{Tables}
\crefname{table}{Tab.}{Tabs.}

\input{macros}

\usepackage{enumitem}
\setlist[itemize]{parsep=0pt,partopsep=0pt,leftmargin=*,itemsep=5pt}
\setlist[enumerate]{parsep=0pt,partopsep=0pt,leftmargin=*,itemsep=5pt}
\makeatletter
\renewcommand\paragraph{\@startsection{paragraph}{4}{\z@}%
    {.5em}%
    {-1em}%
    {\normalfont\normalsize\bfseries}}
\makeatother
\iccvfinalcopy 

\def\iccvPaperID{} 

\begin{document}

\title{
    Neural-PBIR Reconstruction of Shape, Material, and Illumination
}

\author{
Cheng Sun$^{1,2\ast}$ \qquad
Guangyan Cai$^{1,3\ast}$ \qquad
Zhengqin Li$^{1}$ \qquad
Kai Yan$^{3}$ \qquad
Cheng Zhang$^{1}$ \\
Carl Marshall$^{1}$ \qquad
Jia-Bin Huang$^{1,4}$ \qquad
Shuang Zhao$^{3}$ \qquad
Zhao Dong$^{1}$ \\
{\small $^{1}$Meta RLR} \quad
{\small $^{2}$National Tsing Hua University} \quad
{\small $^{3}$University of California, Irvine} \quad
{\small $^{4}$University of Maryland, College Park}
\\{\small $^{\ast}$The authors contribute equally to this paper}
}

\input{teaser}


\ificcvfinal\thispagestyle{empty}\fi

\input{abstract}
\maketitle
\input{intro}

\input{related}

\input{ours}

\input{eval}
\input{conclusion}
\input{acknowledgement}

\setcounter{section}{0}
\renewcommand{\thesection}{\Alph{section}}
\input{stanford_orb}
\input{supp/sec_technical_detail}

\input{supp/sec_additional_results}
{\small
\newpage
\twocolumn[
\begin{center}
\addcontentsline{toc}{section}{References}
\end{center}
]
\bibliographystyle{ieee_fullname}
\bibliography{video_based_recons_arxiv}
}
\end{document}

%% file: macros.tex
\definecolor{orange_cubic}{rgb}{.9765, .5887, .3569}
\definecolor{purple_cubic}{rgb}{.4706, 0, .5216}
\definecolor{green_cubic}{rgb}{.28603, .81178, .5008}

\definecolor{grayLL}{rgb}{.98, .98, .98}
\definecolor{grayL}{rgb}{.9, .9, .9}
\definecolor{purpleL}{rgb}{.9735, .95, .9761}
\definecolor{purpleD}{rgb}{.8941, .8, .9043}
\definecolor{greenL}{rgb}{.9643, .9906, .9750}
\definecolor{greenD}{rgb}{.7145, .9249, .7999}
\definecolor{greenDD}{rgb}{.3145, .6249, .3999}
\definecolor{orangeLL}{rgb}{0.9991, 0.9846, 0.9759}
\definecolor{orangeL}{rgb}{.9982, .9692, .9518}
\definecolor{orangeD}{rgb}{.9929, .8766, .8071}

\definecolor{redL}{rgb}{1.0, 0.95, 0.95}
\definecolor{redD}{rgb}{1.0, 0.8, 0.8}
\definecolor{redDD}{rgb}{1.0, 0.4, 0.4}
\definecolor{yellowL}{rgb}{1.0, 1.0, 0.95}
\definecolor{yellowD}{rgb}{0.95, 0.95, 0.6}
\definecolor{yellowDD}{rgb}{0.8, 0.8, 0.2}
\definecolor{blueLL}{rgb}{0.98, 0.98, 1.0}
\definecolor{blueL}{rgb}{0.95, 0.95, 1.0}
\definecolor{blueD}{rgb}{0.8, 0.8, 1.0}
\definecolor{blueDD}{rgb}{0.6, 0.6, 1.0}

\definecolor{uciBlue}{RGB}{0,100,164}
\definecolor{uciBlueL}{RGB}{127.5, 177.5, 209.5}
\definecolor{uciOrange}{RGB}{247,141,45}
\definecolor{uciOrangeL}{RGB}{251,198,150}

\newcommand{\defeq}{\vcentcolon=}

\makeatletter
\newcommand{\dotr}[1]{%
	\mathpalette\@dotr{#1}%
}
\newcommand*{\@dotr}[2]{%
	\sbox0{$\m@th#1#2$}%
	\usebox{0}%
	\raisebox{\dimexpr\ht0-\height}{$\m@th#1\@smallbullet#1\bullet$}%
	\kern\scriptspace
}
\newcommand*{\@smallbullet}[2]{%
	\scalebox{.4}{$\m@th#1#2$}%
}
\makeatother

\newcommand{\bx}{\boldsymbol{x}}

\newcommand{\bn}{\boldsymbol{n}}

\newcommand{\bo}{\boldsymbol{o}}

\newcommand{\bv}{\boldsymbol{v}}

\newcommand{\bom}{\boldsymbol{\omega}}
\newcommand{\bomi}{\bom_\mathrm{i}}
\newcommand{\bomo}{\bom_\mathrm{o}}

\newcommand{\calL}{\mathcal{L}}
\newcommand{\brdf}{f}

\newcommand{\Real}{\mathbb{R}}
\newcommand{\bbS}{\mathbb{S}}
\newcommand{\Sph}{\bbS^2}

\newcommand{\sdf}{S}
\newcommand{\Lo}{L_\mathrm{o}}
\newcommand{\Li}{L_\mathrm{i}}
\newcommand{\Lenv}{L_\mathrm{env}}
\newcommand{\LenvSG}{\Lenv^\mathrm{SG}}
\newcommand{\Lostudent}{\hat{L}_\mathrm{o}}

\newcommand{\V}{V}
\newcommand{\Vsdf}{\V^\mathrm{(sdf)}}
\newcommand{\Vfeat}{\V^\mathrm{(feat)}}

\newcommand{\Vis}{\mathrm{Vis}}
\newcommand{\IndIl}{\Li^\mathrm{(ind)}}
\newcommand{\wiset}{\Omega}
\newcommand{\vind}{v}

\newcommand{\R}{\mathcal{R}}
\newcommand{\Rmask}{\R^\mathrm{mask}}
\newcommand{\img}{\mathcal{I}}

\newcommand{\loss}{\calL}
\newcommand{\Llap}{\loss_\mathrm{lap}}
\newcommand{\Lphoto}{\loss_\mathrm{photo}}
\newcommand{\Lrgb}{\loss_\mathrm{pp\_rgb}}
\newcommand{\Lsurf}{\loss_\mathrm{surf}}
\newcommand{\Lir}{\loss_\mathrm{IR}}
\newcommand{\Limg}{\loss_\mathrm{img}}
\newcommand{\Lmask}{\loss_\mathrm{mask}}
\newcommand{\Lreg}{\loss_\mathrm{reg}}
\newcommand{\Wmask}{w_\mathrm{mask}}
\newcommand{\Wreg}{w_\mathrm{reg}}
\newcommand{\Ldt}{\loss_\mathrm{distill}}
\newcommand{\Ldtreg}{\loss_\mathrm{v\_reg}}
\newcommand{\Ldtbg}{\loss_\mathrm{bg}}
\newcommand{\Ldttotal}{\loss_\mathrm{distill\_total}}

\newcommand{\weight}{w}
\newcommand{\Wlap}{\weight_\mathrm{lap}}
\newcommand{\Wrgb}{\weight_\mathrm{pp\_rgb}}
\newcommand{\Wdtreg}{\weight_\mathrm{v\_reg}}
\newcommand{\Wdtbg}{\weight_\mathrm{bg}}

\newcommand{\M}{M}
\newcommand{\Ma}{\M_\mathrm{a}}
\newcommand{\Mr}{\M_\mathrm{r}}

\newcommand{\Mn}{\M_\mathrm{n}}

\newcommand{\T}{T}
\newcommand{\Ta}{\T_\mathrm{a}}
\newcommand{\Tr}{\T_\mathrm{r}}

\newcommand{\neus}{\textsf{NeuS}\xspace}
\newcommand{\mii}{\textsf{MII}\xspace}
\newcommand{\nvdiff}{\textsf{nvdiffrec-mc}\xspace}

\newlength{\resLen}

\newcommand{\vtext}[1]{\rotatebox{90}{#1}}

\definecolor{rebuttal_purple}{rgb}{.643, .349, .820}
\definecolor{rebuttal_red}{rgb}{.949, .400, .670}
\definecolor{rebuttal_blue}{rgb}{.172, .827, .882}

%% file: teaser.tex
\twocolumn[{%
    \maketitle
    \renewcommand\twocolumn[1][]{#1}%
    \vspace{-7mm} 
    \centering
    \setlength{\resLen}{.13\linewidth}
    \addtolength{\tabcolsep}{-4pt}
    \small
    \begin{tabular}{ccccccc}
        \includegraphics[width=\resLen]{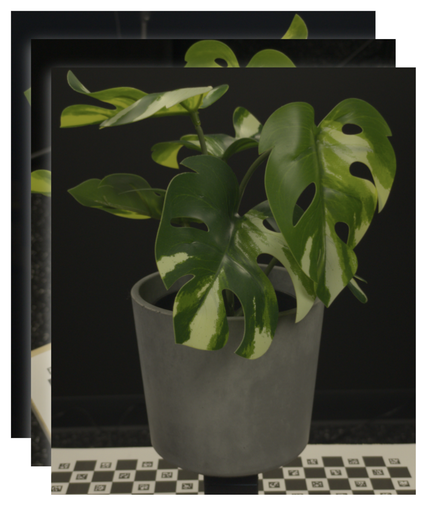} &
        \includegraphics[width=\resLen]{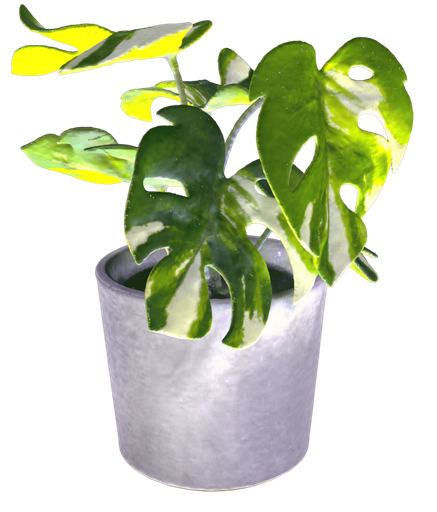} &
        \includegraphics[width=\resLen]{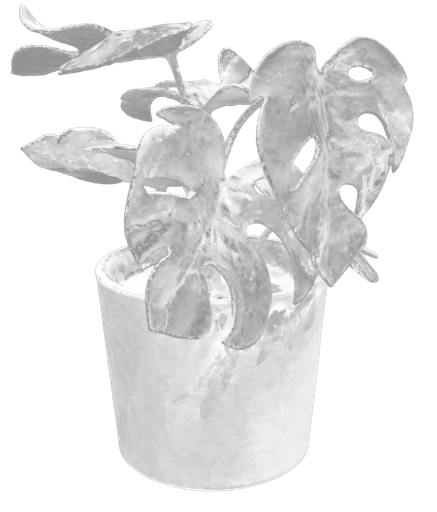} &
        \includegraphics[width=\resLen]{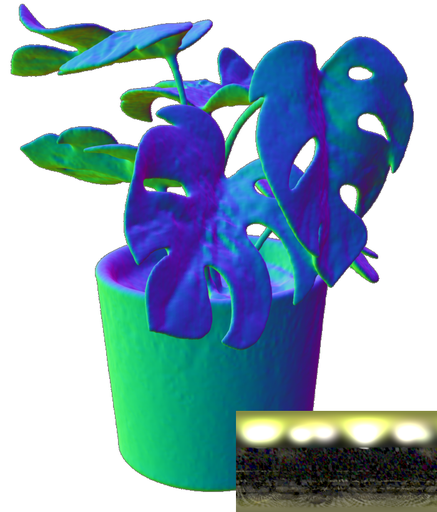} &
        \includegraphics[width=\resLen]{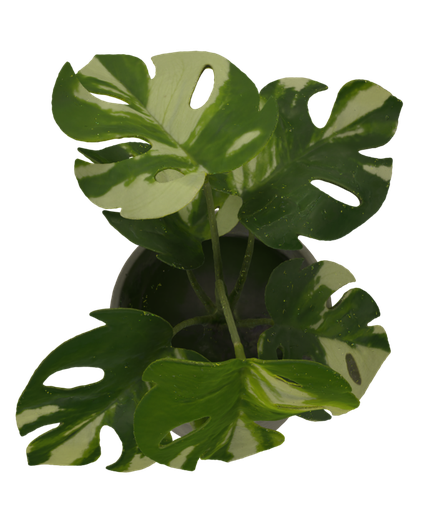} &
        \includegraphics[width=\resLen]{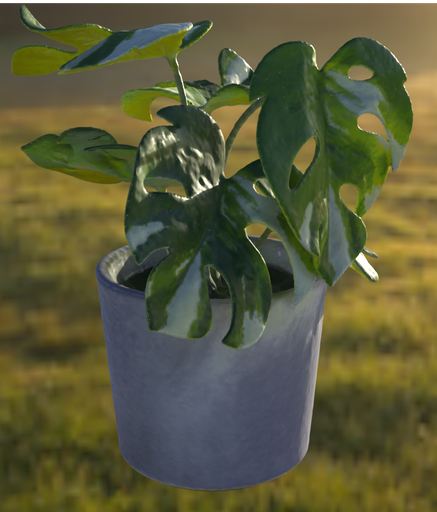} &
        \includegraphics[width=\resLen]{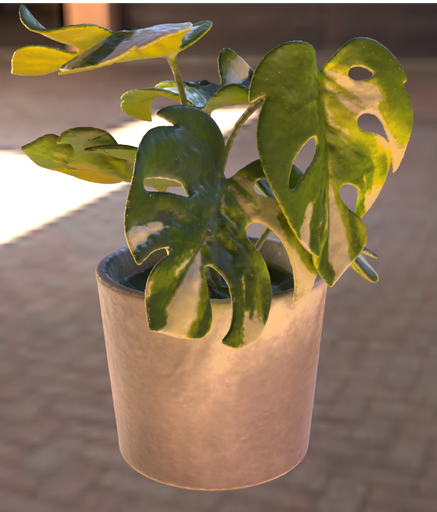}
        \\
        (a) Input images & (b1) Albedo & (b2) Roughness & (b3) Shape \& lighting & (c1) Novel view & (c2) Relit 1 & (c3) Relit 2
    \end{tabular}
    \vspace{-.5em}
    \captionof{figure}{
       \textbf{Neural-PBIR} recovers high-fidelity material (b1,2), shape and lighting (b3), enabling realistic re-rendering (c1-3).
       \label{fig:teaser}
    }
    \vspace{7mm}
}]

%% file: abstract.tex
\begin{abstract}
    Reconstructing the shape and spatially varying surface appearances of a physical-world object as well as its surrounding illumination based on 2D images (e.g., photographs) of the object has been a long-standing problem in computer vision and graphics. In this paper, we introduce an accurate and highly efficient object reconstruction pipeline combining \emph{neural based object reconstruction} and \emph{physics-based inverse rendering (PBIR)}. Our pipeline firstly leverages a neural SDF based shape reconstruction to produce high-quality but potentially imperfect object shape. Then, we introduce a neural material and lighting \emph{distillation} stage to achieve high-quality predictions for material and illumination. In the last stage, initialized by the neural predictions, we perform PBIR to refine the initial results and obtain the final high-quality reconstruction of object shape, material, and illumination. Experimental results demonstrate our pipeline significantly outperforms existing methods quality-wise and performance-wise.
    Code: \url{https://neural-pbir.github.io/}
\end{abstract}

%% file: intro.tex
\section{Introduction}
\label{sec:intro}
Reconstructing geometry, material reflectance, and lighting from images, also known as inverse rendering, is a long-standing challenge in computer vision and graphics. 
Conventionally, the acquisition of the three intrinsic components has been mainly studied independently. 
For instance, multiview-stereo (MVS) \cite{schoenberger2016sfm,schoenberger2016mvs,furukawa2009accurate} and time-of-flight \cite{zhu2010reliability} methods only focus on recovering object geometry, usually based on diffuse reflectance assumption.
Classical material acquisition methods typically assume known or simple geometries (e.g., a planar surface) with highly controlled illuminations \cite{nielsen2015optimal,xu2016minimal,zhou2016sparse}, usually created with a light stage or gantry. This significantly limits their practicality when such capturing conditions are unavailable.

Recently, the advent of novel techniques enables us to jointly reconstruct shape, material, and lighting from 2D images of an object. 
At a high level, these techniques can be classified into two categories.
Neural reconstruction methods encode the appearance of objects into a multi-layer perceptron (MLP) and optimize the network by minimizing the rendering errors from different views through differentiable volume ray tracing. NeRF \cite{MildenhallSTBRN20} reconstructs a density field-based radiance field that allows high-quality view synthesis but not relighting. A series of methods \cite{WangLLTKW21,YarivGKL21,OechsleP021} compute the density field from the signed distance function to achieve high-quality geometry reconstruction. Recent works \cite{ZhangLWBS21,ZhangSDDFB21,BossBJBLL21,BossJBLBL21,MunkbergCHES0GF22,HasselgrenHM22,ZhangSHFJZ22} seek to fully decompose shapes, materials, and lighting from input images. However, due to the high computational cost of volume ray tracing and neural rendering, those methods take hours \cite{MunkbergCHES0GF22,HasselgrenHM22} or a day \cite{BossBJBLL21,ZhangSHFJZ22} to run and usually cannot model more complex indirect illumination \cite{ZhangLWBS21,ZhangSDDFB21,BossBJBLL21,BossJBLBL21,MunkbergCHES0GF22,HasselgrenHM22}, causing shadows and color bleeding to be baked into the material reflectance. Several new methods \cite{SunSC22,mueller2022instant,ChenXGYS22} significantly reduce the computational cost of radiance field reconstruction using hybrid volume representations and efficient MLPs. We are among the first that adopt these novel techniques for efficient joint recovery of geometry, material, and lighting.

\input{tabs/ours_pipeline}

On the contrary, \emph{physics-based inverse rendering} (PBIR) \cite{cai2022physics,luan2021unified,nimier2021material,azinovic2019inverse} optimizes shape, material, and lighting by computing unbiased gradients of image appearance with respect to scene parameters. Leveraging physics-based differentiable renderers \cite{Zhang:2020:PSDR,Jakob2022DrJit,li2018differentiable}, state-of-the-art PBIR pipelines can efficiently handle complex light transport effects such as soft shadow and interreflection. Such complex light transport effects cannot be easily handled through volume-based neural rendering. On the other hand, since PBIR methods rely on gradient-based optimization to refine intrinsic components, they can be prone to local minima and overfitting. Therefore, they may require a good initialization to achieve optimal reconstruction quality.

In this paper, we present a highly efficient and accurate inverse rendering pipeline with the advantages of both neural reconstruction and PBIR methods. Our pipeline attempt to estimate geometry, spatially varying material reflectance, and an HDR environment map from multiple images of an object captured under static but arbitrary lighting.  As shown in Fig. \ref{fig:pipeline}, our pipeline consists of three stages. In the first stage, we propose a hybrid neural volume-based method for fast neural SDF and radiance field reconstruction, which achieves state-of-the-art geometry accuracy and efficiency. In the next stage, based on the reconstructed geometry and radiance field, we design an efficient optimization method to distill materials and lighting by fitting the surface light field. Our method relies on a radiance field to handle visibility and indirect illumination but avoids expensive volume ray tracing to significant computational cost compared to some recent works. Finally, we use an advanced PBIR framework \cite{Zhang:2020:PSDR,Jakob2022DrJit} to jointly refine the geometry, materials, and lighting. Note that our PBIR framework models complex light transport effects such as visibility, occlusion, soft shadows and interreflection in a physically correct and unbiased way while still being much faster than recent inverse rendering methods. 

Concretely, our contributions include the following: 
\begin{itemize}
\vspace{-0.2cm}
    \item A hybrid volume representation for fast and accurate geometry reconstruction. 
\vspace{-0.2cm}
    \item A efficient optimization scheme to distill high-quality initial material and lighting estimation from the reconstructed geometry and radiance field. 
\vspace{-0.2cm}
    \item An advanced PBIR framework that jointly optimizes materials, lighting and geometry with visibility and interreflection handled in a physically unbiased way.
\vspace{-0.2cm}
    \item A end-to-end pipeline that achieves state-of-the-art geometry, material and lighting estimation that enables realistic view synthesis and relighting.
\end{itemize}

%% file: tabs/ours_pipeline.tex
\begin{figure*}[t]
\centering
\includegraphics[trim=0px 200px 0px 20px, clip, width=0.98\textwidth]{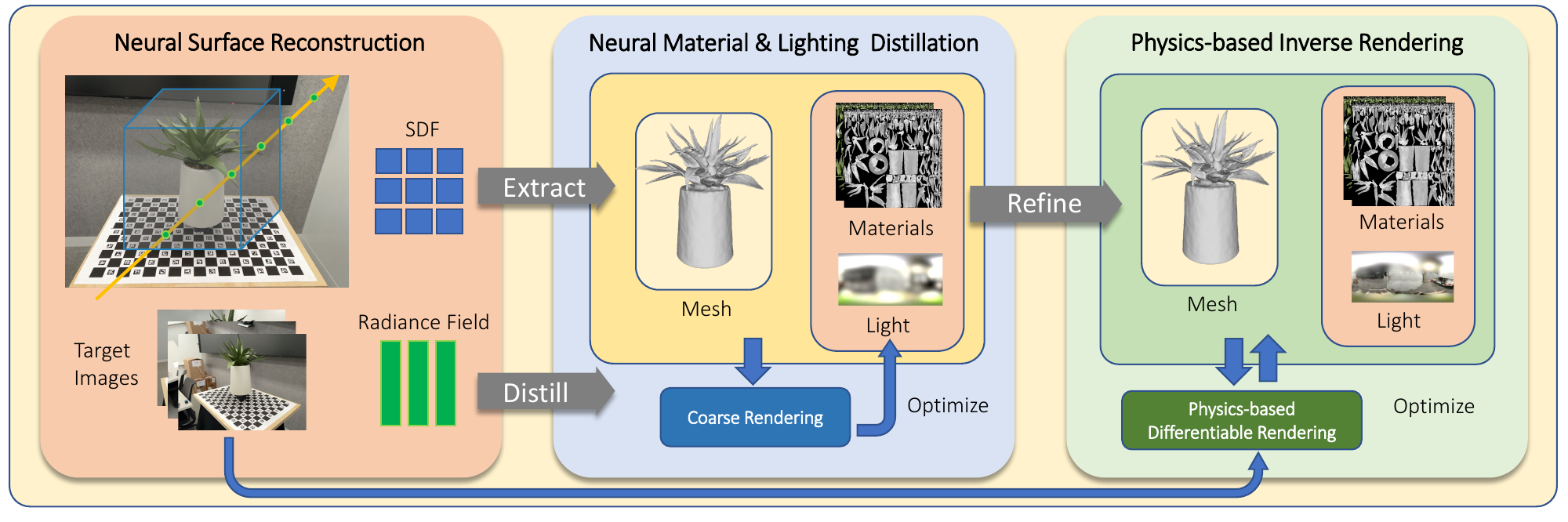}
\vspace{-0.3cm}
\caption{\textbf{Our pipeline for joint shape, material, and lighting estimation.}}
\label{fig:pipeline}
\end{figure*}

%% file: related.tex
\section{Related Works}
\label{sec:related}

\paragraph{Volumetric surface reconstruction.}
Recent progress in volumetric-based surface reconstruction of a static scene shows high-quality and robust results.
NeuS~\cite{WangLLTKW21} and VolSDF~\cite{YarivGKL21} replace NeRF~\cite{MildenhallSTBRN20}'s density prediction with signed distance values and proposes an unbiased and occlusion-aware volume rendering, achieving promising results.
Subsequent works improve quality by introducing regularization losses to encourage surface smoothness~\cite{Zhang0LFMTQ22}, multiview consistency~\cite{FuXOT22}, and manhattan alignment~\cite{GuoPLWZBZ22}.
These methods use a large MLP as their volumetric representation, which is however slow (many hours) to optimize per scene.
Recent advances~\cite{SunSC22,YuFTCR22,mueller2022instant,ChenXGYS22} show great optimization acceleration without loss of quality by using an explicit grid.
Unfortunately, using explicit volume to model SDF leads to bumpy and noisy surfaces~\cite {YuPNSG22}.
Supervisions from SfM sparse point cloud~\cite{Zhang0LFMTQ22,FuXOT22} or monocular depth/normal~\cite{YuPNSG22} may mitigate the difficulty but it's out of our scope.
We propose simple and effective regularization losses for explicit SDF grid optimization, achieving fast and high-quality surfaces without using external priors.

\paragraph{Material and lighting estimation.} 
Several neural reconstruction methods \cite{ZhangLWBS21,ZhangSDDFB21,BossBJBLL21,BossJBLBL21,MunkbergCHES0GF22,HasselgrenHM22,ZhangSHFJZ22} adopt the same setting as ours to simultaneously reconstruct geometry, material, and lighting from multiple images. An earlier work \cite{ZhangSDDFB21} directly optimizes a low-resolution environment map and materials from fixed geometry, leading to noisy lighting reconstruction caused by the highly ill-posed nature of this problem. More recent methods model lighting with a mixture of spherical Gaussians \cite{ZhangLWBS21,BossBJBLL21,ZhangSHFJZ22} or pre-filtered approximation \cite{BossJBLBL21,MunkbergCHES0GF22}, and constrain material reflection with a spherical Gaussian lobe \cite{ZhangLWBS21} or a low dimensional latent code \cite{BossBJBLL21,BossJBLBL21,ZhangSHFJZ22} to improve the reconstruction quality. Most of them only consider direct illumination without occlusion \cite{ZhangLWBS21,BossBJBLL21,MunkbergCHES0GF22} or use a fixed shadow map \cite{ZhangSDDFB21,BossJBLBL21} due to the high computational cost of MLP-based neural rendering. Nvdiffrecmc \cite{HasselgrenHM22} models shadows using a differentiable ray tracer with a denoiser that significantly reduces the number of samples but still cannot model interreflection. MII \cite{ZhangSHFJZ22} trains another network to predict visibility and interreflection from the radiance field, which takes several hours. On the contrary, our hybrid, highly optimized neural SDF framework enables us to efficiently extract geometry, materials, lighting, visibility, and interreflection in less than 10 minutes. Our advanced PBIR framework for the first time allows holistic optimization of lighting, material, and geometry with direct and indirect illumination modeled in a physically unbiased way. 

\paragraph{Physics-based inverse rendering.}
Different from neural-based methods, several recent works \cite{luan2021unified, HasselgrenHM22, cai2022physics} utilize classic image formulation models developed in the graphics community and try to inverse this process using gradient-based optimizations, with the gradients computed using physics-based differentiable renderers. These approaches can handle complex light transport effects during optimization but are prone to local minima and overfitting. In our work, instead of optimizing from scratch, we perform PBIR as a refinement stage for better robustness.

%% file: ours.tex
\section{Our Method}
\label{sec:ours}

Provided multi-view images of an opaque object under fixed unknown illumination (with known camera parameters), our technique reconstructs the shape and reflectance of the object as well as the illumination condition.

As illustrated in \cref{fig:pipeline}, our pipeline is comprised of three main stages.
The first stage (\cref{ssec:ours_surface}) is \emph{a fast and precise surface reconstruction step} that brings direct SDF grid optimization into \textsf{NeuS}~\cite{WangLLTKW21}.
Associated with this surface is an overfitted radiance field that does not fully model the surface reflectance of the object.
Our second stage (\cref{ssec:ours_coarse_fit}) is \emph{an efficient neural distillation method} that converts the radiance fields to physics-based reflectance~\cite{burley2012physically} and illumination models.
Lastly, our third stage (\cref{ssec:ours_pbir}) utilizes \emph{physics-based inverse rendering} (PBIR) to further refine the object geometry and reflectance reconstructed by the first two stages.
This stage leverages physics-based differentiable rendering that captures global illumination (GI) effects such as soft shadows and interreflection.

\input{ours_surface}

\input{ours_coarse_fit}

\input{ours_pbir}

%% file: ours_surface.tex
\subsection{Neural Surface Reconstruction}
\label{ssec:ours_surface}
Taking as input multiple images of an opaque object under fixed illumination (and with known camera parameters), the first stage of our method produces a detailed reconstruction of the object's surface.
To this end, we optimize the object surface and an outgoing radiance field that best describes the input images.
In this stage, for efficiency and robustness, we express the object surface as the zero-level set $\{ \bx \in \Real^3 \;|\; \sdf(\bx) = 0 \}$ of a signed distance field (SDF) $\sdf(\bx)$, and the radiance field $\Lo(\bx, -\bv)$ (for any position $\bx \in \Real^3$ and viewing direction $\bv \in \Sph$) as a non-physics-based general function approximator.

\paragraph{Unbiased volume rendering with SDF.}
To compute the color of a pixel, we first sample $N$ query points $\{ \bx_i = \bo + t_i \bv \}_{i=1}^N$ (with $0 < t_1 < t_2 < \ldots < t_N$) along the corresponding camera ray originated at the camera's location $\bo$ with viewing direction $\bv$.
We then query the scene representation for points sign distance $\sdf(\bx_i)$ and radiance $\Lo(\bx_i, \bv)$.
Following \neus~\cite{WangLLTKW21}'s unbiased rendering, we activate the queried signed distance into alpha for all $1 \leq i \leq N$ via
\begin{small}
\begin{equation}
    \label{eqn:alpha}
    \alpha_i = \max\left(0,\ \frac{\sigma(\sdf(\bx_i)) - \sigma(\sdf(\bx_{i+1}))}{\sigma(\sdf(\bx_i))}\right) \,,
\end{equation}
\end{small}
where $\sigma$ denotes the Sigmoid function.
Then, the pixel color $C$ is computed by the alpha blending of the queried alphas and radiance values
\begin{small}
\begin{equation}
    \label{eqn:pix_C}
    C = \sum_{i=1}^N T_i \,\alpha_i \,\Lo(\bx_i, -\bv) \quad \text{where $T_i = \prod_{j=1}^{i-1} (1-\alpha_j)$.}
\end{equation}
\end{small}

\paragraph{Voxel-based scene representation.}
\neus uses a large MLP to model the SDF $\sdf$ and radiance field $\Lo$.
Unfortunately, since this MLP is expensive to query, \neus' optimization processes can be very time-consuming.

For better performance, we adapt the dense-grid-based scene representation from \textsf{DVGO}~\cite{SunSC22} by setting
\begin{small}
\begin{equation}
\label{eqn:SDF_Lo}
\begin{aligned}
    \sdf(\bx) &= \operatorname{interp}(\bx,\ \Vsdf)\;,\\ 
    \Lo(\bx, -\bv) &= \operatorname{MLP}\left( \operatorname{interp}(\bx,\ \Vfeat),\ \bv \right)\,,
\end{aligned}
\end{equation}
\end{small}
where $\operatorname{interp}()$ indicates trilinear interpolation, $\Vsdf$ is a dense SDF grid, $\Vfeat$ is a dense feature grid, and $\operatorname{MLP}$ is a single-hidden-layer MLP with ReLU activation.

In practice, we use two sets of $\Vsdf$ and $\Vfeat$ grids to model the foreground (i.e., the object of interest) and the background with varying resolutions.
Please refer to the supplement for more details.

\paragraph{Adaptive Huber loss.}
We incorporate a Huber loss to reduce the impact of specular highlights which can cause bumpy artifacts in reconstructed surfaces:
\begin{small}
\begin{equation}
    \label{eqn:Lphoto}
    \Lphoto = \sum_{r} \begin{cases}
        (\img[r] - C[r])^2, & (|\img[r] - C[r]| \leq t)\\
        2t\,|\img[r] - C[r]| - t^2, & (\text{otherwise})
    \end{cases}
\end{equation}
\end{small}
where: %
$\img[r]$ and $C[r]$ denote, respectively, the colors of the $r$-th pixels of the target and rendered images; $t \in \Real_{> 0}$ is a hyper-parameter. %

Intuitively, the infinity norm of the gradient of $\Lphoto$ is clamped by $2t$ to prevent %
bright pixels (\eg, those exhibiting specular highlights) dominating the optimization.
In practice, %
instead of tuning a constant $t$, we adaptively update $t$ by the running mean of the median of $|\img[r] - C[r]|$ in each iteration to clamp gradient for about half of the pixels.

\paragraph{Laplacian regularization.}
To further improve the robustness of our approach and reduce potential artifacts, we regularize our SDF grid $\Vsdf$ using a Laplacian loss:
\begin{small}
\begin{equation}
    \label{eqn:Lap}
    \Llap = \sum_{u} \bigg(\sum_{u' \in N(u)} \left(\Vsdf[u'] - \Vsdf[u]\right)\bigg)^2 ~,
\end{equation}
\end{small}
where $\Vsdf[u]$ indicates the SDF value stored at the grid point $u$, and $N(u)$ denotes the six direct neighbors of $u$.

\paragraph{Training.}
Following \textsf{DVGO}~\cite{SunSC22}, we use progressive grid scaling for efficiency and more coherent results.
Our experiments also indicate that using a per-point RGB loss $\Lrgb$ improves convergence speed and quality:
\begin{small}
\begin{equation}
    \label{eqn:Lrgb}
    \Lrgb = \sum_{r,i} T_i^{(r)} \alpha_i^{(r)} \left| \Lo\left(\bx_i^{(r)}, -\bv^{(r)}\right) - \img[r] \right| \,,
\end{equation}
\end{small}
where $\alpha_i^{(r)}$ and $T_i^{(r)}$ are computed using \cref{eqn:alpha,eqn:pix_C} for each pixel $r$.
In summary, to train our SDF and feature grids $\Vsdf$ and $\Vfeat$ in \cref{eqn:SDF_Lo}, we minimize the objective:
\begin{small}
\begin{equation}
    \label{eqn:Lsurf}
    \Lsurf = \Lphoto + \Wlap \,\Llap + \Wrgb \,\Lrgb ~.
\end{equation}
\end{small}

\paragraph{Postprocessing.}
Optionally, our surface reconstruction stage generates for each input image an anti-aliased mask that separates the object from the background using mesh rasterisation and alpha matting~\cite{ParkSYKK22}.
The generated masks can be used by our physics-based inverse rendering stage (\cref{ssec:ours_pbir}) to facilitate the refinement of object shapes.

%% file: ours_coarse_fit.tex
\subsection{Neural material and lighting distillation}
\label{ssec:ours_coarse_fit}
Provided the optimized object surface and outgoing radiance field from the surface reconstruction stage (\cref{ssec:ours_surface}), the goal of the second stage of our pipeline is to obtain an initial estimation of surface reflectance and illumination condition.
To this end, we leverage the radiance field $\Lo$ as the teacher model to distill the learned surface color into %
physics-based material and illumination models (that can be rendered to reproduce $\Lo$) as follows.

As preprocessing for this stage, we extract a triangle mesh $M^{(0)}$ from the optimized SDF $\sdf(\bx)$ given by \cref{eqn:SDF_Lo} using (non-differentiable) marching cube.
Additionally, for each mesh vertex $\vind$, we compute its normal $M_{\bn}[\vind] \in \Sph$ based on the gradient $\nabla\sdf$ of the SDF.

\paragraph{Material and illumination models.}
To model spatially varying surface reflectance, we use the widely-adopted Disney microfacet BRDF~\cite{karis2013real} parameterized by surface albedo and roughness. %
In practice, instead of using a large MLP to model the spatially varying BRDF parameters~\cite{ZhangLWBS21,ZhangSDDFB21,MunkbergCHES0GF22,ZhangSHFJZ22}, we store the albedo $\Ma[\vind] \in [0, 1)$ and roughness $\Mr[\vind] \in \Real_{> 0}$ for each vertex $\vind$ of the extracted mesh $M^{(0)}$ (and interpolate them in the interior of triangle faces) for better performance. %
To model environmental illumination, we use mixtures of spherical Gaussians~\cite{WangRGSG09,MederB18,ZhangLWBS21,ZhangSHFJZ22} with $256$ lobes. %

\paragraph{Coarse rendering.}
For fast training, we opt to use numerical integration on stratified pre-sampled light directions $\wiset$ for efficiency (we use $256$ samples in practice).
Specifically, for each vertex $\vind$ of the extracted mesh $M^{(0)}$ and light direction $\bomi \in \wiset$, we precompute the visibility $\Vis[\vind, \bomi]$ and indirect illumination $\IndIl[\vind, \bomi]$ by tracing a ray from the vertex in the direction $\bomi$.
If this ray intersects the object surface at some $\bx$, we set $\Vis[\vind, \bomi] = 0$ and %
$\IndIl[\vind, \bomi] = \Lo(\bx, -\bomi)$~\cite{ZhangSHFJZ22}.
Otherwise, we set $\Vis[\vind, \bomi] = 1$.
Then, the incident radiance at each vertex can be expressed as
\begin{small}
\begin{equation}
    \label{eqn:Li}
    \Li(\bomi) = \LenvSG(\bomi)\,\Vis[\bomi] + \IndIl[\bomi]\left(1-\Vis[\bomi]\right) ~,
\end{equation}
\end{small}
where $\LenvSG$ is the SG-based environmental illumination.
Lastly, the outgoing radiance is computed by
\begin{small}
\begin{equation}
    \label{eqn:coarse_render}
    \Lostudent(\bomo) = \frac{1}{Z} \sum_{\bomi \in \wiset} \Li(\bomi) \,\brdf(\bomi, \,\bomo, \,\Mn) \,(\Mn \cdot \bomi) ~,
\end{equation}
\end{small}
where $Z$ is a normalization factor, $\brdf$ denotes the BRDF function parameterized by $\Ma$ and $\Mr$.
In \cref{eqn:Li,eqn:coarse_render}, we omit dependencies on vertex $\vind$ for brevity.

\paragraph{Training.}
In each iteration, we randomly sample a outgoing direction $\bomo$ (with $\Mn[\vind] \cdot \bomo[\vind] > 0$) for each vertex $\vind$ and minimize the absolute difference with the teacher model:
\begin{small}
\begin{equation}
    \label{eq:loss_dist}
    \Ldt = \sum_{\vind} \left| \Lostudent\left(\vind, \bomo[\vind]\right) - \Lo\left(\vind, \bomo[\vind]\right) \right| \,.
\end{equation}
\end{small}
To encourage smoothness, we apply total variation loss on the per-vertex albedo $\Ma$ and roughness $\Mr$:
\begin{small}
\begin{equation}
    \label{eq:loss_dist_tv}
    \Ldtreg = \sum_{(\vind_1,\vind_2)\in\text{edge}} \;\sum_{* \in \{ \mathrm{a}, \mathrm{r} \}} \left| M_*[\vind_1] - M_*[\vind_2] \right| \,.
\end{equation}
\end{small}
Also, we regularize the SG-based illumination $\LenvSG$ using
\begin{small}
\begin{equation}
    \Ldtbg = \sum_{\bomi \in \wiset} \left| \LenvSG(\bomi) - (\LenvSG)'(\bomi) \right| \,,
\end{equation}
\end{small}
where $(\LenvSG)'$ indicates the averaged background observation (see the supplement for more details).

In summary, we optimize per-vertex albedo $\Ma$, roughness $\Mr$, and the environmental illumination $\LenvSG$ by minimizing
\begin{small}
\begin{equation}
    \Ldttotal = \Ldt + \Wdtreg\,\Ldtreg + \Wdtbg\,\Ldtbg ~.
\end{equation}
\end{small}

\paragraph{Postprocessing.}
With the per-vertex attributes $\Ma$ and $\Mr$ obtained, we UV-parameterize the extracted mesh $M^{(0)}$ and pixelize $\Ma$ and $\Mr$, respectively, into an abledo map $\Ta^{(0)}$ and a roughness map $\Tr^{(0)}$.
These maps will be refined by our following inverse-rendering stage (\cref{ssec:ours_pbir}).

%% file: ours_pbir.tex
\subsection{Physics-Based Inverse Rendering}
\label{ssec:ours_pbir}
Reconstructions produced by our shape reconstruction and neural material distillation stages (\cref{ssec:ours_surface,ssec:ours_coarse_fit}) comprise three components: (i)~environmental illumination $\Lenv^{(0)} \defeq \LenvSG$ expressed as spherical Gaussians; (ii)~object reflectance parameterized with an albedo map $\Ta^{(0)}$ and a roughness map $\Tr^{(0)}$; and (iii)~object surface mesh $M^{(0)}$.

Despite being high-fidelity, these reconstructions can still lack sharp details and are not immune to artifacts (see \cref{fig:abla_pbir}).
To further improve reconstruction quality, we utilize a physics-based inverse rendering stage.
Specifically, initialized our neural reconstructions (i.e., $\Lenv^{(0)}$, $\Ta^{(0)}$, $\Tr^{(0)}$, and $M^{(0)}$), we apply gradient-based optimization to minimize the \emph{inverse-rendering loss}:
\begin{small}
\begin{equation}
    \label{eqn:Lir}
    \Lir = \Limg + \Wmask\,\Lmask + \Wreg\,\Lreg,
\end{equation}
\end{small}
where
\begin{small}
\begin{equation}
    \label{eqn:Limg}
    \Limg(\Lenv, \Ta, \Tr, M) = \sum_j \left\| \img_j - \R_j(\Lenv, \Ta, \Tr, M) \right\|_1,
\end{equation}
\end{small}
is the \emph{image loss} given by the sum of L1 losses between the $j$-th target image $\img_j$ and the corresponding rendered image $\R_j(\Lenv, \Ta, \Tr, M)$.
Additionally, $\Lreg$ and $\Lmask$ in \cref{eqn:Lir} denote, respectively, the \emph{regularization} and the \emph{mask} losses---which we will discuss in the following.

\paragraph{Environment map optimization.}
We recall that the initial illumination $\Lenv^{(0)} = \LenvSG$ is a coarse reconstruction expressed as a set of spherical Gaussians (SGs).
To further refine it, we employ a two-step process as follows.
In the first step, we directly optimize the SG parameters (i.e., per-lobe means and variances).
In the second step, we pixelize the optimized SG representation into an environment map (using the latitude-longitude parameterization) and perform per-pixel optimization. %

\paragraph{SVBRDF optimization.}

Since our initializations $\Ta^{(0)}$ and $\Tr^{(0)}$ are already high-quality, we perform per-texel optimization for the albedo and roughness maps $\Ta$ and $\Tr$.
Further, to make our optimization less prune to Monte Carlo noises produced by our physics-based renderer (discussed later), we regularize $\Tr$ using a total variation loss:
\begin{small}
\begin{equation}
    \label{eqn:Lreg}
    \Lreg(\Tr) = \sum_{(x,y)} \sum_{(x',y')} \left| \,\Tr[x',y'] - \Tr[x,y] \,\right| \,,
\end{equation}
\end{small}
where $\Tr[x, y]$ denotes the $(x, y)$-th texel of the roughness map $\Tr$, and $(x', y') \in \{(x + 1, y),\ (x, y + 1) \}$ are two direct neighbors of $(x, y)$.

\paragraph{Shape refinement.}
In all our experiments, object geometries predicted by our neural stages accurately recover object topology.
Thus, although it is possible to directly optimize SDFs in inverse rendering~\cite{Bangaru2022NeuralSDFReparam,Vicini2022sdf}, 
we opt to use explicit mesh-based representations for the object surface $M$ in this stage for better efficiency.
To make our per-vertex optimization %
more robust to Monte Carlo noises, we utilize Nicolet et al.'s \textsf{AdamUniform} optimizer~\cite{Nicolet2021Large}.

To further improve efficiency, we leverage object masks produced either as input or by our surface reconstruction (\cref{ssec:ours_surface}) and introduce a mask loss:
\begin{small}
\begin{equation}
    \label{eqn:Lmask}
    \Lmask(M) =\sum_{j} \| S_j - \Rmask_j(M) \|_1,
\end{equation}
\end{small}
where $S_j$ is our predicted mask for the $j$-th target image and $\Rmask_j$ the corresponding rendered mask.

\paragraph{Differentiable rendering.}
To differentiate the image and the mask losses defined in Eqs.~\eqref{eqn:Limg} and \eqref{eqn:Lmask}, we develop a physics-based Monte Carlo differentiable renderer that implements path-space differentiable rendering~\cite{Zhang:2020:PSDR} and utilizes state-of-the-art numerical backend~\cite{Jakob2022DrJit}.

Most, if not all, previous techniques including \mii~\cite{ZhangSHFJZ22} and \nvdiff~\cite{HasselgrenHM22} rely on highly simplified differentiable rendering processes that typically neglect global-illumination (GI) effects and produce biased gradients with respect to geometry.
On the contrary, our differentiable renderer offers unbiased gradients and the generality of differentiating GI (and anti-aliased masks) with respect to surface geometries.
Consequently, our pipeline is capable of generating higher-quality reconstructions than state-of-the-art methods, which we will demonstrate in \cref{sec:evaluation}.

%% file: eval.tex
\section{Results}
\label{sec:evaluation}
To demonstrate the effectiveness of our method, we show our reconstructions on synthetic input images and captured photographs in \cref{ssec:synthetic_data,ssec:real_data}, respectively.
We compare reconstructions obtained with our technique with two state-of-the-art baselines: \mii~\cite{ZhangSHFJZ22} and \nvdiff~\cite{HasselgrenHM22}.
Additionally, we conduct ablation studies to evaluate several components of our pipeline in \cref{ssec:ablation}.
Please refer to the supplement for more results.

\input{tabs/mii_table}

\input{tabs/mii_synthetic_visual}

\subsection{Synthetic Data}
\label{ssec:synthetic_data}
We assess the effectiveness of our proposed method using the synthetic dataset made available by \mii~\cite{ZhangSHFJZ22}.
This dataset comprises four virtual scenes---each of which includes multi-view renderings (with object masks) of an object under some natural environmental illumination (with the ground truth environment map provided).
For each scene, the dataset provides 100 target images accompanied by camera poses and object masks for training.
Additionally, the testing set consists of renderings of ground truth albedo, roughness, and the object under two novel lighting conditions in 200 poses. 
We apply our technique and the baselines %
to the posed training target images and masks (but not the GT environment maps).

\cref{tab:mii} presents the quantitative comparisons where we compare our method's reconstructions with the baselines by rendering them in the testing poses and comparing them with the ground truth images. In line with \mii, we report PSNR, SSIM, and LPIPS for the relit and albedo images, and the MSE for the roughness images averaged across all scenes. 
\mii also reports error metrics on aligned albedo to reduce the impact of albedo-light ambiguity \cite{ZhangSHFJZ22}. 
Specifically, we compute an RGB scale that minimizes the difference between the reconstructed albedo images and the ground truth albedo images, and reconstructed albedo images are scaled before evaluation. Our method outperforms the baselines and takes less time. %

We show qualitative results in \cref{fig:mii_vis}. Since \mii imposes a sparsity constraint to reconstruct a sparse set of materials, the results tend to be over-blurred.
Despite being able to produce sharp texture maps, \nvdiff suffers from inaccurate illumination reconstructions, causing their reconstructed albedo maps to be color-shifted.
Compared with these two baselines, our method produces significantly more accurate material and illumination reconstructions.

\input{tabs/ours_real_table}

\input{tabs/ours_real_nv}

\input{tabs/ours_real_relit}

\input{tabs/ours_real_relit_novel}

\input{tabs/ablation_pbir}

\subsection{Real Data}
\label{ssec:real_data}
To further demonstrate the robustness of our method, we captured five real scenes under indoor lighting and again compared the reconstruction quality with \mii and \nvdiff.
For each scene, we use around 200 measured images for training and 10--20 for testing.
We place a checkerboard beneath the object for the geometric calibration of camera parameters. %
For each object, we manually annotate an object 3D bounding box to crop the reconstructed meshes to focus the comparisons on the object of interest.
In addition, we measure ground truth environment map (which we use for evaluation only) for each scene using an HDR 360 camera. As there are no ground truth foreground masks while our baselines require them, we use masks generated in our surface reconstruction stage (\S\ref{ssec:ours_surface}).

We compare the qualities of reconstructions produced by our method and the baselines using novel-view renderings (since we do not have the GTs).
As shown in \cref{fig:ours_real_nv}, our reconstructions produce detailed renderings that closely resemble the GTs, while \mii and \nvdiff's results suffer from various artifacts (\eg lacking details and bumpy surfaces).
In addition, to evaluate the quality of reconstructed object shape and reflectance, we compare renderings under captured GT illumination in \cref{fig:ours_real_relit}.
The quantitative comparison is summarized in \cref{tab:ours_real_table} where we achieve the best novel-view synthesis results under both the reconstructed and the captured environment maps.
We show qualitative comparison under novel-illumination in \cref{fig:ours_real_relit_wild}.

\subsection{Evaluations and Ablations}
\label{ssec:ablation}
\paragraph{Surface reconstruction.}
To evaluate the quality of reconstructed geometries produced by our surface reconstruction stage (\cref{ssec:ours_surface}), we use the subsampled 15 scenes of DTU dataset~\cite{JensenDVTA14}.
Each scene has 49 or 64 images with camera parameters and masks provided.
The quantitative results are concluded in \cref{tab:dtu}.
Our direct SDF grid optimization uses significantly less time than \neus~\cite{WangLLTKW21} while still outperforming \neus surface accuracy measured in chamfer distance (CD).
Our simple surface reconstruction with dense grid also achieves similar performance comparing to the most recent \textsf{Voxurf}~\cite{WuWPXTLL22} and \textsf{NeuS2}~\cite{WangHHDTL22}.

\input{tabs/dtu_table}
\input{tabs/ablation_surface}

We conduct a comprehensive ablation study for the three losses that help us achieve direct SDF grid optimization.
As shown in \cref{tab:dtu_ablate}, naively adapting \textsf{DVGO} without regularization leads poor quality.
Among the regularization, $\Llap$ is the most significant, which enforces the SDF grid to evolve smoothly during optimization.
Adding $\Lrgb$ and using adaptive Huber loss also shows good improvement.
Combing all three losses offers the best results.

\setlength{\columnsep}{6pt}
\begin{wrapfigure}{r}{0.55\linewidth}
\centering
\vspace{-1em}
\includegraphics[width=.97\linewidth]{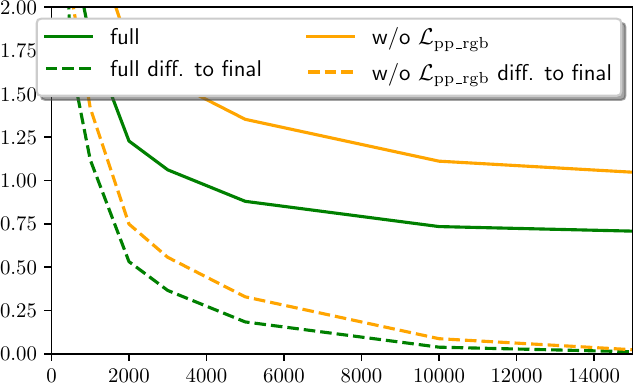}
\vspace{-2em}
\end{wrapfigure}
\definecolor{fig_green}{rgb}{.21, .50, .13}
\definecolor{fig_orange}{rgb}{.95, .66, .23}
We also show the chamfer distance (CD) in the first 15k iterations \textcolor{fig_green}{with} and \textcolor{fig_orange}{without} $\Lrgb$ here.
The solid lines are CD, while the dashed lines are the difference to the final CD.
The results are averaged over the 15 DTU scenes.
Using $\Lrgb$ improves final quality and speeds up convergence to the final CD in fewer iterations.

Please note that our results on DTU dataset are directly from the shape reconstruction stage.
We skip evaluating the shape refinement of our physics-based inverse rendering as DTU exhibit vary light occlusion from robot arms.
Please see the supplement for more results.

\paragraph{Neural material distilling.}
We now demonstrate the usefulness of our neural material distilling stage (\cref{ssec:ours_coarse_fit})---which provides high-quality material predictions used to initialize our PBIR stage (\cref{ssec:ours_pbir}).
Specifically, when using constant initializations for surface reflectance (with $\Ta = 0.5$, $\Tr = 0.5$) and illumination (with a gray environment map), it takes our PBIR pipeline 37 minutes (see the row labeled as ``Ours - const. init.'' in \cref{tab:mii}) to generate reconstructions with a similar level of accuracy as the neural distilling stage does in less than 15 minutes (see the row labeled as ``Ours - distilled only'').

\paragraph{Importance of PBIR.}
The distilling method is efficient by sidestepping ray tracing and approximating global illumination but it sacrifices accuracy and tends to "bake" interreflection into the materials.
We use our physically-based differentiable renderer to further refine the material.
In \ref{fig:abla_pbir}, we show the albedo maps  before and after PBIR on our real dataset.
We achieve much higher fidelity with less baking after PBIR.
The result is consistent with the quantitative results on synthetic dataset (refer to "Ours - Distilled only" versus "Ours - Full") in \ref{tab:mii}.

\paragraph{Importance of GI.}
Our differentiable renderer used in the PBIR stage (\cref{ssec:ours_pbir}) is capable of handling global-illumination (GI) effects such as interrelfection.
To demonstrate the importance of GI, we run PBIR optimizations with and without GI and compare the material reconstruction qualities.
As shown in rows labeled as ``Ours - w/o GI'' and ``Ours - w/o shape ref.'' in \cref{tab:mii}, enabling GI improves the accuracy of material and lighting reconstructions.  

A main reason for the usefulness of GI is that, without GI, inverse-rendering optimizations tend to ``bake'' effects like interreflections into reflectance (\eg albdeo) maps, limiting their overall accuracy.
We demonstrate this by comparing albedo maps optimized with and without GI in \cref{fig:ablation_gi} (using the balloons data from the \mii dataset).

\input{tabs/ablation_gi}

\input{tabs/abla_shape_refine_pig}

\paragraph{Shape refinement.}
Lastly, we demonstrate the usefulness of shape refinement (by optimizing the vertex positions of the extracted mesh) using rows labeled as ``Ours - w/o shape ref.'' and ``Ours - Full'' in \cref{tab:mii,fig:our_vis_abla_2}.
Please refer to the supplement for more examples.

%% file: tabs/mii_table.tex
\begin{table*}
\centering
{
\resizebox{.9\textwidth}{!}{
\begin{tabular}{ @{}l@{\hskip 5pt} @{}c@{} c@{\hskip 7pt} @{}c@{\hskip 5pt} c@{\hskip 5pt} c@{} c@{\hskip 7pt} @{}c@{\hskip 5pt} c@{\hskip 5pt} c@{} c@{\hskip 7pt} @{}c@{\hskip 5pt} c@{\hskip 5pt} c@{} c@{\hskip 7pt} @{}c@{} }
    \toprule
    &Speed && \multicolumn{3}{c}{Relighting} & &\multicolumn{3}{c}{Aligned albedo} & & \multicolumn{3}{c}{Albedo} && Rough. \\
    \cmidrule{2-2} 
    \cmidrule{4-6}
    \cmidrule{8-10}
    \cmidrule{12-14}
    \cmidrule{16-16}
    Method & Time$\downarrow$ & & PSNR$\uparrow$ & SSIM$\uparrow$ & LPIPS$\downarrow$ & &PSNR$\uparrow$ & SSIM$\uparrow$ & LPIPS$\downarrow$ & & PSNR$\uparrow$ & SSIM$\uparrow$ & LPIPS$\downarrow$ && MSE$\downarrow$ \\
    \midrule
    \nvdiff~\cite{HasselgrenHM22} & $\sim$2 h & & 23.93 & 0.946 & 0.074 && \textbf{29.72} & \textbf{0.959} & \textbf{0.057} &&  18.25 & 0.899 & 0.103 && \underline{0.009} \\
    \mii~\cite{ZhangSHFJZ22} & $\sim$10 h & & 27.53 & 0.947 & 0.087 && 25.77 & 0.935 & \underline{0.066} &&  24.62 & 0.931 & \textbf{0.064} && \textbf{0.008} \\
    \midrule
    Ours - Distilled only & $<$15 m & & 30.26 & 0.961 & 0.059 && 27.67 & 0.933 & 0.079 &&  26.20 & 0.931 & 0.093 && \underline{0.009} \\
    Ours - Const. init. & $\sim$ 37 m & & 30.25 & {\bf 0.970} & 0.050 && 28.55 & 0.940 & 0.070 &&  25.83 & 0.940 & 0.080 && \underline{0.010} \\
    Ours - w/o GI & $\sim$45 m & & 30.57 & 0.960 & 0.050 && 27.71 & 0.940 & 0.070 &&  26.38 & 0.940 & 0.082 && \underline{0.009} \\
    Ours - w/o shape ref. & $\sim$45 m & & \underline{30.61} & 0.965 & \underline{0.049} && 28.74 & 0.944 & 0.067 &&  \underline{26.84} & \underline{0.941} & 0.082 && \textbf{0.008} \\
    Ours - Full & $\sim$1 h & & \textbf{30.73} & \underline{0.966} & \textbf{0.047} && \underline{29.06} & \underline{0.946} & 0.067 &&  \textbf{26.85} & \textbf{0.944} & \underline{0.080} && \textbf{0.008} \\
    \bottomrule
\end{tabular}
}
}
\vspace{-0.5em}
\caption{
{\bf Relighting, material reconstruction, and view-interpolation quality on MII dataset~\cite{ZhangSHFJZ22}.} We compare our method with MII and Nvdiffrec-mc. The highest performing number is presented in bold, while the second best is underscored. For ``Ours - Distilled only'', we did not run the PBIR stage. For ``Ours - Const. init.'' and ``Ours - w/o GI'', we did not run the shape refinement. Please refer to \cref{ssec:synthetic_data,ssec:ablation} for more details.
}
\label{tab:mii}
\end{table*}

%% file: tabs/mii_synthetic_visual.tex
\begin{figure}
    \centering
    \includegraphics[width=.98\columnwidth]{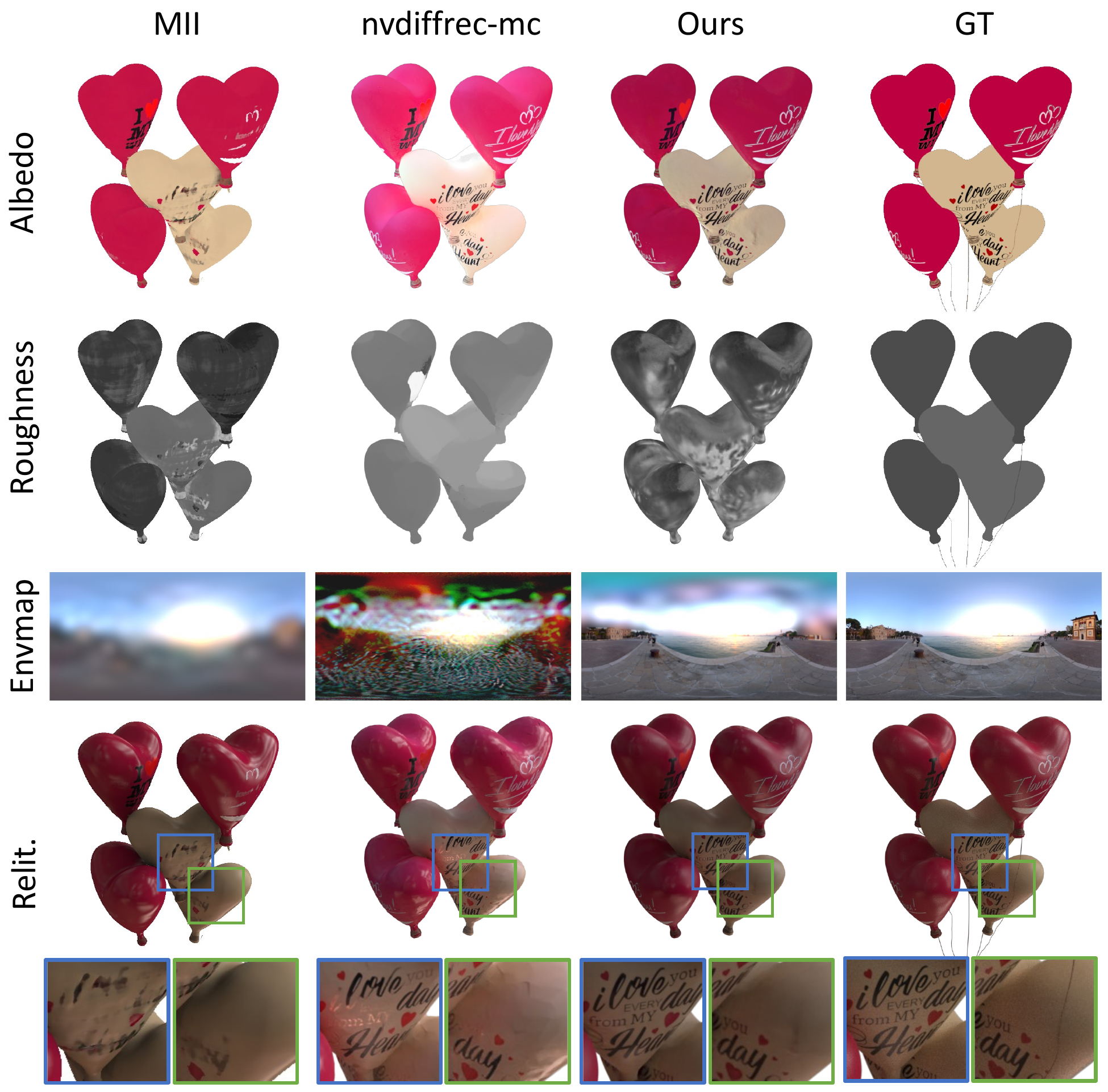}
    \vspace{-0.3cm}
    \caption{\bf Qualitative comparisons on the MII data.}
    \label{fig:mii_vis}
\end{figure}

%% file: tabs/ours_real_table.tex
\begin{table}
\centering
\resizebox{\columnwidth}{!}{
\begin{tabular}[t]{ @{}l@{\hskip 5pt} @{}c@{\hskip 3pt} c@{} c@{\hskip 5pt} @{}c@{\hskip 3pt} c@{} c@{\hskip 5pt} @{}c@{\hskip 3pt} c@{} }
    \toprule
    \multirow{3}{*}{Methods} & \multicolumn{2}{c}{Novel-view} && \multicolumn{5}{c}{Captured light re-rendering} \\
    \cmidrule{5-9}
    & & & & \multicolumn{2}{c}{raw} && \multicolumn{2}{c}{aligned} \\
    \cmidrule{2-3}
    \cmidrule{5-6}
    \cmidrule{8-9}
    
    & PSNR$\uparrow$ & SSIM$\uparrow$ && PSNR$\uparrow$ & SSIM$\uparrow$ && PSNR$\uparrow$ & SSIM$\uparrow$ \\
    \midrule
    {\footnotesize Nvdiffrecmc} & 30.3 & 0.94 && 21.8 & 0.92 && 28.0 & 0.94 \\
    MII                         & 28.9 & 0.94 && 27.5 & 0.94 && 28.6 & 0.94 \\
    Ours                        & {\bf 31.6} & {\bf 0.96} && {\bf 28.8} & {\bf 0.95} && {\bf 30.7} & {\bf 0.95} \\
    \bottomrule
\end{tabular}
}
\vspace{-0.3cm}
\caption{
{\bf Quantitative comparison on Our Real Dataset.}
To inspect material quality for relighting, we capture 360 images for our dataset and evaluate the rendering results under the captured lighting.
}
\label{tab:ours_real_table}
\end{table}

%% file: tabs/ours_real_nv.tex
\begin{table*}
\setlength{\resLen}{.8in}
\centering
\begin{tabular}{@{}c@{\hskip 2pt}c@{\hskip 3pt}c@{\hskip 3pt}c@{\hskip 3pt}c@{\hskip 3pt}c@{}}
    & {\it dinosaur} & {\it greenplant} & {\it plantpot} & {\it pumpkin} & {\it shoe} \\
    \vtext{\footnotesize \textsf{Groundtruth}} &
    \includegraphics[height=\resLen]{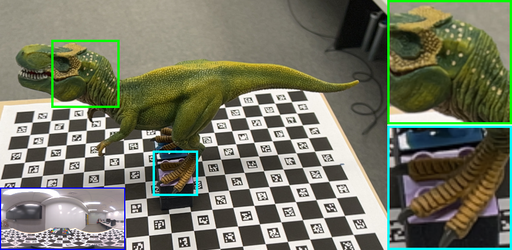} &
    \includegraphics[height=\resLen]{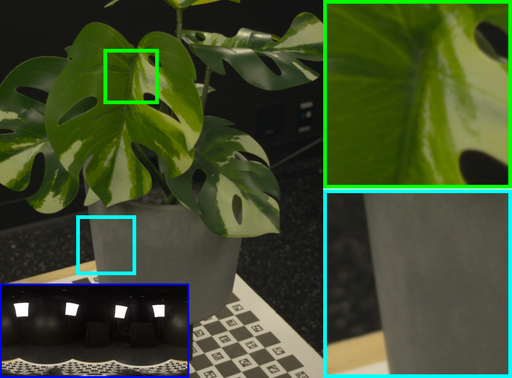} &
    \includegraphics[height=\resLen]{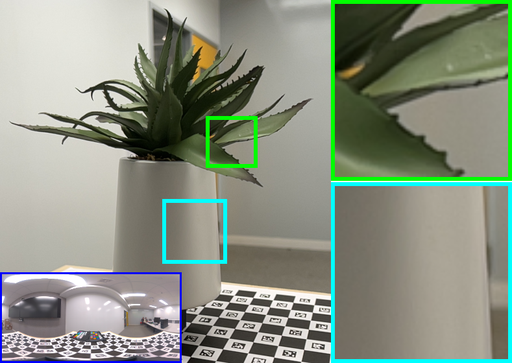} &
    \includegraphics[height=\resLen]{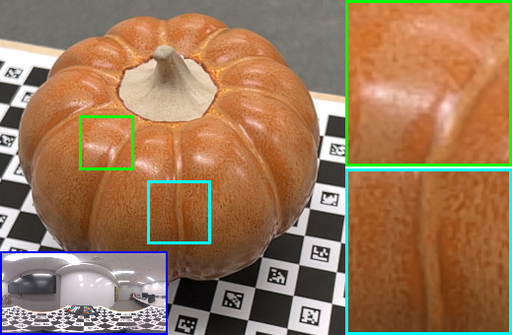} &
    \includegraphics[height=\resLen]{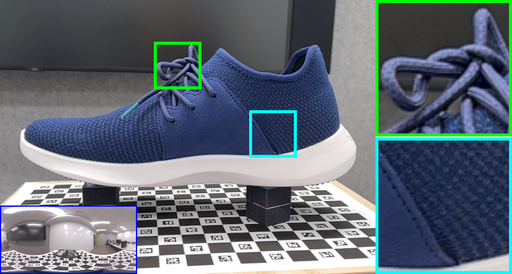} \\
    \vtext{\footnotesize \textsf{Ours}} &
    \includegraphics[height=\resLen]{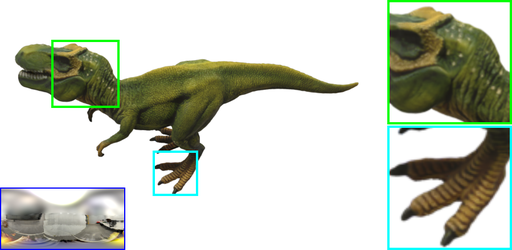} &
    \includegraphics[height=\resLen]{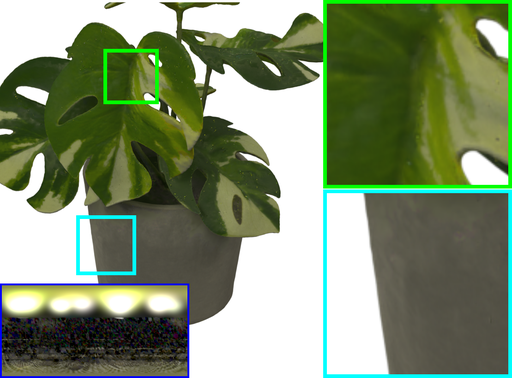} &
    \includegraphics[height=\resLen]{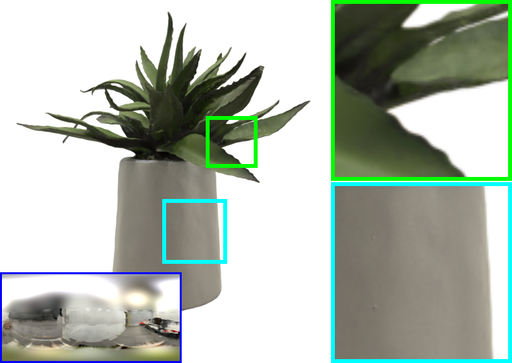} &
    \includegraphics[height=\resLen]{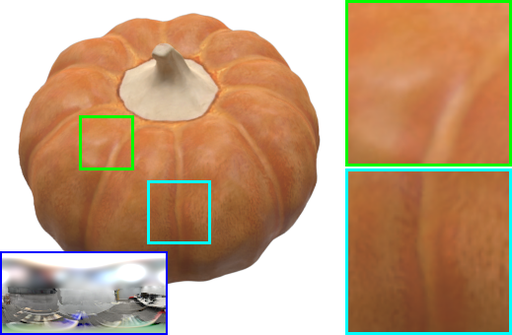} &
    \includegraphics[height=\resLen]{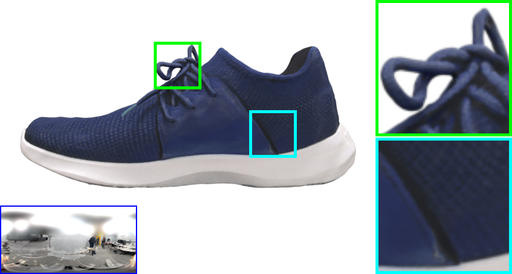} \\
    & {\bf\scriptsize 33.7 / 0.97} & {\bf\scriptsize 29.2 / 0.93} & {\bf\scriptsize 32.1 / 0.98} & {\scriptsize 33.3 / 0.97} & {\bf\scriptsize 29.9 / 0.93} \\
    \vtext{\footnotesize \nvdiff} &
    \includegraphics[height=\resLen]{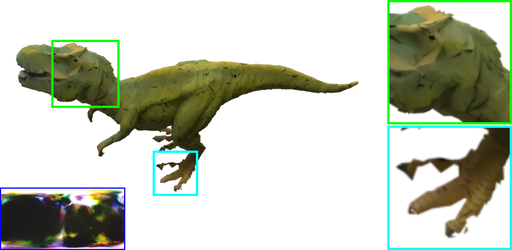} &
    \includegraphics[height=\resLen]{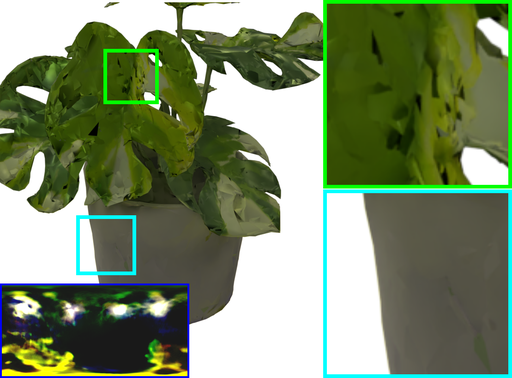} &
    \includegraphics[height=\resLen]{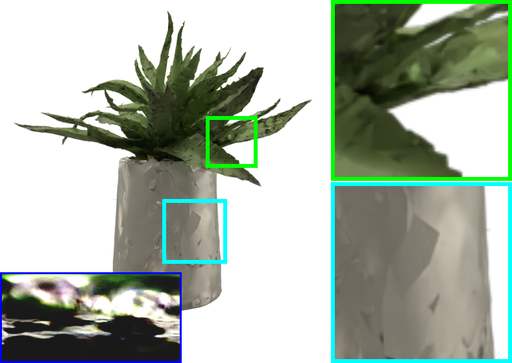} &
    \includegraphics[height=\resLen]{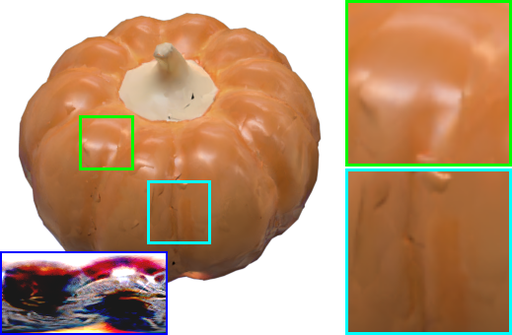} &
    \includegraphics[height=\resLen]{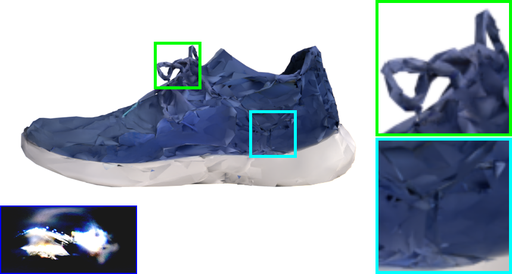} \\
    & {\scriptsize 32.2 / 0.96} & {\scriptsize 27.5 / 0.89} & {\scriptsize 30.5 / 0.97} & {\bf\scriptsize 34.3 / 0.97} & {\scriptsize 27.4 / 0.91} \\
    \vtext{\footnotesize \mii} &
    \includegraphics[height=\resLen]{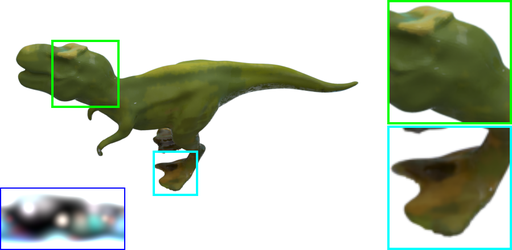} &
    \includegraphics[height=\resLen]{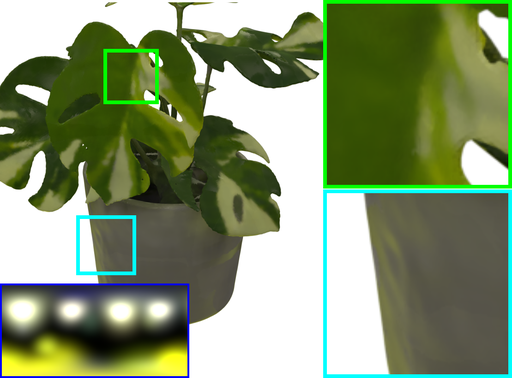} &
    \includegraphics[height=\resLen]{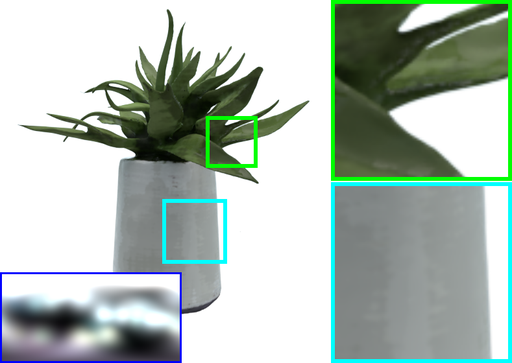} &
    \includegraphics[height=\resLen]{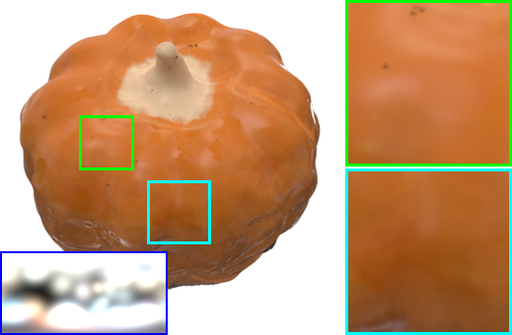} &
    \includegraphics[height=\resLen]{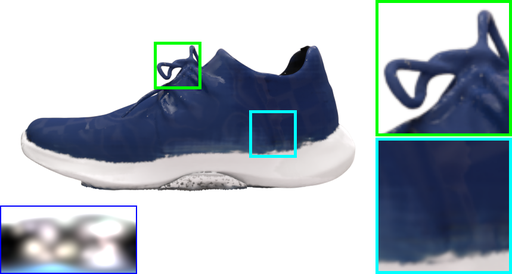} \\
    & {\scriptsize 31.2 / 0.96} & {\scriptsize 27.7 / 0.91} & {\scriptsize 29.2 / 0.97} & {\scriptsize 29.4 / 0.95} & {\scriptsize 26.9 / 0.92}
\end{tabular}
\vspace{-0.3cm}
\captionof{figure}{%
    {\bf Novel-view interpolation on our real dataset.}
    Our technique produces high-fidelity reconstructions with minimal artifacts.
    We report the average PSNR$\uparrow$ and SSIM$\uparrow$ below each image.
}
\label{fig:ours_real_nv}
\end{table*}

%% file: tabs/ours_real_relit.tex
\begin{figure}
\centering
{\small
\begin{tabular}{@{}c@{\hskip 3pt}c@{\hskip 3pt}c@{\hskip 3pt}c@{}}
{\small \mii} & {\small \nvdiff} & {\small Ours} & {\small Groundtruth} \\
\includegraphics[width=0.24\linewidth]{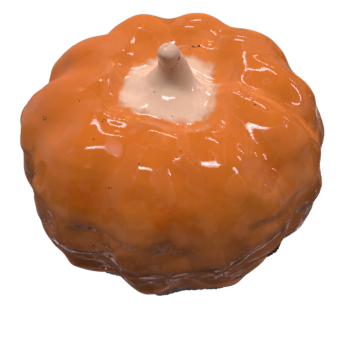} &
\includegraphics[width=0.24\linewidth]{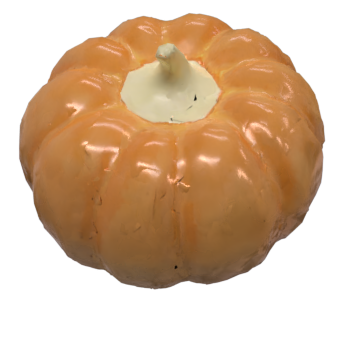} &
\includegraphics[width=0.24\linewidth]{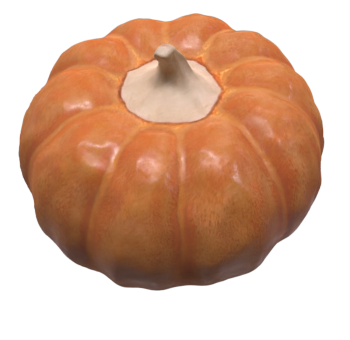} &  
\includegraphics[width=0.24\linewidth]{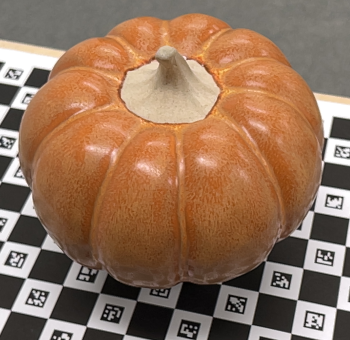}\\
30.1 / 0.94 & 33.0 / 0.96 & {\bf 33.6 / 0.96} & PSNR$\uparrow$ / SSIM$\uparrow$ \\
\end{tabular}
}
\vspace{-0.3cm}
\caption{%
    \textbf{Rerendering of reconstruction results under captured (GT) illumination.}
    We rescale all renderings to match the overall brightness of the GT image.
}
\label{fig:ours_real_relit_capture}

\label{fig:ours_real_relit}
\end{figure}

%% file: tabs/ours_real_relit_novel.tex
\begin{figure}
\setlength{\resLen}{.32\linewidth}
\centering
\begin{tabular}{@{}c@{\hskip 3pt}c@{\hskip 3pt}c@{}}
    {\small \mii} & {\small \nvdiff} & {\small Ours} \\
    \adjincludegraphics[Clip={0} {.25\height} {0} {.0625\height}, width=\resLen]{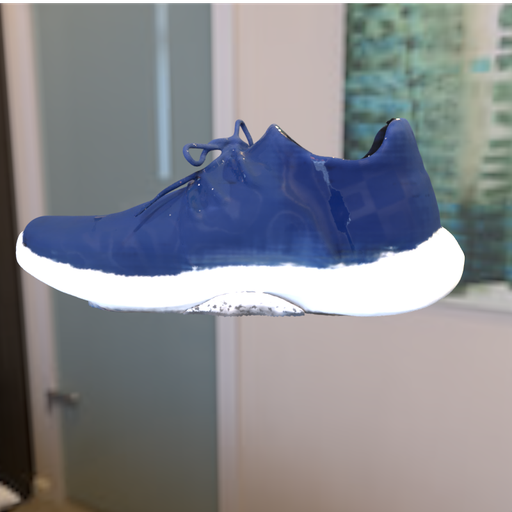} &
    \adjincludegraphics[Clip={0} {.25\height} {0} {.0625\height}, width=\resLen]{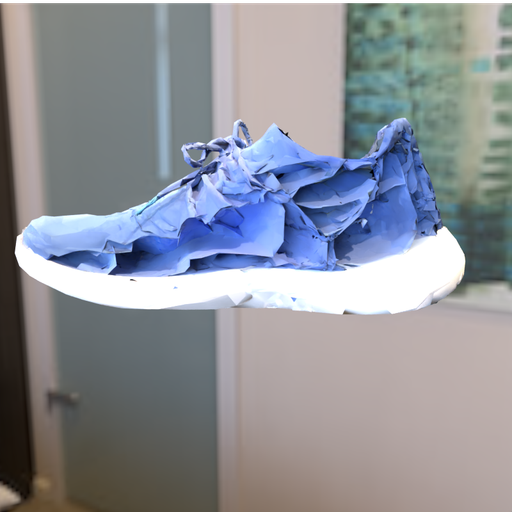} &
    \adjincludegraphics[Clip={0} {.25\height} {0} {.0625\height}, width=\resLen]{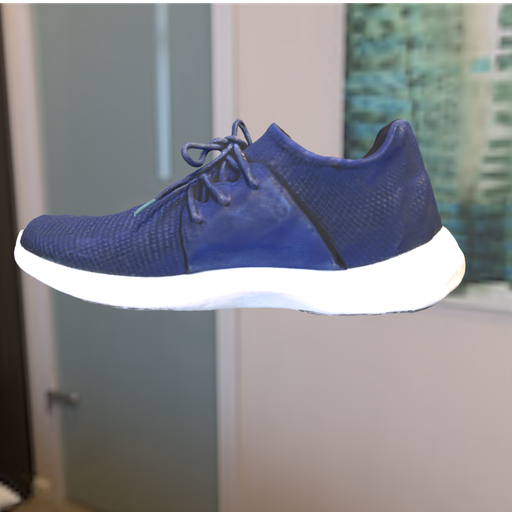}
    \\
    \adjincludegraphics[Clip={0} {.225\height} {0} {.0875\height}, width=\resLen]{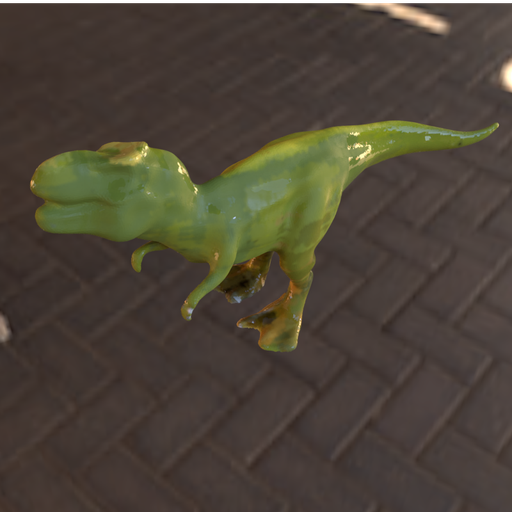} &
    \adjincludegraphics[Clip={0} {.225\height} {0} {.0875\height}, width=\resLen]{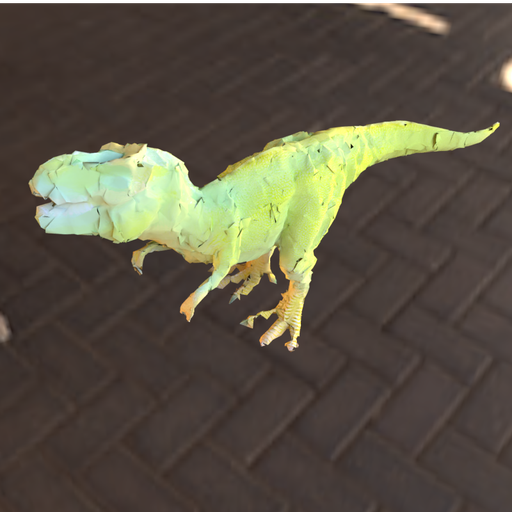} &
    \adjincludegraphics[Clip={0} {.225\height} {0} {.0875\height}, width=\resLen]{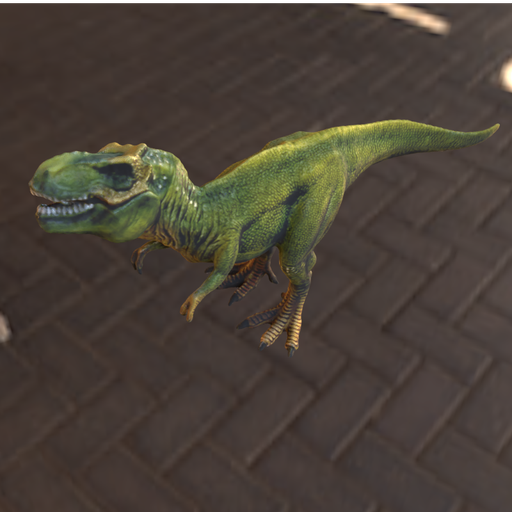}
    \\
    \adjincludegraphics[Clip={0} {.2125\height} {0} {0}, width=\resLen]{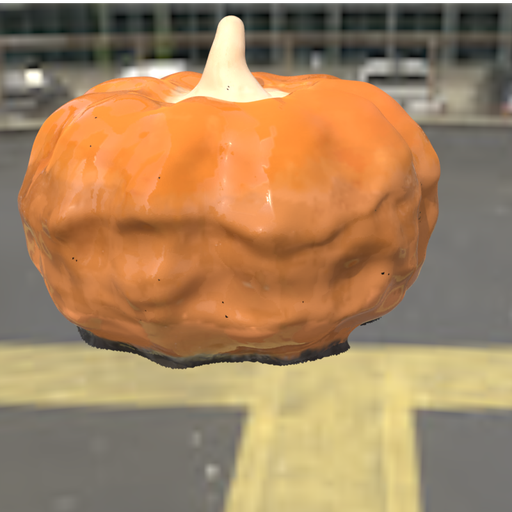} &
    \adjincludegraphics[Clip={0} {.2125\height} {0} {0}, width=\resLen]{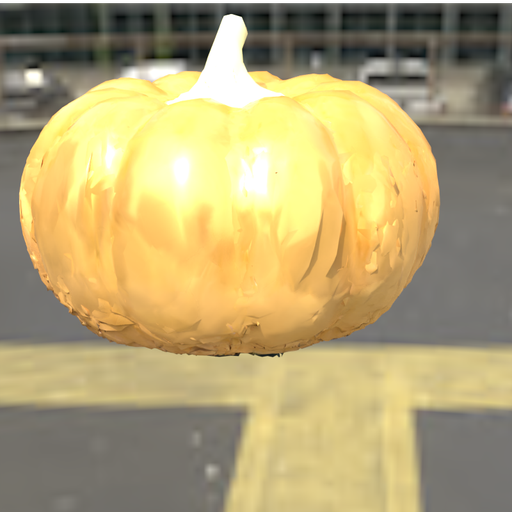} &
    \adjincludegraphics[Clip={0} {.2125\height} {0} {0}, width=\resLen]{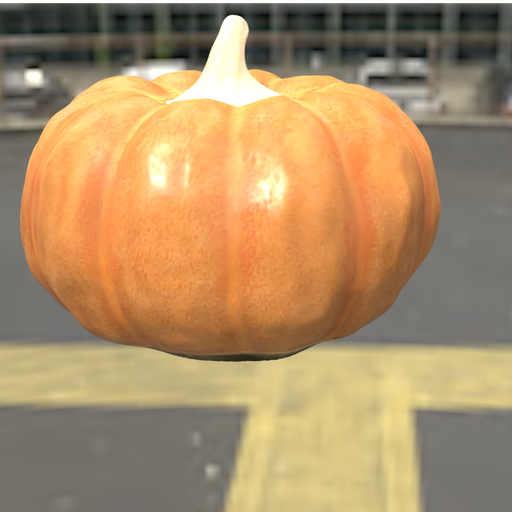}
    \\
    \adjincludegraphics[Clip={0} {0} {0} {.08\height}, width=\resLen]{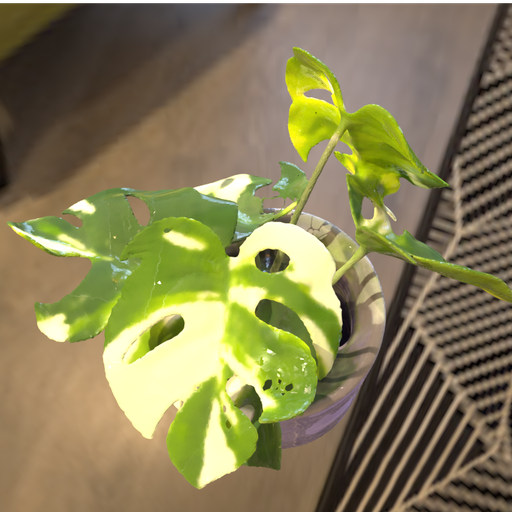} &
    \adjincludegraphics[Clip={0} {0} {0} {.08\height}, width=\resLen]{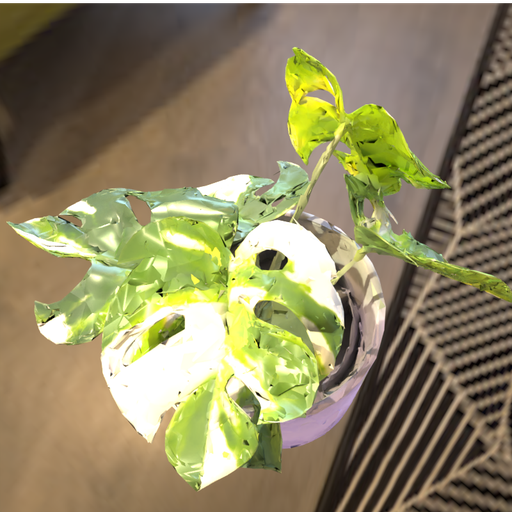} &
    \adjincludegraphics[Clip={0} {0} {0} {.08\height}, width=\resLen]{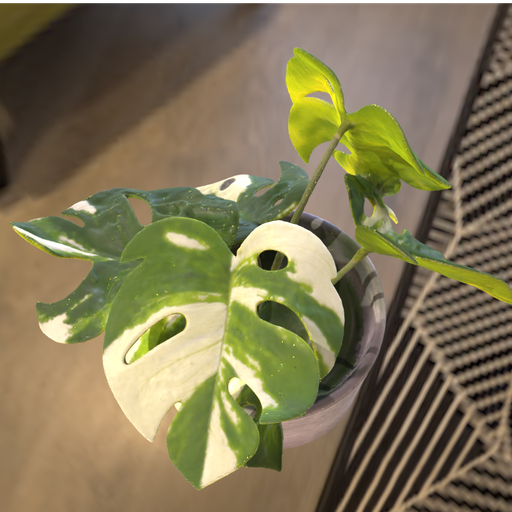}
    \\
    \adjincludegraphics[Clip={.075\width} {.1\height} {.1\width} {.05\height}, width=\resLen]{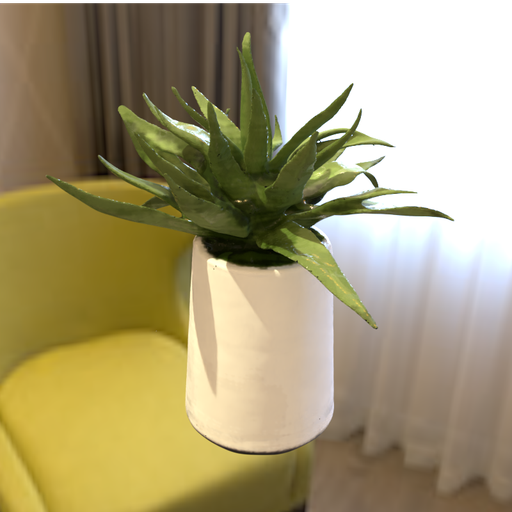} &
    \adjincludegraphics[Clip={.075\width} {.1\height} {.1\width} {.05\height}, width=\resLen]{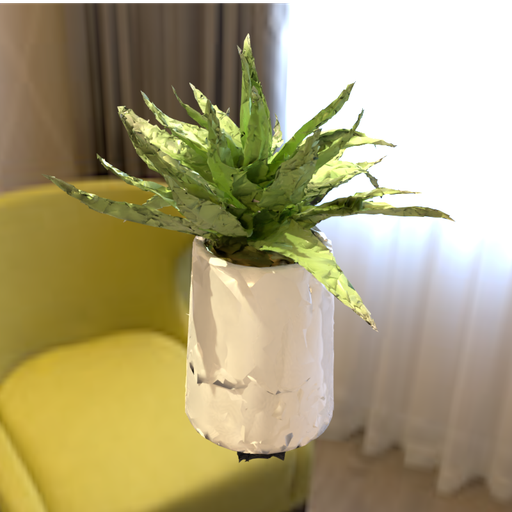} &
    \adjincludegraphics[Clip={.075\width} {.1\height} {.1\width} {.05\height}, width=\resLen]{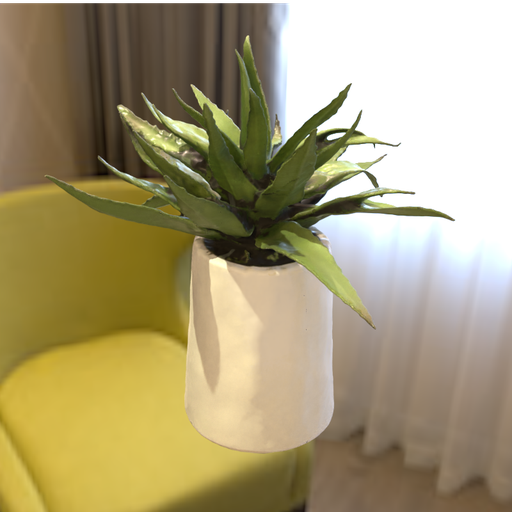}
\end{tabular}
\vspace{-0.3cm}
\caption{%
    {\bf Novel-illumination renderings of the reconstructed models.}
    Our results offer the highest overall quality with minimal artifacts.
}
\label{fig:ours_real_relit_wild}
\end{figure}

%% file: tabs/ablation_pbir.tex
\begin{table*}
    \centering
    {\small
    \begin{tabular}{@{}c@{\hskip 1pt}cc@{\hskip 1pt}c@{}}
    \includegraphics[width=0.24\linewidth]{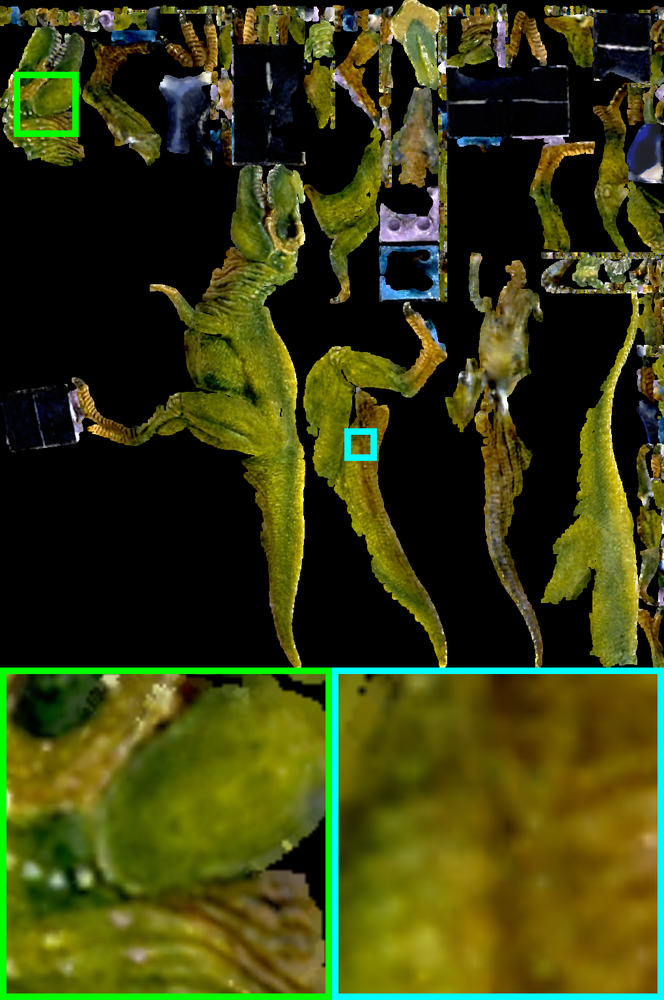} &
    \includegraphics[width=0.24\linewidth]{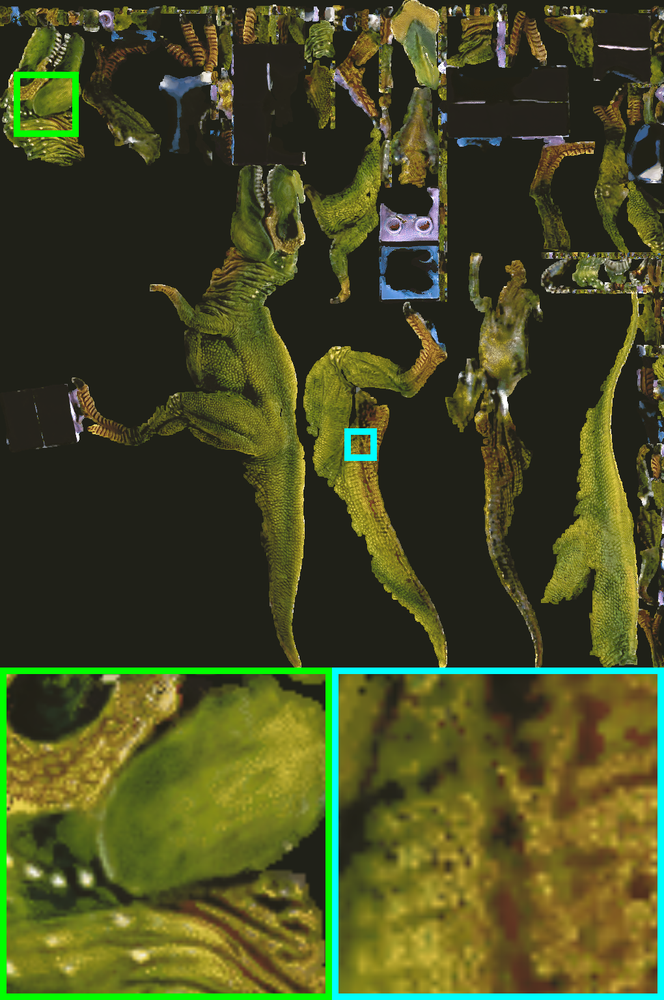} &
    \includegraphics[width=0.24\linewidth]{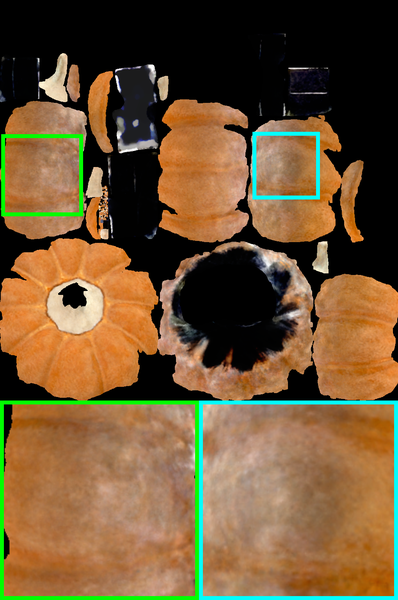} &
    \includegraphics[width=0.24\linewidth]{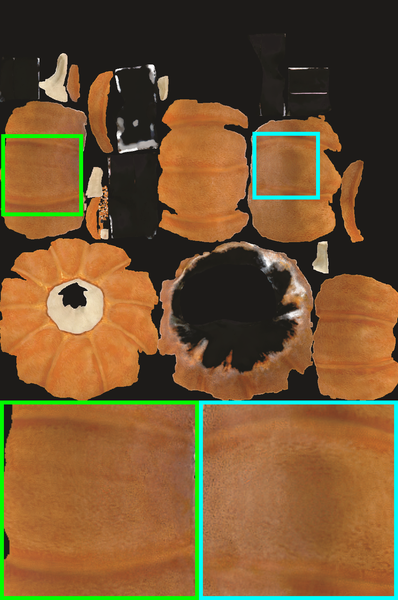} \\
    Distilling & PBIR & Distilling & PBIR
    \end{tabular}
    }
    \vspace{-0.3cm}
    \captionof{figure}{{\bf Reconstructed albedo before and after PBIR.}
    Despite our neural distilling outperform previous arts by a large margin, we still observe blurriness and light baking in the reconstructed materials.
    Our PBIR stage can provide sharper details and remove light baking efficiently.}
    \label{fig:abla_pbir}
\end{table*}

%% file: tabs/dtu_table.tex
\begin{table}
\centering
{\small
\begin{tabular}{@{}l  c@{\hskip 8pt} c@{\hskip 8pt} c@{\hskip 8pt} c@{\hskip 8pt} c@{}}
    \toprule
    Methods   & COLMAP & NeuS & Voxurf & NeuS2 & Our \\
    \midrule
    Runtime$\downarrow$ & 1 hrs & 5.5 hrs & 16 mins & {\bf 5 mins} & {\bf 5 mins} \\
    CD (mm)$\downarrow$ & 1.36 & 0.77 & 0.72 & 0.70 & {\bf 0.66} \\
    \bottomrule
\end{tabular}
}
\vspace{-0.3cm}
\caption{
{\bf Surface reconstruction quality on DTU dataset~\cite{JensenDVTA14}.}
The results are averaged across the 15 scenes.
We include two recent works, Voxurf~\cite{WuWPXTLL22} and NeuS2~\cite{WangHHDTL22}, for reference.
See supp. for results breakdown.
}
\label{tab:dtu}
\end{table}

%% file: tabs/ablation_surface.tex
\begin{table}
\centering
{\small
\begin{tabular}{ @{}l c@{\hskip 8pt} c@{\hskip 8pt} c@{\hskip 8pt} c@{\hskip 8pt} c@{\hskip 8pt} c@{\hskip 8pt} c@{\hskip 8pt} c@{} }
\toprule
$\Llap$           & &            &            & \checkmark &            & \checkmark & \checkmark & \checkmark \\
$\Lrgb$       & &            & \checkmark &            & \checkmark &            & \checkmark & \checkmark \\
{\footnotesize ada. huber} & & \checkmark &            &            & \checkmark & \checkmark &            & \checkmark \\
\midrule
CD (mm)$\downarrow$        & 1.50 & 1.37 & 1.24 & 1.11 & 1.03 & 1.00 & 0.89 & {\bf 0.68} \\
\bottomrule
\end{tabular}
}
\vspace{-0.3cm}
\caption{
{\bf Ablation study of SDF grid regularizations.}
The results are averaged over the 15 objects on DTU dataset.
}
\label{tab:dtu_ablate}
\end{table}

%% file: tabs/ablation_gi.tex
\begin{figure}
    \centering
    \resizebox{\columnwidth}{!}{%
    \begin{tikzpicture}[x=1pt, y=1pt]
        \node[anchor=south west] (img1) at (0pt,0pt) {\includegraphics[width=100pt]{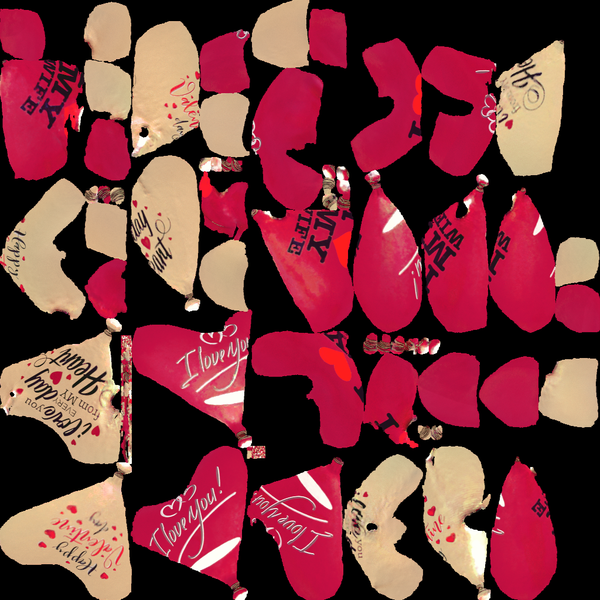}};
        \draw[cyan, line width=1pt] (54pt,4pt) rectangle +(21pt,21pt);
        \draw[green, line width=1pt] (44pt,49pt) rectangle +(21pt,21pt);
        \node[anchor=south west, draw=cyan, inner sep=1pt, line width=1pt] at (103pt,2pt) {\adjincludegraphics[Clip={.4882\width} {0} {.2929\width} {.7812\height}, width=49pt]{figs/ablation/gi/aligned_albedo_w_gi.png}};
        \node[anchor=south west, draw=green, inner sep=1pt, line width=1pt] at (103pt,52pt) {\adjincludegraphics[Clip={.3906\width} {.4394\height} {.3906\width} {.3418\height}, width=49pt]{figs/ablation/gi/aligned_albedo_w_gi.png}};
        \node[anchor=south west] (img2) at (155pt,0) {\includegraphics[width=100pt]{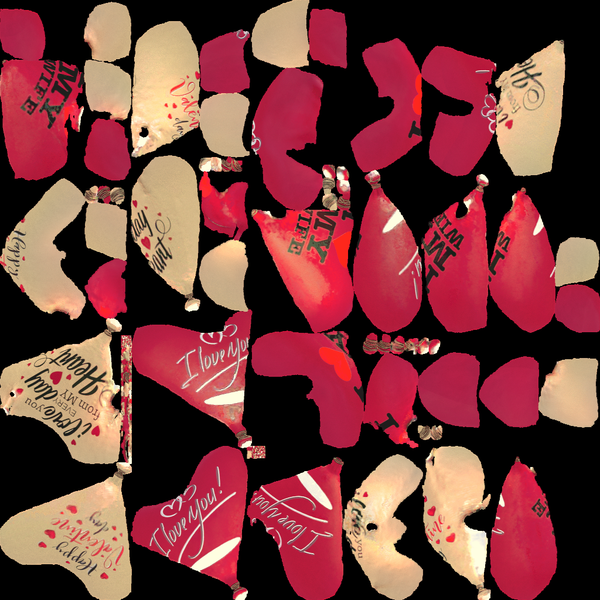}};
        \draw[cyan, line width=1pt] (209pt,4pt) rectangle +(21pt,21pt);
        \draw[green, line width=1pt] (199pt,49pt) rectangle +(21pt,21pt);
        \node[anchor=south west, draw=cyan, inner sep=1pt, line width=1pt] at (258pt,2pt) {\adjincludegraphics[Clip={.4882\width} {0} {.2929\width} {.7812\height}, width=49pt]{figs/ablation/gi/aligned_albedo_wo_gi.png}};
        \node[anchor=south west, draw=green, inner sep=1pt, line width=1pt] at (258pt,52pt) {\adjincludegraphics[Clip={.3906\width} {.4394\height} {.3906\width} {.3418\height}, width=49pt]{figs/ablation/gi/aligned_albedo_wo_gi.png}};
        \node[below] at (img1.south) {w/ GI (27.69 / 0.940)};
        \node[below] at (img2.south) {w/o GI (26.58 / 0.934)};
    \end{tikzpicture}
    }
    \vspace{-0.7cm}
    \caption{\textbf{Reconstructed albedo with/without GI.} We show the impact of modeling global illumination (GI) on material reconstruction. The above images are the reconstructed albedo maps with GI on/off, with their errors listed below (PSNR$\uparrow$ / SSIM$\uparrow$). Without GI, the optimization tends to "bake" the indirect lighting into the albedo map. } 
    \label{fig:ablation_gi}
\end{figure}

%% file: tabs/abla_shape_refine_pig.tex
\begin{figure}
  \centering
  \setlength{\resLen}{0.15\textwidth}
  \addtolength{\tabcolsep}{-4.5pt}
  \small
  \begin{tabular}{ccc}
    \textbf{Before stage 3} & \textbf{After stage 3} & \textbf{GT}
    \\[-8pt]
    \begin{overpic}[width=\resLen]{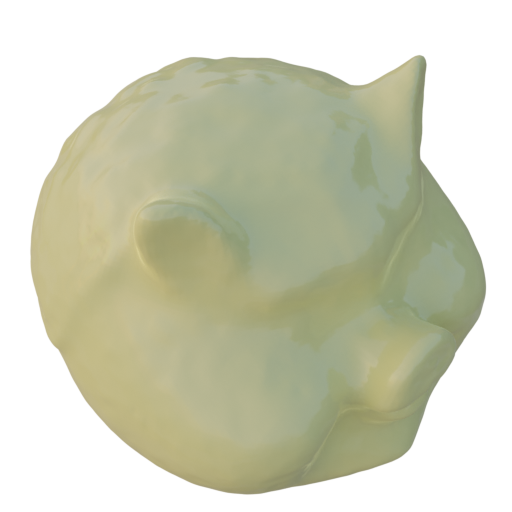}
      \put(2, 82){\color{white} \contour{black}{Training}}
      \put(0, 0){\includegraphics[width=.45\resLen]{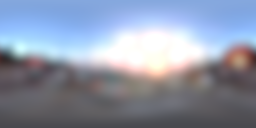}}
    \end{overpic}
    &
    \begin{overpic}[width=\resLen]{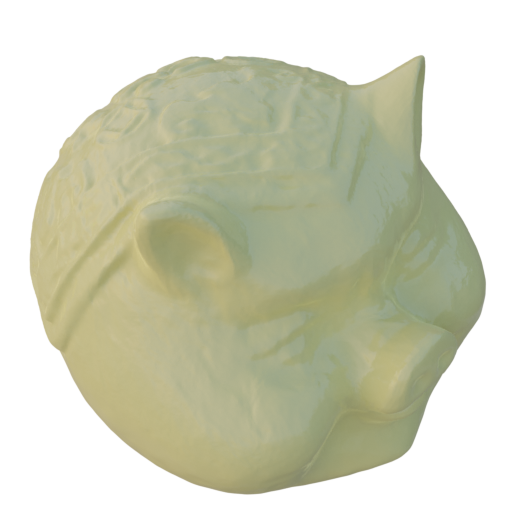}
      \put(0, 0){\includegraphics[width=.45\resLen]{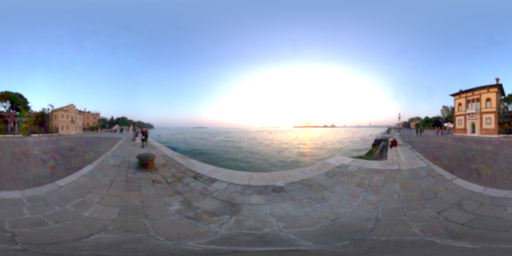}}
    \end{overpic}
    &
    \begin{overpic}[width=\resLen]{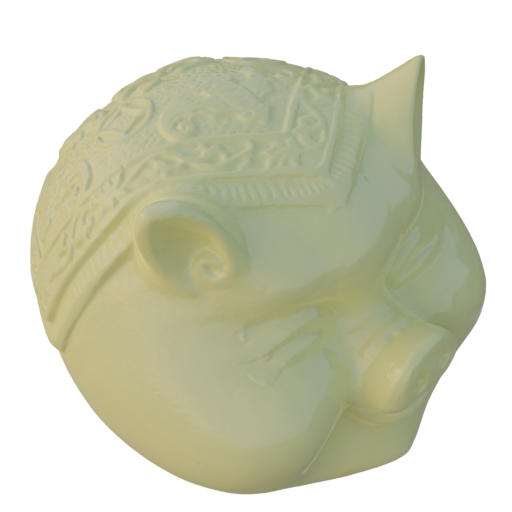}
      \put(0, 0){\includegraphics[width=.45\resLen]{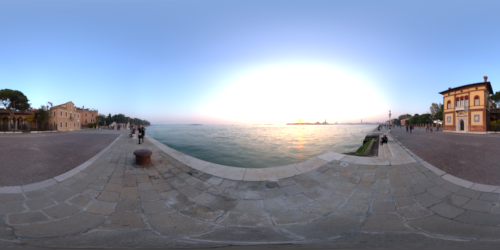}}
    \end{overpic}
    \\[-5pt]
    \begin{overpic}[width=\resLen]{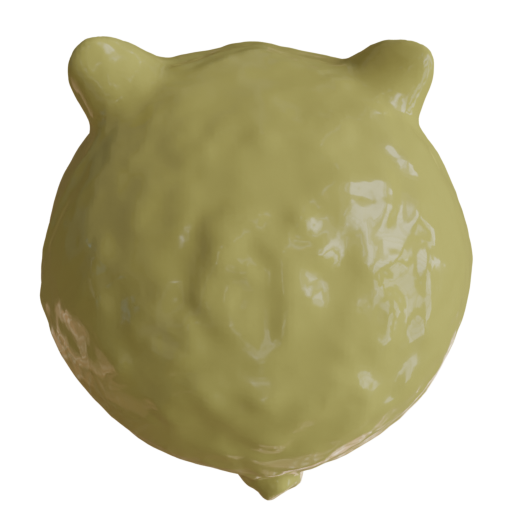}
      \put(2, 82){\color{white} \contour{black}{Re-render}}
      \put(2, 8){\color{white} \contour{black}{(novel light)}}
    \end{overpic}
    &
    \includegraphics[width=\resLen]{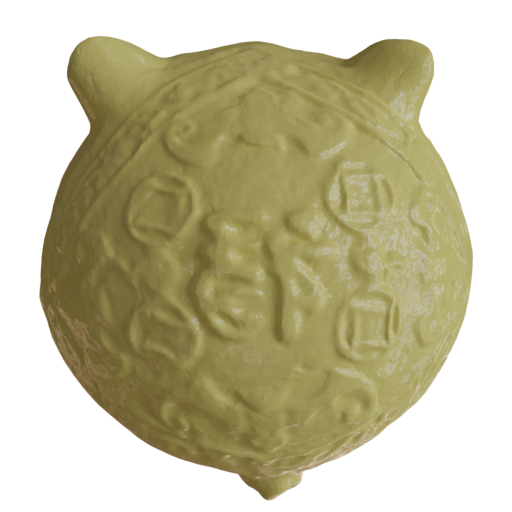} &
    \includegraphics[width=\resLen]{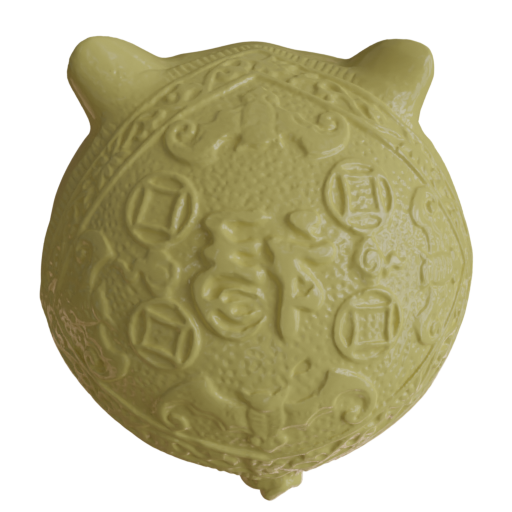}
  \end{tabular}
  \vspace{-1.5em}
  \caption{
    {\bf An\ example showcasing the usefulness of our PBIR shape refinement.}
    Our stage~1 can sometimes lack geometry details while our PBIR in stage~3 can recover these details.
  }
  \label{fig:our_vis_abla_2}
\end{figure}

%% file: conclusion.tex
\section{Discussion and Conclusion}
\label{sec:conclusion}
\paragraph{Limitations.}
Due to the fundamentally under-constrained nature of inverse rendering, our technique is not immunue to ``baking'' artifacts---especially when predictions provided by our neural stages are far from the groundtruth.
Also, although our technique handles global-illumination effects including interreflection, it assumes for opaque materials and does not currently support the reconstruction of transparent or translucent objects.

\paragraph{Conclusion.}
We introduced a new method for reconstructing an object's shape and reflectance under unknown environmental illumination.
Our technique is comprised of three stages: (i) a surface reconstruction stage fitting an implicit surface and a radiance field from input images; (ii) a neural distilling stage decomposing material and lighting from the radiance field; and (iii) a PBIR stage jointly refining the geometry, material, and lighting.
Broad experiments show our effectiveness.

%% file: acknowledgement.tex
\paragraph{Acknowledgement.}
This project has been partially supported by NSF grant 1900927.

%% file: stanford_orb.tex
\section{Results on Stanford ORB dataset}
\label{sec:stanford_orb}
We evaluate Neural-PBIR on the recently released Stanford-ORB dataset~\cite{kuang2023stanfordorb}\footnote{We add Neural-PBIR's result in Dec 2023.}.

\paragraph{Stanford-ORB.}
The dataset scans 14 real-world objects each under 3 different lighting condition, resulting in 42 capturing sequence.
The ground-truth mesh and material are also scanned from studio.
The relative camera poses of an object captured from different lighting are also provided for evaluating the equality of re-lighting quality.

\paragraph{Evaluation protocal.}
We follow official guideline to train our method on each of the 42 scenes separately under benchmark resolution of $512\times 512$.
The same set hyperparameters as the MII dataset are applied for all the Stanford-ORB scenes.
We follow official train-test split and use official script {\scriptsize (\url{https://github.com/StanfordORB/Stanford-ORB})} for authentic evaluation scores.

\begin{figure*}
    \centering
    \begin{overpic}[trim=0 350 0 0, clip,width=.95\linewidth, clip]{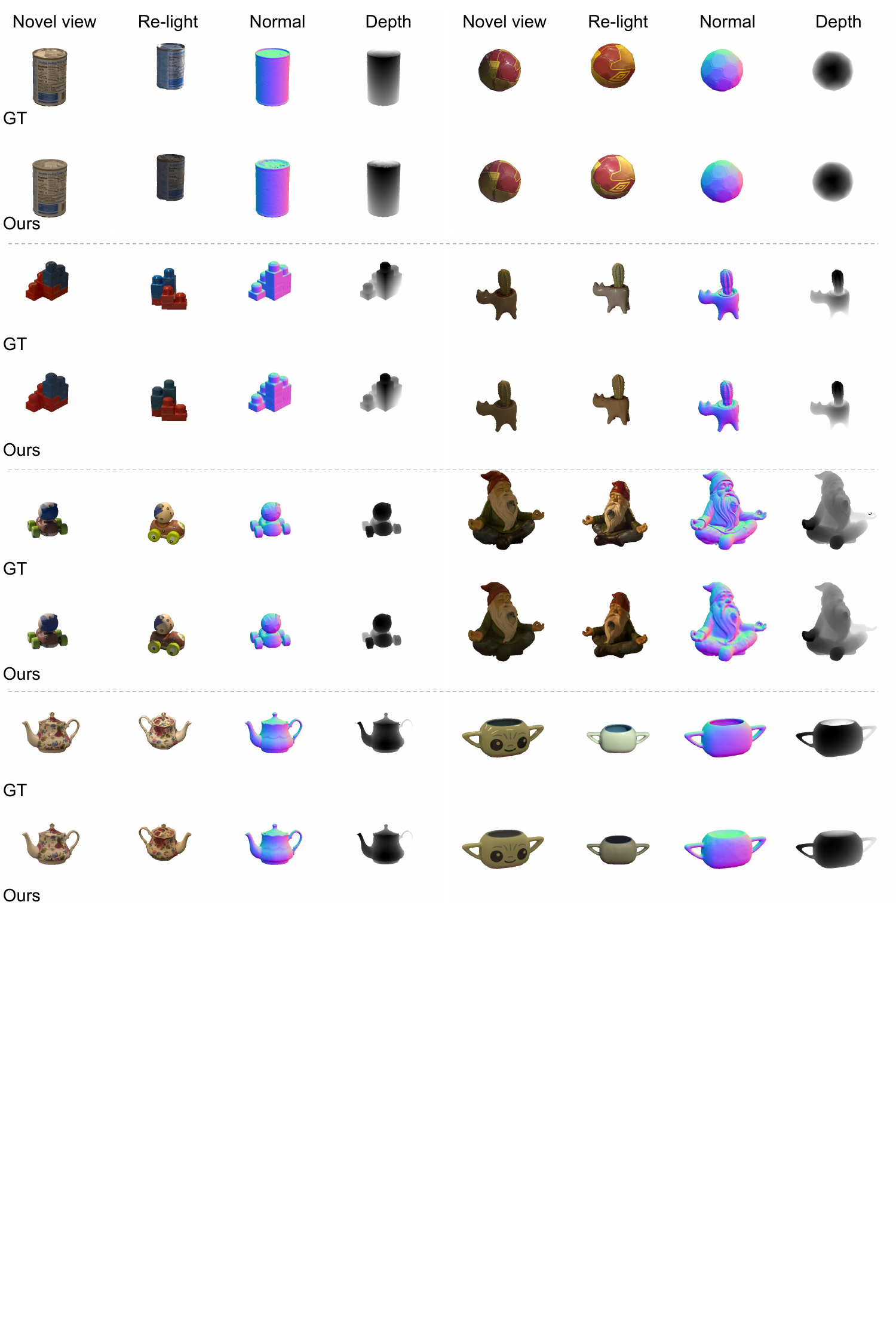}
    \end{overpic}
    \vspace{-1.5em}
    \caption{
    {\bf Qualitative results on Stanford-ORB dataset~\cite{kuang2023stanfordorb}.}
    }
    \label{fig:stanfordorb}
\end{figure*}

\paragraph{Results.}
The quantitative comparison with previous arts is provided in \cref{tab:stanfordorb}, where our Neural-PBIR outperforms previous methods on most metrics.
We show some qualitative results in \cref{fig:stanfordorb}.

\begin{table*}
\centering
{
\resizebox{1\textwidth}{!}{
\begin{tabular}{ @{}l@{\hskip 5pt} @{}c@{} @{}c@{\hskip 5pt} c@{\hskip 5pt} c@{} c@{\hskip 5pt} @{}c@{\hskip 5pt} @{}c@{\hskip 5pt} c@{\hskip 5pt} c@{} c@{\hskip 5pt} @{}c@{\hskip 5pt} @{}c@{\hskip 5pt} c@{\hskip 5pt} c@{} @{}c@{} }
    \toprule
    && \multicolumn{3}{c}{Geometry} && \multicolumn{4}{c}{Novel Scene Relighting} && \multicolumn{4}{c}{Novel View Synthesis} \\
    \cmidrule{3-5}
    \cmidrule{7-10}
    \cmidrule{12-15}
    Method && Depth$\downarrow$ & Normal$\downarrow$ & Shape$\downarrow$ && PSNR-H$\uparrow$ & PSNR-L$\uparrow$ & SSIM$\uparrow$ & LPIPS$\downarrow$ & & PSNR-H$\uparrow$ & PSNR-L$\uparrow$ & SSIM$\uparrow$ & LPIPS$\downarrow$ \\
    \midrule
    PhySG~\cite{ZhangLWBS21} &&
    1.90 & 0.17 & 9.28 && 21.81 & 28.11 & 0.960 & 0.055 && 24.24 & 32.15 & 0.974 & 0.047
    \\
    NVDiffRec~\cite{MunkbergCHES0GF22} &&
    \underline{0.31} & 0.06 & 0.62 && 22.91 & 29.72 & 0.963 & 0.039 && 21.94 & 28.44 & 0.969 & 0.030
    \\
    NeRD~\cite{BossBJBLL21} &&
    1.39 & 0.28 & 13.7 && 23.29 & 29.65 & 0.957 & 0.059 && 25.83 & 32.61 & 0.963 & 0.054
    \\
    NeRFactor~\cite{ZhangSDDFB21} &&
    0.87 & 0.29 & 9.53 && 23.54 & 30.38 & 0.969 & 0.048 && 26.06 & 33.47 & 0.973 & 0.046
    \\
    InvRender~\cite{wu2023invrender} &&
    0.59 & \underline{0.06} & \underline{0.44} && 23.76 & 30.83 & 0.970 & 0.046 && 25.91 & 34.01 & 0.977 & 0.042
    \\
    NVDiffRecMC~\cite{HasselgrenHM22} &&
    0.32 & \textbf{0.04} & 0.51 && \underline{24.43} & \underline{31.60} & \underline{0.972} & \underline{0.036} && \underline{28.03} & \underline{36.40} & \underline{0.982} & \underline{0.028}
    \\
    Neural-PBIR &&
    \textbf{0.30} & \underline{0.06} & \textbf{0.43} && \textbf{26.01} & \textbf{33.26} & \textbf{0.979} & \textbf{0.023} && \textbf{28.83} & \textbf{36.80} & \textbf{0.986} & \textbf{0.019}
    \\
    \bottomrule
\end{tabular}
}
}
\vspace{-0.5em}
\caption{
{\bf Geometry, relighting, and view-interpolation quality on Stanford-ORB dataset~\cite{kuang2023stanfordorb}.}
}
\label{tab:stanfordorb}
\end{table*}

%% file: supp/sec_technical_detail.tex
\section{Technical Details}
\label{sec:tech_details}
In what follows, we elaborate technical details of our Neural-PBIR pipeline's three main stages.
Code will be released upon internal approval for future extension and reproduction.

\input{supp/detail_surface}
\input{supp/detail_neural_distill}
\input{supp/detail_pbir}

%% file: supp/detail_surface.tex
\subsection{Neural Surface Reconstruction}
\label{ssec:surface_detail}
\paragraph{Sharpness term in unbiased volume rendering.}
Following \neus~\cite{WangLLTKW21}, we use a scaled sigmoid $\sigma_s$ function in the SDF for alpha activation:
\begin{equation}
    \label{eqn:alpha_detail}
    \alpha_i = \max\left(0,\ \frac{\sigma_s(\sdf(\bx_i)) - \sigma_s(\sdf(\bx_{i+1}))}{\sigma_s(\sdf(\bx_i))}\right) \,,
\end{equation}
where:
\begin{itemize}
    \item $\sdf$ is the signed-distance function;
    \item $\sigma_s(y) = (1 + \exp(-sy))^{-1}$ with $s > 0$ being the \emph{sharpness term};
    \item $\{ \bx_i = \bo + t_i \bv \}_{i=1}^N$ (with $0 < t_1 < t_2 < \ldots < t_N$) are the $N$ sampled points on the camera ray originated at the camera's location $\bo$ with viewing direction $\bv$.
\end{itemize}
Specifically, we start with $s = 30$ if foreground masks are provided (\eg, for the synthetic and DTU datasets) and $s = 5$ otherwise (\eg for our measured real-world dataset).
In practice, we use a \emph{scheduled} sharpness $s$ (instead of updating $s$ with gradient descent) as we find it more stable.
Then, we update the sharpness $s$ by setting $s \gets \min(s + 0.02,\ 300)$ after each iteration.

\paragraph{Background modeling.}
As stated in Sec.~3.1 of the main paper, %
we use two sets of $\Vsdf$ and $\Vfeat$ grids to model the foreground and the background (via Eq.~(3) in the main paper), respectively.
Specifically, the foreground region is defined as the volume inside a (predetermined) small bounding box.
The background, on the other hand, is the volume inside a much larger bounding box.%
\footnote{In practice, we use background bounding boxes that are $16\times$ as large as the foreground ones.}
Given a camera ray, we categorize the sample points $\{ \bx_i = \bo + t_i \bv \}_{i=1}^N$ along the ray as foreground or background and then evaluate signed distance~$\sdf(\bx_i)$ and radiance~$\Lo(\bx_i, -\bv)$ for each $\bx_i$ using the corresponding grids.

Thanks to the background scene volume, our method can work without external mask supervision (e.g., our own real-world dataset).

\paragraph{Points sampling on rays.}
When sampling 3D points $\{ \bx_i = \bo + t_i \bv \}_{i=1}^N$ along a camera ray, we use $t_i = i \Delta t$ with $\Delta t$ being half the size of a grid voxel for all $i = 1, 2, \ldots, N$.

\paragraph{Coarse-to-fine optimization.}
For better efficiency and more coherent results, when optimizing the $\Vsdf$ and $\Vfeat$ grids, we leverage a coarse-to-fine scheme by
doubling the number of voxels every 1k iterations for the first 10k iterations.
The final voxel resolutions are $300^3$ for the foreground grids (which contain the object of interest) and $160^3$ for the background ones.

\paragraph{Optimization details.}
We optimizing the $\Vsdf$ and $\Vfeat$ grids for the foreground and the background jointly using the \textsf{Adam}~\cite{KingmaB14} method with $\beta = (0.9, 0.99)$ and $\epsilon = 10^{-12}$ in 20k iterations.
When computing the loss, we use weights $\Wlap=10^{-8}$ and $\Wrgb=0.01$.
Also, when using the running means to update the threshold $t$ in the adaptive Huber loss, we set the momentum to $0.99$ and clamp $t$ to a minimum of $0.01$.

When training the SDF grids $\Vsdf$, we use an initial learning rate of $0.01$ that then decays to $0.001$ at 10k iterations.
When training the outgoing radiance field $\Lo$, we use a learning rate of $0.001$ for the MLPs and $0.1$ for the feature grids $\Vfeat$.

%% file: supp/detail_neural_distill.tex
\subsection{Neural Distillation of Material and Lighting}
\label{ssec:neural_distill_detail}

\paragraph{Initialization.}
We initialize the roughnesses to $\Mr[\vind] = 0.25$ for each vertex $\vind$.
For per-vertex albedo, we initialize $\Ma[\vind]$ to the median of the outgoing radiance from the teacher model $\Lo$:
\begin{multline}
    \Ma[\vind] = \operatorname{Median} \big\{ \Lo(\bx[\vind], \bomo) \mid\\
    \bomo \in \wiset,\ (\bomo\cdot\Mn[\vind])>0 \big\}~,
\end{multline}
where $\bx[\vind]$ and $\Mn[\vind]$ indicate, respectively, the position and the normal of vertex $\vind$, and $\wiset$ is the predetermined set of outgoing directions.

\paragraph{Fresnel term.}
In addition to albedo and roughness, we also need the Fresnel term $F_0$~\cite{karis2013real} to model specular reflection.
Following \mii, we assume the object to be reconstructed is dielectric and make $F_0$ constant.
We set $F_0 = 0.02$ for all synthetic data since it is used by \mii's open-source implementation, and $F_0 = 0.04$ for real-world data since it is %
the industrial standard.

\begin{figure}
    \centering
    \includegraphics[width=.495\linewidth]{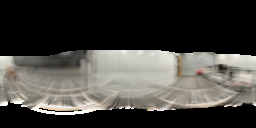}
    \hfill
    \includegraphics[width=.495\linewidth]{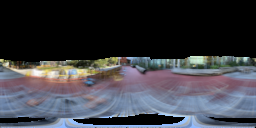}
    \caption{{\bf Examples of the averaged background observation.}
    The black region indicates missing observation.}
    \label{fig:avg_bg_examples}
\end{figure}

\paragraph{Averaged background constraint.}
Recap that we regularize our SG-based illumination $\LenvSG$ to be similar to the averaged background observation $(\LenvSG)'$.
We now detail how the latter is obtained.

First, we gather all ``background'' training pixels (toward which the camera rays miss our reconstructed mesh).
Then, we compute $(\LenvSG)'$ as an environment map under the latitude-and-longitude representation as follows.
For each background pixel with intensity $I$ and viewing direction $\bv$, we set the value of the corresponding pixel $j$ in the environment map $(\LenvSG)'$---based on latitude and longitude coordinates of $\bv$---as $(\LenvSG)'[j] = I$.
When multiple pixels (from different camera locations) contribute to one pixel $j$ of $(\LenvSG)'$, we set $(\LenvSG)'[j]$ using the average intensity of all such pixels.

We show some examples of the averaged background observations $(\LenvSG)'$ in \cref{fig:avg_bg_examples}.
We only compute the regularization loss for the observed viewing directions.

\paragraph{Optimization details.}
To optimize per-vertex appearance parameters, we use the \textsf{Adam} method with $\beta = (0.9, 0.999)$ and $\epsilon = 10^{-8}$ in 2k iterations.
When computing losses, we use the weights $\Wdtreg=0.1$ and $\Wdtbg=10$. %
We use a learning rate $0.01$ for per-vertex attributes and $0.001$ for the spherical Gaussian (SG) parameters (representing the illumination $\LenvSG$).

%% file: supp/detail_pbir.tex
\subsection{Physics-Based Inverse Rendering}
\label{ssec:pbir_detail}

\paragraph{Optimization details.} Initialized using the mesh $M_0$ predicted by the surface reconstruction stage as well as albedo/roughness maps $\Ta^{(0)}$, $\Tr^{(0)}$ (for surface reflectance) and SG-based illumination $\LenvSG$ produced by the neural distillation stage, our physics-based inverse rendering (PBIR) stage involves the following three steps:
\begin{enumerate}
    \item We jointly optimize (using 1k iterations) the albedo/roughness maps $\Ta$, $\Tr$ and the SG parameters $\LenvSG$ while keeping the mesh geometry fixed.
    \item We first pixelize the SG-based $\LenvSG$ into an environment map $\Lenv$ and then perform joint per-pixel optimizations (using 1k iterations) for the albedo, roughness, and environment maps $\Ta$, $\Tr$, and $\Lenv$.
    \item We jointly optimize (using 500 iterations) all maps and the mesh geometry (per-vertex).
\end{enumerate}
In practice, when optimizing albedo and roughness maps $\Ta$ and $\Tr$ in all three steps, we use the \textsf{Adam} optimizer with $\beta = (0.9, 0.999)$, $\epsilon = 10^{-8}$, and the learning rates $10^{-2}$ for $\Ta$ and $5 \times 10^{-3}$ for $\Tr$.
When computing losses, we use  $\Wmask \approx 10$ and $\Wreg \approx 0.1$ (which we slightly adjust for each example). %

Additionally, in the first step, we use the \textsf{Adam} optimizer~\cite{KingmaB14} for the SG parameters with $\beta = (0.9, 0.999)$, $\epsilon = 10^{-8}$, and learning rates around 0.001 (which we slightly adjust per example).
In the second step, to suppress the impact of Monte Carlo noises during environment map optimization, we utilize the \textsf{AdamUniform} optimizer~\cite{Nicolet2021Large} with $\lambda = 1$ and a learning rate of 0.01.
In the last step, when optimizing the mesh geometry, we again use the \textsf{AdamUniform} optimizer with $\lambda = 100$. %

%% file: supp/sec_additional_results.tex
\section{Additional Results and Evaluations}
\label{sec:addtional_results}

\subsection{Additional Results}
\label{ssec:addtl_res}
\input{supp/eval_relit_video}
\input{supp/eval_real_outdoor}
\input{supp/eval_mii}

\input{supp/eval_ours_synthetic}

\subsection{Additional Evaluations}
\label{ssec:addtl_eval}
\input{supp/eval_surface}

%% file: supp/eval_relit_video.tex
\paragraph{Video for view synthesis and relighting.}
Since results of novel-view synthesis and relighting are best viewed animated, we encourage readers to see our supplementary video (\href{run:video.mp4}{video.mp4}) for a more convincing comparison on our five real-world objects.

Similar to the results shown in Fig.~6 of the main paper, our method significantly outperforms \nvdiff~\cite{HasselgrenHM22} and \mii~\cite{ZhangSHFJZ22}.
\nvdiff's geometry and material reconstructions %
contain heavy artifacts.
Despite \nvdiff showing better novel-view results than \mii (main paper's Fig.~4), the artifacts become visually prominent under novel illuminations as can be seen in the video and main paper's Fig.~6.
\mii offers better overall albedo than \nvdiff but %
suffer from over-blurring in both geometry and material reconstructions.
Overall, our results show significant better quality in both geometry and material.

%% file: supp/eval_real_outdoor.tex
\paragraph{Outdoor illumination.}
The five real-world objects presented the main paper are captured under indoor lighting.
In \cref{fig:ours_real_nv_outdoor}, we showcase the results of two of these objects re-captured under outdoor illumination.
Same as the results under indoor lighting, our reconstructions are more detailed, allowing their rerenderings (under novel views) to achieve better PSNR and SSIM.

\input{supp/figure_ours_outdoor}

%% file: supp/figure_ours_outdoor.tex
\begin{figure*}
\setlength{\resLen}{1.65in}
\centering
\begin{tabular}{@{}c@{\hskip 2pt}c@{\hskip 2pt}c@{\hskip 2pt}c@{\hskip 2pt}c@{}}
& {\footnotesize \mii} & {\footnotesize \nvdiff} & {\footnotesize Ours} & {\footnotesize GT}
\\
    \vtext{\it plantpot outdoor} &
    \includegraphics[width=\resLen]{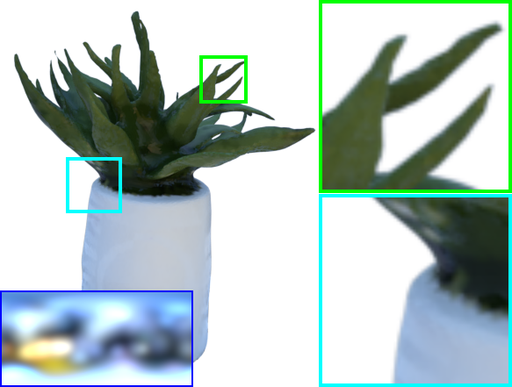} &
    \includegraphics[width=\resLen]{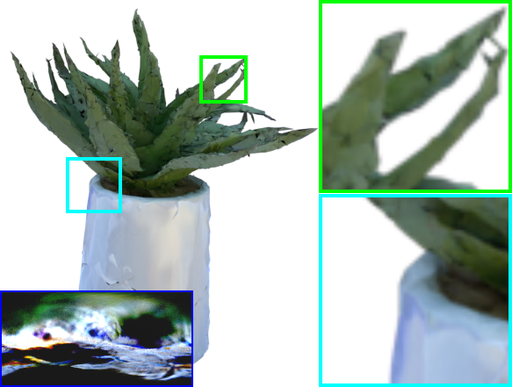} &
    \includegraphics[width=\resLen]{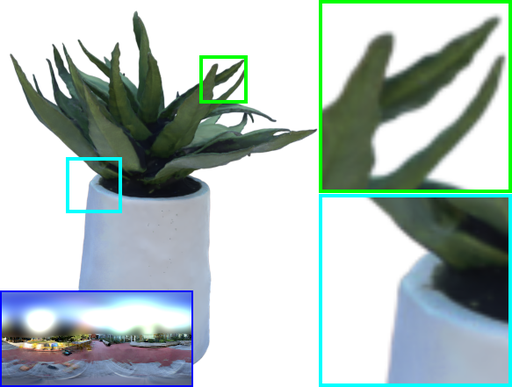} &
    \includegraphics[width=\resLen]{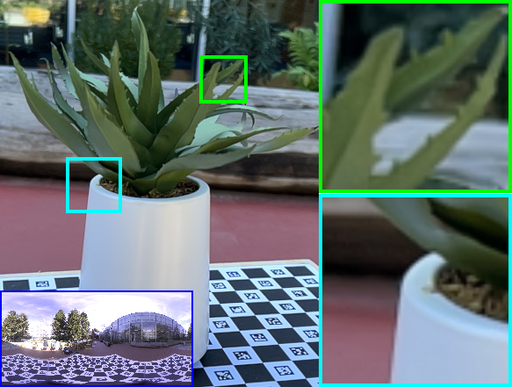}
    \\
    &
    {\scriptsize 30.2 / 0.97} &
    {\scriptsize 31.7 / 0.97} &
    {\bf\scriptsize 33.0 / 0.98} &
    \\
    \vtext{\it pumpkin outdoor} &
    \includegraphics[width=\resLen]{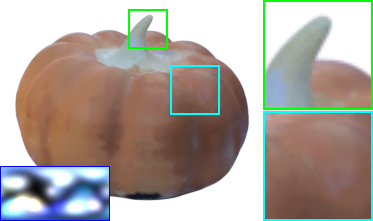} &
    \includegraphics[width=\resLen]{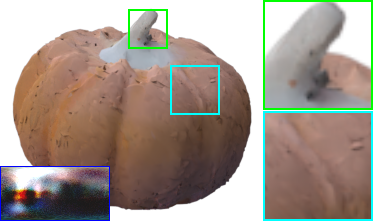} &
    \includegraphics[width=\resLen]{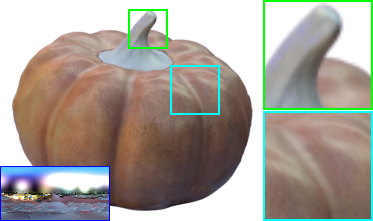} &
    \includegraphics[width=\resLen]{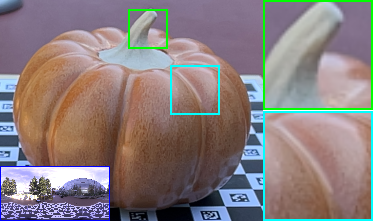}
    \\
    &
    {\scriptsize 32.9 / 0.98} &
    {\scriptsize 33.1 / 0.97} &
    {\bf\scriptsize 34.7 / 0.98} &
\end{tabular}
\caption{%
    {\bf Novel-view interpolation on our additional two real outdoor data.}
    We report the average PSNR$\uparrow$ and SSIM$\uparrow$ below each image.
    The results show that our method achieves good quality and outperforms previous arts under outdoor lighting as well.
}
\label{fig:ours_real_nv_outdoor}
\end{figure*}

%% file: supp/eval_mii.tex
\paragraph{Synthetic \mii dataset.}
The authors of \mii have kindly shared their rendered results for us to compare.
As their evaluation scripts are unavailable, we use our own implementation for all the quantitative results.
Due to the different implementation of the evaluation metrics, \mii's quantitative results presented in our main paper differ slightly from those reported in their paper.

In \cref{fig:mii_vis_air_balloons,fig:mii_vis_chair,fig:mii_vis_hotdog,fig:mii_vis_jugs}, we show more qualitative results on the synthetic \mii dataset.
Overall, our method offers more detailed albedo reconstructions than the baseline methods.
On the other hand, none of the methods performs well on roughness estimation---likely due to the lack of robust priors.
The qualitative results are consistent with the quantitative comparison in Tab.~1 of the main paper.

\input{supp/figure_mii_more}

%% file: supp/figure_mii_more.tex
\newcommand{\miiimage}[2]{\adjincludegraphics[#1]{figs/mii_synthetic/#2}}
\setlength{\resLen}{.77in}

\begin{figure*}
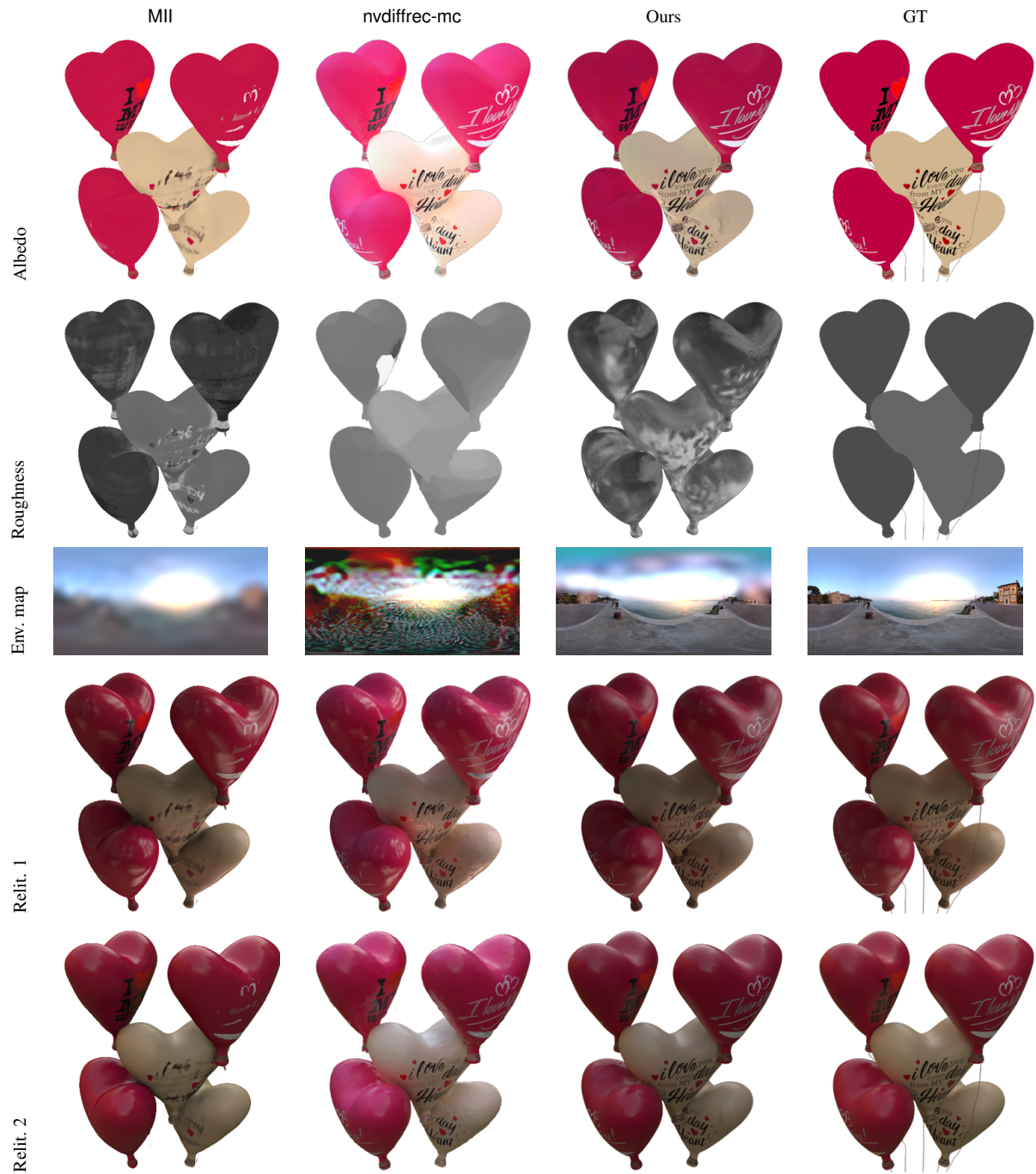

    \setlength{\resLen}{1.45in}
    \addtolength{\tabcolsep}{-3pt}
    \centering
    \begin{tabular}{ccccc}
    & {\footnotesize \mii} & {\footnotesize \nvdiff} & {\footnotesize Ours} & {\footnotesize GT}
    \\
    \vtext{{\footnotesize Albedo}} & 
    \miiimage{trim={{0.15\width} {0.1375\height} {0.15\width} {0.125\height}},clip,width=\resLen}{mii/air_balloons/im_0145_albedo.png} & 
    \miiimage{trim={{0.15\width} {0.1375\height} {0.15\width} {0.125\height}},clip,width=\resLen}{nvdiffrec/air_balloons/im_0145_albedo.png} & 
    \miiimage{trim={{0.15\width} {0.1375\height} {0.15\width} {0.125\height}},clip,width=\resLen}{ours/air_balloons/im_0145_albedo.png} & 
    \miiimage{trim={{0.15\width} {0.1375\height} {0.15\width} {0.125\height}},clip,width=\resLen}{GT/air_balloons/im_0145_albedo.png}
    \\
    \vtext{{\footnotesize Roughness}} &
    \miiimage{trim={{0.15\width} {0.1375\height} {0.15\width} {0.125\height}},clip,width=\resLen}{mii/air_balloons/im_0145_roughness.png} & 
    \miiimage{trim={{0.15\width} {0.1375\height} {0.15\width} {0.125\height}},clip,width=\resLen}{nvdiffrec/air_balloons/im_0145_roughness.png} & 
    \miiimage{trim={{0.15\width} {0.1375\height} {0.15\width} {0.125\height}},clip,width=\resLen}{ours/air_balloons/im_0145_roughness.png} & 
    \miiimage{trim={{0.15\width} {0.1375\height} {0.15\width} {0.125\height}},clip,width=\resLen}{GT/air_balloons/im_0145_roughness.png}
    \\
    \vtext{{\footnotesize Env. map}} & 
    \miiimage{width=.9\resLen}{mii/air_balloons/envmap.png} & 
    \miiimage{width=.9\resLen}{nvdiffrec/air_balloons/envmap.png} & 
    \miiimage{width=.9\resLen}{ours/air_balloons/envmap.png} &
    \miiimage{width=.9\resLen}{GT/air_balloons/envmap.png}
    \\
    \vtext{\footnotesize Relit. 1} & 
    \miiimage{trim={{0.15\width} {0.1375\height} {0.15\width} {0.125\height}},clip,width=\resLen}{mii/air_balloons/im_0145_envmap6.png} & 
    \miiimage{trim={{0.15\width} {0.1375\height} {0.15\width} {0.125\height}},clip,width=\resLen}{nvdiffrec/air_balloons/im_0145_envmap6.png} & 
    \miiimage{trim={{0.15\width} {0.1375\height} {0.15\width} {0.125\height}},clip,width=\resLen}{ours/air_balloons/im_0145_envmap6.png} & 
    \miiimage{trim={{0.15\width} {0.1375\height} {0.15\width} {0.125\height}},clip,width=\resLen}{GT/air_balloons/im_0145_envmap6.png}
    \\
    \vtext{\footnotesize Relit. 2} & 
    \miiimage{trim={{0.15\width} {0.1375\height} {0.15\width} {0.125\height}},clip,width=\resLen}{mii/air_balloons/im_0145_envmap12.png} & 
    \miiimage{trim={{0.15\width} {0.1375\height} {0.15\width} {0.125\height}},clip,width=\resLen}{nvdiffrec/air_balloons/im_0145_envmap12.png} & 
    \miiimage{trim={{0.15\width} {0.1375\height} {0.15\width} {0.125\height}},clip,width=\resLen}{ours/air_balloons/im_0145_envmap12.png} & 
    \miiimage{trim={{0.15\width} {0.1375\height} {0.15\width} {0.125\height}},clip,width=\resLen}{GT/air_balloons/im_0145_envmap12.png}
    \end{tabular}
    \caption{{\bf Qualitative comparisons of {\it air\_balloons} from the \mii dataset.}}
    \label{fig:mii_vis_air_balloons}
\end{figure*}

\begin{figure*}
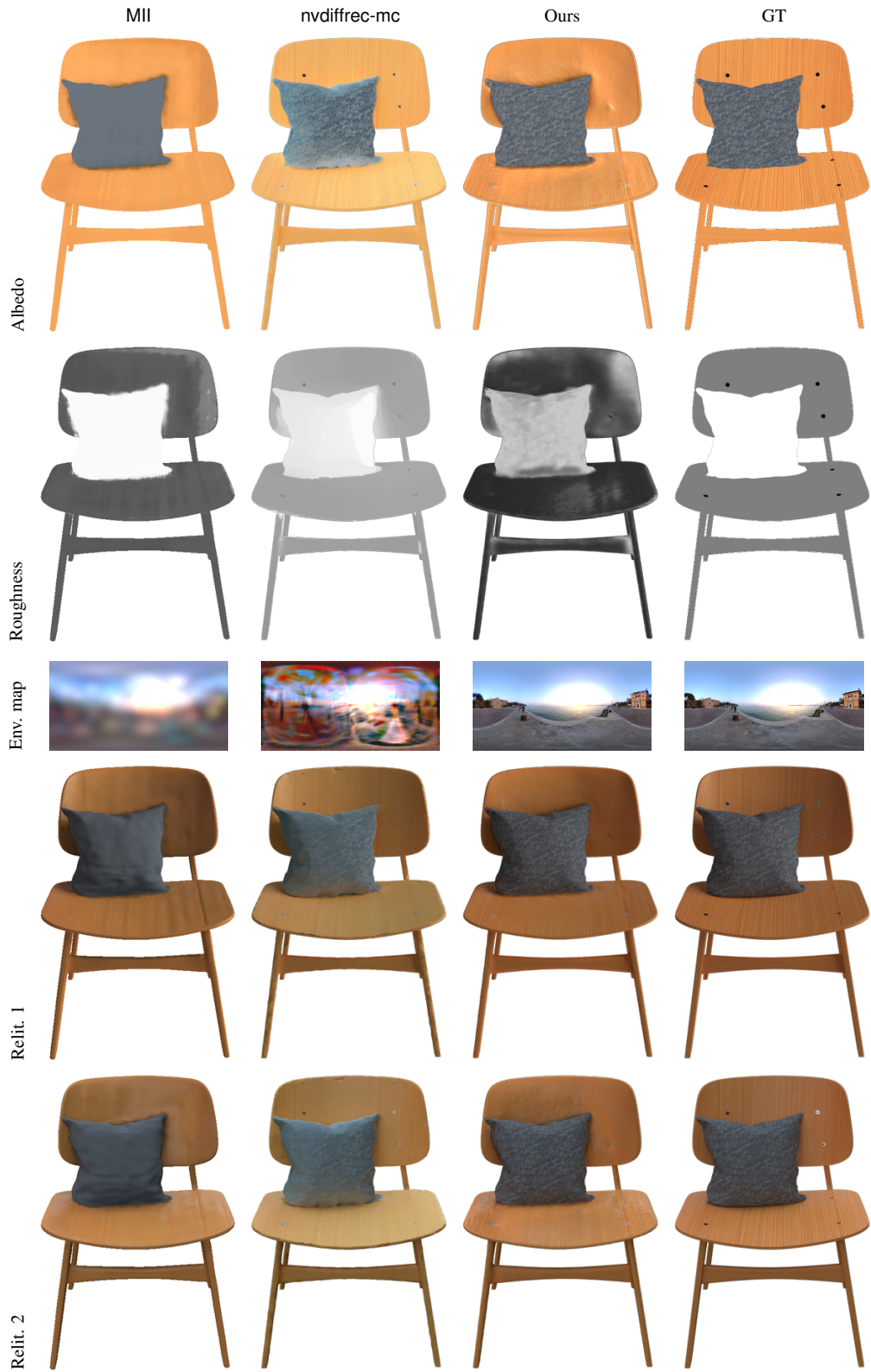

    \setlength{\resLen}{1.2in}
    \addtolength{\tabcolsep}{-3pt}
    \centering
    \begin{tabular}{ccccc}
    & {\footnotesize \mii} & {\footnotesize \nvdiff} & {\footnotesize Ours} & {\footnotesize GT}
    \\
    \vtext{{\footnotesize Albedo}} & 
    \miiimage{trim={{0.225\width} {0.1625\height} {0.25\width} {0.0375\height}},clip,width=\resLen}{mii/chair/im_0051_albedo.png} & 
    \miiimage{trim={{0.225\width} {0.1625\height} {0.25\width} {0.0375\height}},clip,width=\resLen}{nvdiffrec/chair/im_0051_albedo.png} & 
    \miiimage{trim={{0.225\width} {0.1625\height} {0.25\width} {0.0375\height}},clip,width=\resLen}{ours/chair/im_0051_albedo.png} & 
    \miiimage{trim={{0.225\width} {0.1625\height} {0.25\width} {0.0375\height}},clip,width=\resLen}{GT/chair/im_0051_albedo.png}
    \\
    \vtext{{\footnotesize Roughness}} &
    \miiimage{trim={{0.225\width} {0.1625\height} {0.25\width} {0.0375\height}},clip,width=\resLen}{mii/chair/im_0051_roughness.png} & 
    \miiimage{trim={{0.225\width} {0.1625\height} {0.25\width} {0.0375\height}},clip,width=\resLen}{nvdiffrec/chair/im_0051_roughness.png} & 
    \miiimage{trim={{0.225\width} {0.1625\height} {0.25\width} {0.0375\height}},clip,width=\resLen}{ours/chair/im_0051_roughness.png} & 
    \miiimage{trim={{0.225\width} {0.1625\height} {0.25\width} {0.0375\height}},clip,width=\resLen}{GT/chair/im_0051_roughness.png}
    \\[5pt]
    \vtext{{\footnotesize Env. map}} & 
    \miiimage{width=.9\resLen}{mii/chair/envmap.png} & 
    \miiimage{width=.9\resLen}{nvdiffrec/chair/envmap.png} & 
    \miiimage{width=.9\resLen}{ours/chair/envmap.png} &
    \miiimage{width=.9\resLen}{GT/chair/envmap.png}
    \\
    \vtext{\footnotesize Relit. 1} & 
    \miiimage{trim={{0.225\width} {0.1625\height} {0.25\width} {0.0375\height}},clip,width=\resLen}{mii/chair/im_0051_envmap6.png} & 
    \miiimage{trim={{0.225\width} {0.1625\height} {0.25\width} {0.0375\height}},clip,width=\resLen}{nvdiffrec/chair/im_0051_envmap6.png} & 
    \miiimage{trim={{0.225\width} {0.1625\height} {0.25\width} {0.0375\height}},clip,width=\resLen}{ours/chair/im_0051_envmap6.png} & 
    \miiimage{trim={{0.225\width} {0.1625\height} {0.25\width} {0.0375\height}},clip,width=\resLen}{GT/chair/im_0051_envmap6.png}
    \\
    \vtext{\footnotesize Relit. 2} & 
    \miiimage{trim={{0.225\width} {0.1625\height} {0.25\width} {0.0375\height}},clip,width=\resLen}{mii/chair/im_0051_envmap12.png} & 
    \miiimage{trim={{0.225\width} {0.1625\height} {0.25\width} {0.0375\height}},clip,width=\resLen}{nvdiffrec/chair/im_0051_envmap12.png} & 
    \miiimage{trim={{0.225\width} {0.1625\height} {0.25\width} {0.0375\height}},clip,width=\resLen}{ours/chair/im_0051_envmap12.png} & 
    \miiimage{trim={{0.225\width} {0.1625\height} {0.25\width} {0.0375\height}},clip,width=\resLen}{GT/chair/im_0051_envmap12.png}
    \end{tabular}
    \caption{{\bf Qualitative comparisons of {\it chair} from the \mii dataset.}
    }
    \label{fig:mii_vis_chair}
\end{figure*}

\begin{figure*}
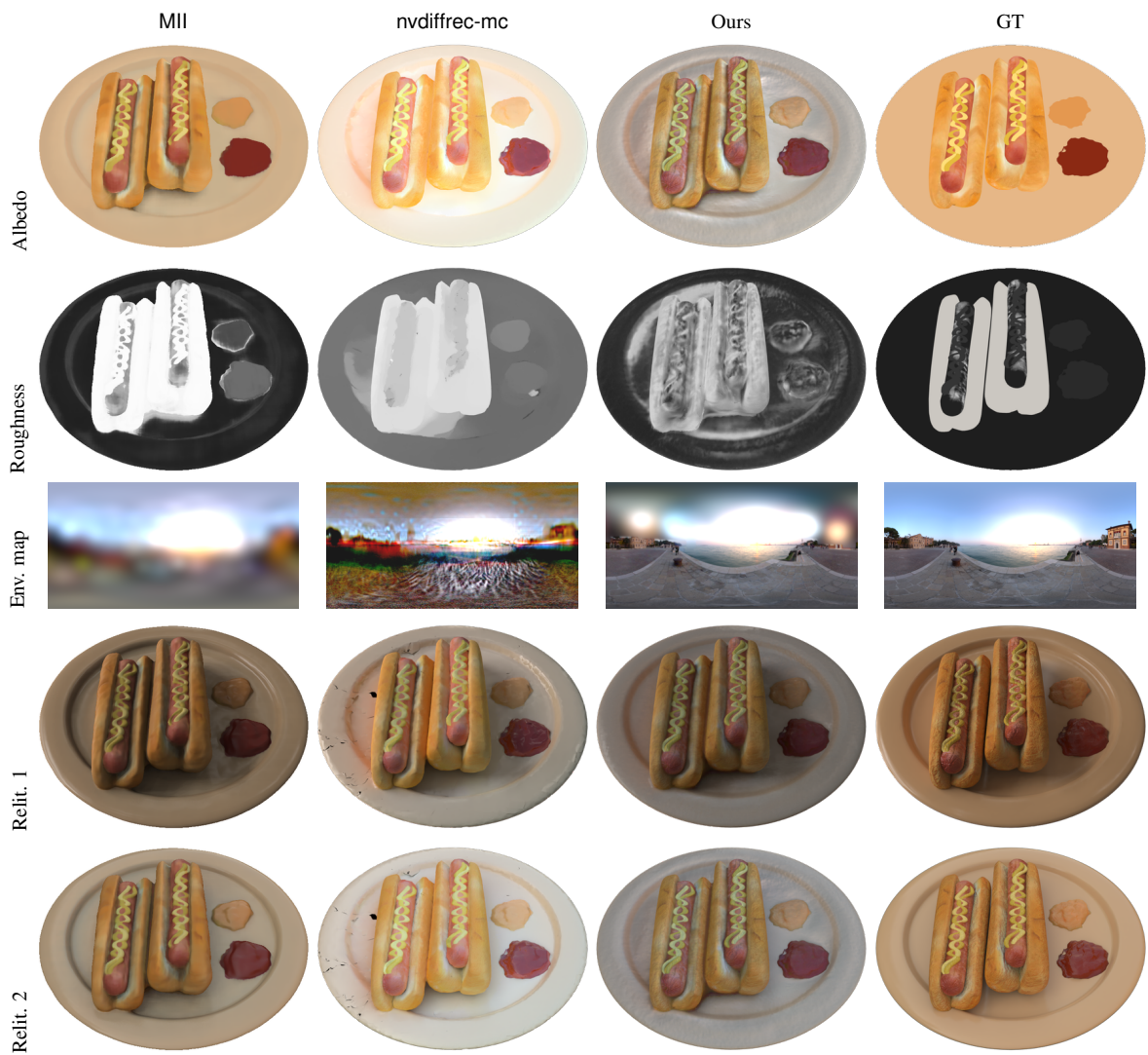

\setlength{\resLen}{1.5in}
\centering
\begin{tabular}{@{}c@{\hskip 2pt}c@{\hskip 0pt}c@{\hskip 0pt}c@{\hskip 0pt}c@{}}
& {\footnotesize \mii} & {\footnotesize \nvdiff} & {\footnotesize Ours} & {\footnotesize GT}
\\
\vtext{{\footnotesize Albedo}} & 
\miiimage{trim={{0.075\width} {0.125\height} {0.075\width} {0.225\height}},clip,width=\resLen}{mii/hotdog/im_0001_albedo.png} & 
\miiimage{trim={{0.075\width} {0.125\height} {0.075\width} {0.225\height}},clip,width=\resLen}{nvdiffrec/hotdog/im_0001_albedo.png} & 
\miiimage{trim={{0.075\width} {0.125\height} {0.075\width} {0.225\height}},clip,width=\resLen}{ours/hotdog/im_0001_albedo.png} & %
\miiimage{trim={{0.075\width} {0.125\height} {0.075\width} {0.225\height}},clip,width=\resLen}{GT/hotdog/im_0001_albedo.png}
\\
\vtext{{\footnotesize Roughness}} &
\miiimage{trim={{0.075\width} {0.125\height} {0.075\width} {0.225\height}},clip,width=\resLen}{mii/hotdog/im_0001_roughness.png} & 
\miiimage{trim={{0.075\width} {0.125\height} {0.075\width} {0.225\height}},clip,width=\resLen}{nvdiffrec/hotdog/im_0001_rough.png} & 
\miiimage{trim={{0.075\width} {0.125\height} {0.075\width} {0.225\height}},clip,width=\resLen}{ours/hotdog/im_0001_roughness.png}& %
\miiimage{trim={{0.075\width} {0.125\height} {0.075\width} {0.225\height}},clip,width=\resLen}{GT/hotdog/im_0001_rough.png}
\\
\vtext{{\footnotesize Env. map}} & 
\miiimage{width=.9\resLen}{mii/hotdog/envmap.png} & 
\miiimage{width=.9\resLen}{nvdiffrec/hotdog/envmap.png} & 
 \miiimage{width=.9\resLen}{ours/hotdog/envmap.png} & %
\miiimage{width=.9\resLen}{GT/chair/envmap.png}
\\
\vtext{\footnotesize Relit. 1} & 
\miiimage{trim={{0.075\width} {0.125\height} {0.075\width} {0.225\height}},clip,width=\resLen}{mii/hotdog/im_0001_envmap6.png} & 
\miiimage{trim={{0.075\width} {0.125\height} {0.075\width} {0.225\height}},clip,width=\resLen}{nvdiffrec/hotdog/im_0001_envmap6.png} & 
\miiimage{trim={{0.075\width} {0.125\height} {0.075\width} {0.225\height}},clip,width=\resLen}{ours/hotdog/im_0001_envmap6.png} & %
\miiimage{trim={{0.075\width} {0.125\height} {0.075\width} {0.225\height}},clip,width=\resLen}{GT/hotdog/im_0001_envmap6.png}
\\
\vtext{\footnotesize Relit. 2} & 
\miiimage{trim={{0.075\width} {0.125\height} {0.075\width} {0.225\height}},clip,width=\resLen}{mii/hotdog/im_0001_envmap12.png} & 
\miiimage{trim={{0.075\width} {0.125\height} {0.075\width} {0.225\height}},clip,width=\resLen}{nvdiffrec/hotdog/im_0001_envmap12.png} & 
\miiimage{trim={{0.075\width} {0.125\height} {0.075\width} {0.225\height}},clip,width=\resLen}{ours/hotdog/im_0001_envmap12.png}& %
\miiimage{trim={{0.075\width} {0.125\height} {0.075\width} {0.225\height}},clip,width=\resLen}{GT/hotdog/im_0001_envmap12.png}
\end{tabular}
\caption{{\bf Qualitative comparisons of {\it hotdog} from the \mii dataset.}
}
\label{fig:mii_vis_hotdog}
\end{figure*}

\begin{figure*}
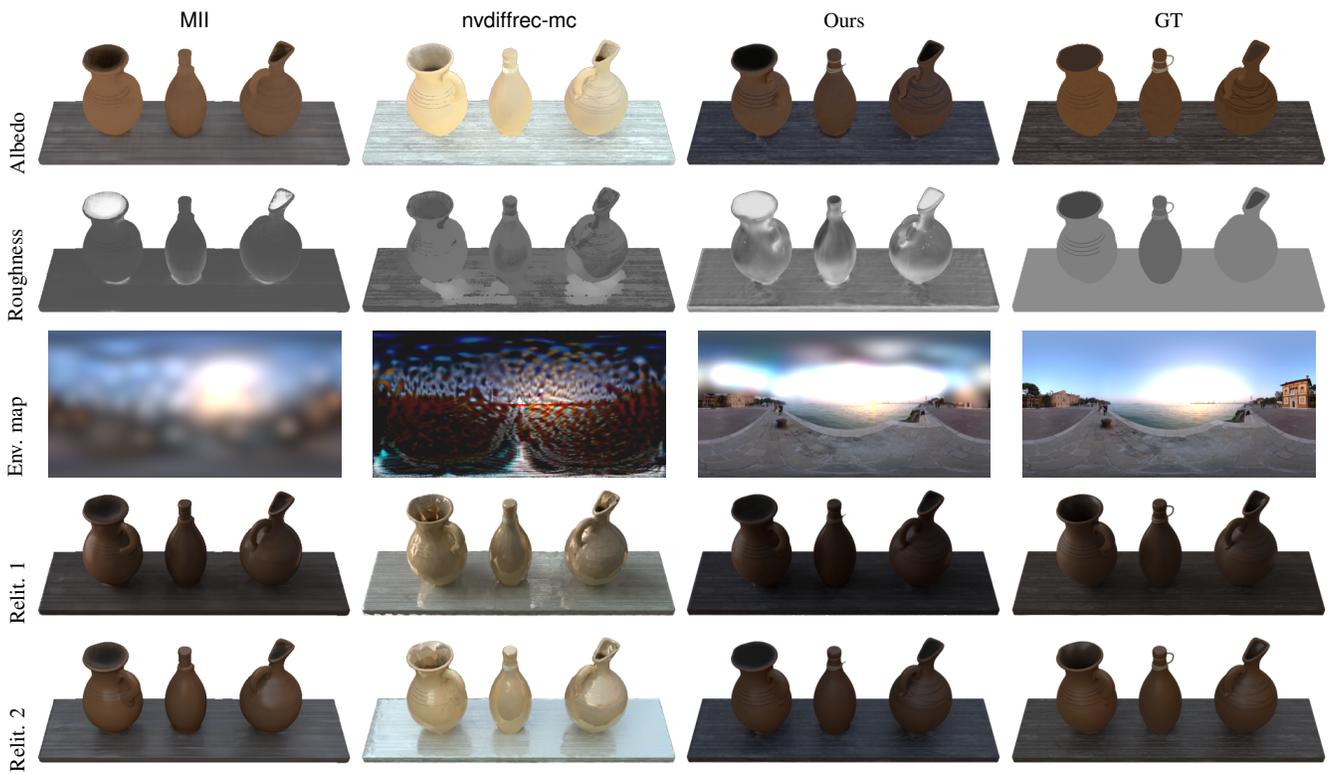

\centering
\setlength{\resLen}{1.7in}
\begin{tabular}{@{}c@{\hskip 2pt}c@{\hskip 0pt}c@{\hskip 0pt}c@{\hskip 0pt}c@{}}
& {\footnotesize \mii} & {\footnotesize \nvdiff} & {\footnotesize Ours} & {\footnotesize GT}
\\
\vtext{{\footnotesize Albedo}} & 
\miiimage{trim={{0.05\width} {0.3625\height} {0.0375\width} {0.25\height}},clip,width=\resLen}{mii/jugs/im_0001_albedo.png} & 
\miiimage{trim={{0.05\width} {0.3625\height} {0.0375\width} {0.25\height}},clip,width=\resLen}{nvdiffrec/jugs/im_0001_albedo.png} & 
\miiimage{trim={{0.05\width} {0.3625\height} {0.0375\width} {0.25\height}},clip,width=\resLen}{ours/jugs/im_0001_albedo.png} & 
\miiimage{trim={{0.05\width} {0.3625\height} {0.0375\width} {0.25\height}},clip,width=\resLen}{GT/jugs/im_0001_albedo.png}
\\
\vtext{{\footnotesize Roughness}} &
\miiimage{trim={{0.05\width} {0.3625\height} {0.0375\width} {0.25\height}},clip,width=\resLen}{mii/jugs/im_0001_roughness.png} & 
\miiimage{trim={{0.05\width} {0.3625\height} {0.0375\width} {0.25\height}},clip,width=\resLen}{nvdiffrec/jugs/im_0001_roughness.png} & 
\miiimage{trim={{0.05\width} {0.3625\height} {0.0375\width} {0.25\height}},clip,width=\resLen}{ours/jugs/im_0001_roughness.png} & 
\miiimage{trim={{0.05\width} {0.3625\height} {0.0375\width} {0.25\height}},clip,width=\resLen}{GT/jugs/im_0001_roughness.png}
\\
\vtext{{\footnotesize Env. map}} & 
\miiimage{width=.9\resLen}{mii/jugs/envmap.png} & 
\miiimage{width=.9\resLen}{nvdiffrec/jugs/envmap.png} & 
\miiimage{width=.9\resLen}{ours/jugs/envmap.png} &
\miiimage{width=.9\resLen}{GT/chair/envmap.png}
\\
\vtext{\footnotesize Relit. 1} & 
\miiimage{trim={{0.05\width} {0.3625\height} {0.0375\width} {0.25\height}},clip,width=\resLen}{mii/jugs/im_0001_envmap6.png} & 
\miiimage{trim={{0.05\width} {0.3625\height} {0.0375\width} {0.25\height}},clip,width=\resLen}{nvdiffrec/jugs/im_0001_envmap6.png} & 
\miiimage{trim={{0.05\width} {0.3625\height} {0.0375\width} {0.25\height}},clip,width=\resLen}{ours/jugs/im_0001_envmap6.png} & 
\miiimage{trim={{0.05\width} {0.3625\height} {0.0375\width} {0.25\height}},clip,width=\resLen}{GT/jugs/im_0001_envmap6.png}
\\
\vtext{\footnotesize Relit. 2} & 
\miiimage{trim={{0.05\width} {0.3625\height} {0.0375\width} {0.25\height}},clip,width=\resLen}{mii/jugs/im_0001_envmap12.png} & 
\miiimage{trim={{0.05\width} {0.3625\height} {0.0375\width} {0.25\height}},clip,width=\resLen}{nvdiffrec/jugs/im_0001_envmap12.png} & 
\miiimage{trim={{0.05\width} {0.3625\height} {0.0375\width} {0.25\height}},clip,width=\resLen}{ours/jugs/im_0001_envmap12.png} & 
\miiimage{trim={{0.05\width} {0.3625\height} {0.0375\width} {0.25\height}},clip,width=\resLen}{GT/jugs/im_0001_envmap12.png}
\end{tabular}
\caption{{\bf Qualitative comparisons of {\it jugs} from the \mii dataset.}
}
\label{fig:mii_vis_jugs}
\end{figure*}

%% file: supp/eval_ours_synthetic.tex
\paragraph{Our synthetic dataset.}
Since the \mii dataset does not contain groundtruth meshes, it is difficult to evaluate the accuracy of reconstructed shapes.
To address this, we create two extra synthetic scenes---\textit{buddha} and \textit{lion}---with groundtruth meshes for evaluation.
For each scene, the training set includes 190 posed images with masks.
The testing set consists of visualizations of groundtruth albedo, roughness, and renderings of the object under seven novel lighting conditions in 10 poses. %

Table \ref{tab:ours_synthetic} shows quantitative comparisons between our method and the baselines. In addition to metrics used in the MII dataset, we also measure Chamfer distances~\cite{barrow1977parametric} between optimized and groundtruth shapes (normalized so that the groundtruth has unit bounding boxes).
Our method again outperforms the baselines. 

As shown in \cref{fig:ours_vis_lion,fig:ours_vis_buddha}, since the background is fully visible (i.e., each pixel of $\Lenv$ is visible as the background of at least one input image), our method is capable of reconstructing the environment map almost perfectly.
Because of this, our albedo reconstructions are not hindered by the albedo-light ambiguity---as demonstrated in \cref{tab:ours_synthetic} where the error metrics barely change with or without albedo alignment.
We note that this might not apply to all scenarios, for instance, the background might not be fully visible, as shown in the \mii dataset.
Reconstructing indoor lighting perfectly is also challenging even if the background is completely visible, because it breaks the assumption of environmental (i.e., distant) lighting. %

\input{supp/ours_synthetic_table}
\input{supp/ours_synthetic_visual}

%% file: supp/ours_synthetic_table.tex
\begin{table*}
\centering
{
\resizebox{\textwidth}{!}{
\begin{tabular}{ @{}l@{\hskip 5pt} @{}r@{} c@{\hskip 7pt} @{}c@{\hskip 5pt} c@{\hskip 5pt} c@{} c@{\hskip 7pt} @{}c@{\hskip 5pt} c@{\hskip 5pt} c@{} c@{\hskip 7pt} @{}c@{\hskip 5pt} c@{\hskip 5pt} c@{} c@{\hskip 7pt} @{}c@{} c@{\hskip 7pt} @{}c@{}}
    \toprule
    &Speed && \multicolumn{3}{c}{Relighting} & &\multicolumn{3}{c}{Aligned albedo} & & \multicolumn{3}{c}{Albedo} && Rough. && Shape\\
    \cmidrule{2-2} 
    \cmidrule{4-6}
    \cmidrule{8-10}
    \cmidrule{12-14}
    \cmidrule{16-16}
    \cmidrule{18-18}
    
    Method & Time$\downarrow$ & & PSNR$\uparrow$ & SSIM$\uparrow$ & LPIPS$\downarrow$ & &PSNR$\uparrow$ & SSIM$\uparrow$ & LPIPS$\downarrow$ & & PSNR$\uparrow$ & SSIM$\uparrow$ & LPIPS$\downarrow$ && MSE$\downarrow$ && CD$\downarrow$\\
    \midrule
    \nvdiff~\cite{HasselgrenHM22} & $\sim$ 2 h & & 21.77 & 0.936 & 0.071 && 33.29 & 0.964 & 0.037 &&  16.14 & 0.910 & 0.068 && 0.013 && 9.42e-5\\
    \mii~\cite{ZhangSHFJZ22} & $\sim$ 10 h & & 24.94 & 0.952 & 0.051 && 30.92 & 0.962 & 0.044 &&  19.80 & 0.923 & 0.065 && \underline{0.003} && 5.92e-5\\
    \midrule
    Ours - Distilled only & $<$ 15 m && 33.90 & 0.976 & 0.034 && 34.09 & 0.971 & 0.034 && 34.09 & 0.972 & 0.034 && 0.005 && \underline{2.61e-5} \\
    Ours - w/o shape ref. & $\sim$ 45 m & & \underline{34.18} & \underline{0.980} & \underline{0.028} && \underline{35.57} & \underline{0.983} & \underline{0.026} && \underline{35.57} & \underline{0.983} & \underline{0.026} && \underline{0.003} && \underline{2.61e-5} \\
    Ours - Full & $\sim$ 1 h & & \textbf{35.30} & \textbf{0.982} & \textbf{0.026} && \textbf{37.69} & \textbf{0.985} & \textbf{0.023} && \textbf{37.68} & \textbf{0.985} & \textbf{0.023} && \textbf{0.002} && \textbf{2.56e-5} \\
    \bottomrule
\end{tabular}
}
}
\caption{\textbf{Relighting, material reconstruction, and mesh quality on our synthetic dataset.} We compare our
method with \mii and \nvdiff. The highest performing number is presented in bold, while the second best is underscored. We measure the shape quality using Chamfer distances (CD). }
\label{tab:ours_synthetic}
\end{table*}

%% file: supp/ours_synthetic_visual.tex
\newcommand{\oursynimage}[2]{\adjincludegraphics[#1]{figs/ours_synthetic/#2}}

\begin{figure*}
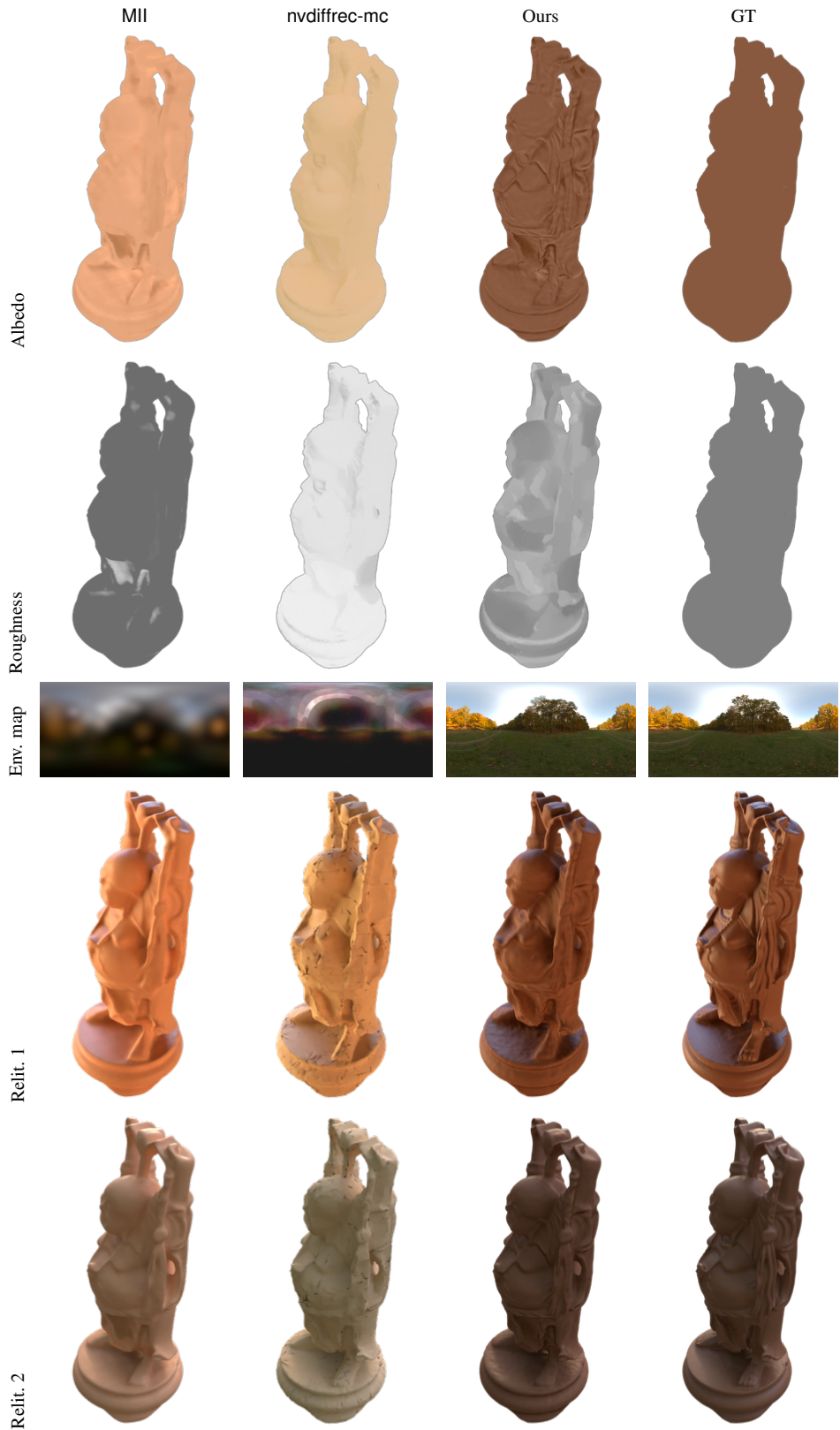

    \setlength{\resLen}{.8in}
    \addtolength{\tabcolsep}{-3pt}
    \centering
    \begin{tabular}{ccccc}
    & {\footnotesize \mii} & {\footnotesize \nvdiff} & {\footnotesize Ours} & {\footnotesize GT}
    \\
    \vtext{{\footnotesize Albedo}} & 
    \oursynimage{trim={{0.35\width} {0.19375\height} {0.35\width} {0.10\height}},clip,width=\resLen}{mii/buddha/im_0003_albedo.png} & 
    \oursynimage{trim={{0.35\width} {0.19375\height} {0.35\width} {0.10\height}},clip,width=\resLen}{nvdiffrec/buddha/im_0003_albedo.png} & 
    \oursynimage{trim={{0.35\width} {0.19375\height} {0.35\width} {0.10\height}},clip,width=\resLen}{ours/buddha/im_0003_albedo.png} & 
    \oursynimage{trim={{0.35\width} {0.19375\height} {0.35\width} {0.10\height}},clip,width=\resLen}{GT/buddha/im_0003_albedo.png}
    \\
    \vtext{{\footnotesize Roughness}} &
    \oursynimage{trim={{0.35\width} {0.19375\height} {0.35\width} {0.10\height}},clip,width=\resLen}{mii/buddha/im_0003_roughness.png} & 
    \oursynimage{trim={{0.35\width} {0.19375\height} {0.35\width} {0.10\height}},clip,width=\resLen}{nvdiffrec/buddha/im_0003_roughness.png} & 
    \oursynimage{trim={{0.35\width} {0.19375\height} {0.35\width} {0.10\height}},clip,width=\resLen}{ours/buddha/im_0003_roughness.png} & 
    \oursynimage{trim={{0.35\width} {0.19375\height} {0.35\width} {0.10\height}},clip,width=\resLen}{GT/buddha/im_0003_roughness.png}
    \\
    \vtext{{\footnotesize Env. map}} & 
    \oursynimage{width=1.4\resLen}{mii/buddha/envmap.png} & 
    \oursynimage{width=1.4\resLen}{nvdiffrec/buddha/envmap.png} & 
    \oursynimage{width=1.4\resLen}{ours/buddha/envmap.png} &
    \oursynimage{width=1.4\resLen}{GT/buddha/envmap.png}
    \\
    \vtext{\footnotesize Relit. 1} & 
    \oursynimage{trim={{0.35\width} {0.19375\height} {0.35\width} {0.10\height}},clip,width=\resLen}{mii/buddha/im_0003_courtyard.png} & 
    \oursynimage{trim={{0.35\width} {0.19375\height} {0.35\width} {0.10\height}},clip,width=\resLen}{nvdiffrec/buddha/im_0003_courtyard.png} & 
    \oursynimage{trim={{0.35\width} {0.19375\height} {0.35\width} {0.10\height}},clip,width=\resLen}{ours/buddha/im_0003_courtyard.png} & 
    \oursynimage{trim={{0.35\width} {0.19375\height} {0.35\width} {0.10\height}},clip,width=\resLen}{GT/buddha/im_0003_courtyard.png}
    \\
    \vtext{\footnotesize Relit. 2} & 
    \oursynimage{trim={{0.35\width} {0.19375\height} {0.35\width} {0.10\height}},clip,width=\resLen}{mii/buddha/im_0003_forest.png} & 
    \oursynimage{trim={{0.35\width} {0.19375\height} {0.35\width} {0.10\height}},clip,width=\resLen}{nvdiffrec/buddha/im_0003_forest.png} & 
    \oursynimage{trim={{0.35\width} {0.19375\height} {0.35\width} {0.10\height}},clip,width=\resLen}{ours/buddha/im_0003_forest.png} & 
    \oursynimage{trim={{0.35\width} {0.19375\height} {0.35\width} {0.10\height}},clip,width=\resLen}{GT/buddha/im_0003_forest.png}
    \end{tabular}
    \caption{
        {\bf Qualitative comparisons of {\it buddha} from our dataset.}
    }
    \label{fig:ours_vis_buddha}
\end{figure*}

\begin{figure*}
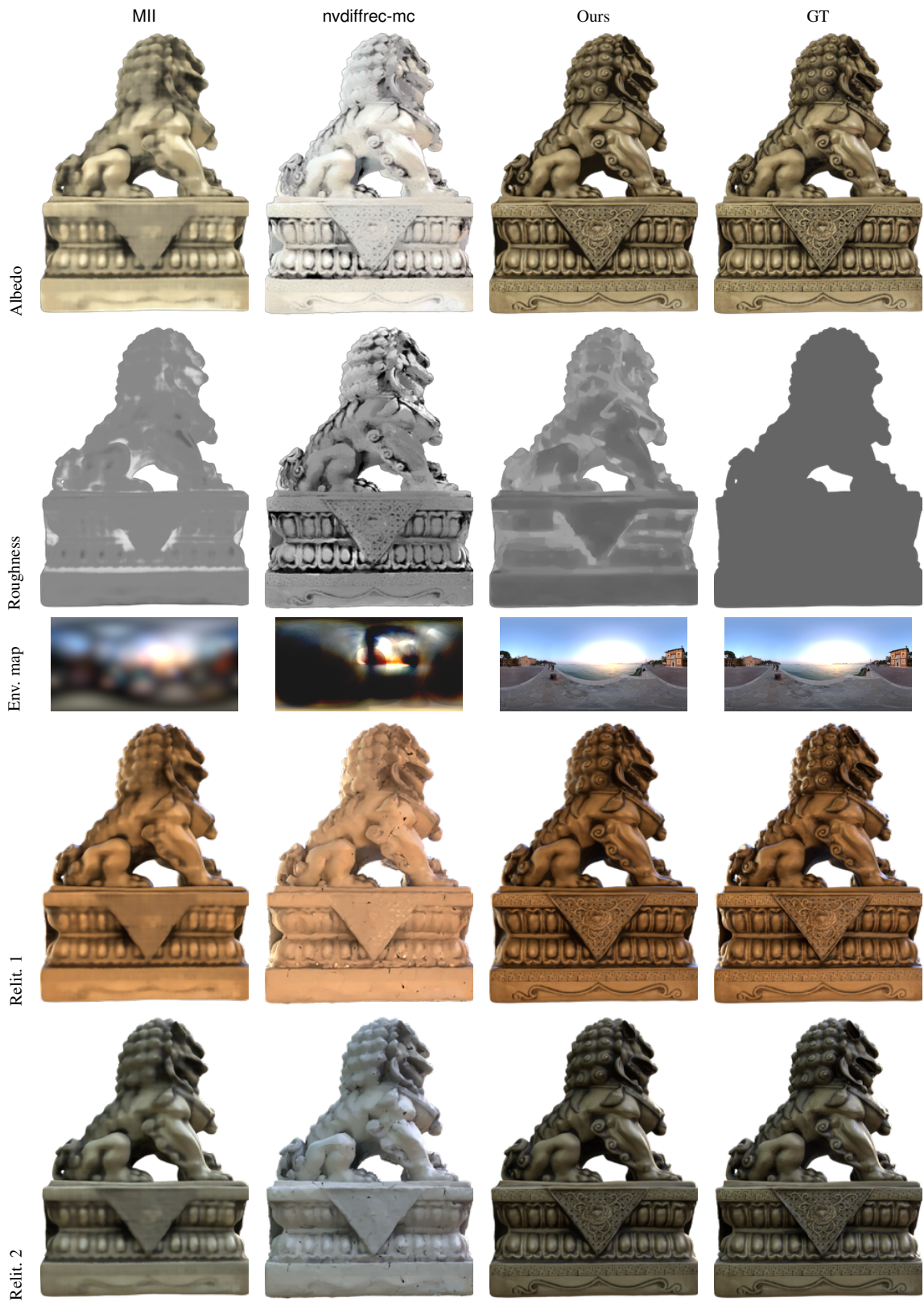

    \setlength{\resLen}{1.35in}
    \addtolength{\tabcolsep}{-2pt}
    \centering
    \begin{tabular}{ccccc}
    & {\footnotesize \mii} & {\footnotesize \nvdiff} & {\footnotesize Ours} & {\footnotesize GT}
    \\
    \vtext{{\footnotesize Albedo}} & 
    \oursynimage{trim={{0.2875\width} {0.3\height} {0.2875\width} {0.1125\height}},clip,width=\resLen}{mii/lion/im_0006_albedo.png} & 
    \oursynimage{trim={{0.2875\width} {0.3\height} {0.2875\width} {0.1125\height}},clip,width=\resLen}{nvdiffrec/lion/im_0006_albedo.png} & 
    \oursynimage{trim={{0.2875\width} {0.3\height} {0.2875\width} {0.1125\height}},clip,width=\resLen}{ours/lion/im_0006_albedo.png} & 
    \oursynimage{trim={{0.2875\width} {0.3\height} {0.2875\width} {0.1125\height}},clip,width=\resLen}{GT/lion/im_0006_albedo.png}
    \\
    \vtext{{\footnotesize Roughness}} &
    \oursynimage{trim={{0.2875\width} {0.3\height} {0.2875\width} {0.1125\height}},clip,width=\resLen}{mii/lion/im_0006_roughness.png} & 
    \oursynimage{trim={{0.2875\width} {0.3\height} {0.2875\width} {0.1125\height}},clip,width=\resLen}{nvdiffrec/lion/im_0006_roughness.png} & 
    \oursynimage{trim={{0.2875\width} {0.3\height} {0.2875\width} {0.1125\height}},clip,width=\resLen}{ours/lion/im_0006_roughness.png} & 
    \oursynimage{trim={{0.2875\width} {0.3\height} {0.2875\width} {0.1125\height}},clip,width=\resLen}{GT/lion/im_0006_roughness.png}
    \\
    \vtext{{\footnotesize Env. map}} & 
    \oursynimage{width=.9\resLen}{mii/lion/envmap.png} & 
    \oursynimage{width=.9\resLen}{nvdiffrec/lion/envmap.png} & 
    \oursynimage{width=.9\resLen}{ours/lion/envmap.png} &
    \oursynimage{width=.9\resLen}{GT/lion/envmap.png}
    \\
    \vtext{\footnotesize Relit. 1} & 
    \oursynimage{trim={{0.2875\width} {0.3\height} {0.2875\width} {0.1125\height}},clip,width=\resLen}{mii/lion/im_0006_courtyard.png} & 
    \oursynimage{trim={{0.2875\width} {0.3\height} {0.2875\width} {0.1125\height}},clip,width=\resLen}{nvdiffrec/lion/im_0006_courtyard.png} & 
    \oursynimage{trim={{0.2875\width} {0.3\height} {0.2875\width} {0.1125\height}},clip,width=\resLen}{ours/lion/im_0006_courtyard.png} & 
    \oursynimage{trim={{0.2875\width} {0.3\height} {0.2875\width} {0.1125\height}},clip,width=\resLen}{GT/lion/im_0006_courtyard.png}
    \\
    \vtext{\footnotesize Relit. 2} & 
    \oursynimage{trim={{0.2875\width} {0.3\height} {0.2875\width} {0.1125\height}},clip,width=\resLen}{mii/lion/im_0006_forest.png} & 
    \oursynimage{trim={{0.2875\width} {0.3\height} {0.2875\width} {0.1125\height}},clip,width=\resLen}{nvdiffrec/lion/im_0006_forest.png} & 
    \oursynimage{trim={{0.2875\width} {0.3\height} {0.2875\width} {0.1125\height}},clip,width=\resLen}{ours/lion/im_0006_forest.png} & 
    \oursynimage{trim={{0.2875\width} {0.3\height} {0.2875\width} {0.1125\height}},clip,width=\resLen}{GT/lion/im_0006_forest.png}
    \end{tabular}
    \caption{{\bf Qualitative comparisons of {\it lion} from our dataset.}}
    \label{fig:ours_vis_lion}
\end{figure*}

%% file: supp/eval_surface.tex
\paragraph{Surface quality on the DTU dataset.}
We show quantitative results breakdown for the 15 scenes from DTU dataset~\cite{JensenDVTA14} in \cref{tab:dtu_detail}.
We use the official evaluation script to measure Chamfer distances.
Please note that our results evaluated here are directly from the shape reconstruction stage.
We skip evaluating the shape refinement of our physics-based inverse rendering on DTU dataset as DTU exhibit vary light occlusion from robot arms.

\input{supp/table_dtu_breakdown}

\paragraph{Usefulness of shape refinement.}
Lastly, we demonstrate the usefulness of our shape refinement (as the last step of the physics-based inverse rendering stage) via an ablation.
As shown in \cref{fig:our_vis_abla} and \cref{tab:ours_synthetic}, our shape refinement improves the accuracy of reconstructed object geometries.

\input{supp/ours_synthetic_abla}

%% file: supp/table_dtu_breakdown.tex
\definecolor{grayresult}{rgb}{.7, .7, .7}
\newcommand{\concur}[1]{\textcolor{grayresult}{#1}}

\begin{table*}
\centering
\begin{tabular}{@{}l@{\hskip 8pt}c|c@{\hskip 12pt}c@{\hskip 8pt}c@{\hskip 8pt}c@{\hskip 8pt}c@{\hskip 8pt}c@{\hskip 8pt}c@{\hskip 8pt}c@{\hskip 8pt}c@{\hskip 8pt}c@{\hskip 8pt}c@{\hskip 8pt}c@{\hskip 8pt}c@{\hskip 8pt}c@{\hskip 8pt}c@{\hskip 8pt}c@{}}
    \hline
    Method & Time & avg. & 24 & 37 & 40 & 55 & 63 & 65 & 69 & 83 & 97 & 105 & 106 & 110 & 114 & 118 & 122
    \\
    \hline\hline
    COLMAP & 1 h & 1.36 & 0.81 & 2.05 & 0.73 & 1.22 & 1.79 & 1.58 & 1.02 & 3.05 & 1.40 & 2.05 & 1.00 & 1.32 & 0.49 & 0.78 & 1.17
    \\
    NeuS & 5.5 h & 0.77 & 0.83       & 0.98       & 0.56       & 0.37       & 1.13       & {\bf 0.59} & {\bf 0.60} & 1.45       & {\bf 0.95} & 0.78       & {\bf 0.52} & 1.43       & {\bf 0.36} & 0.45 & {\bf 0.45}
    \\
    Ours & {\bf 5 m} & {\bf 0.66} & {\bf 0.52} & {\bf 0.72} & {\bf 0.36} & {\bf 0.35} & {\bf 0.97} & 0.68 & 0.61 & {\bf 1.27} & 1.06 & {\bf 0.71} & {\bf 0.52} & {\bf 0.78} & {\bf 0.36} & {\bf 0.43} & 0.56
    \\
    \hline
    \hline
\end{tabular}
\par\medskip
\newcommand{\dtuimage}[1]{\includegraphics[width=0.245\linewidth]{#1}}
\begin{tabular}{@{}c@{\hskip 3pt}c@{\hskip 3pt}c@{\hskip 3pt}c@{}}
NeuS & Ours & NeuS & Ours \\
\dtuimage{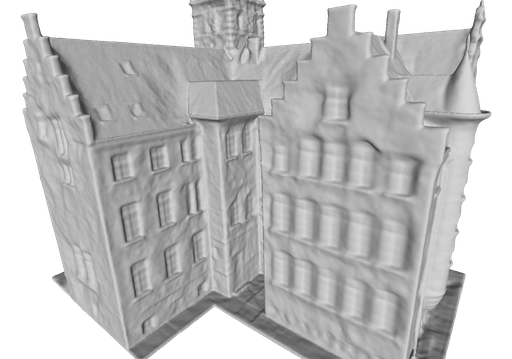} & \dtuimage{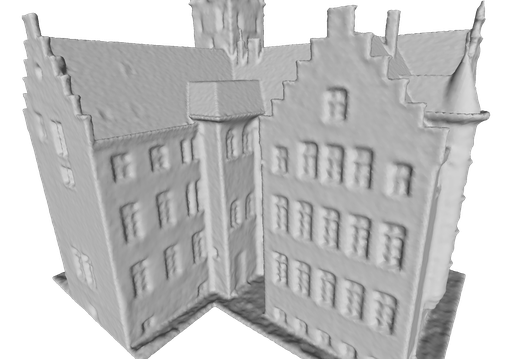} & \dtuimage{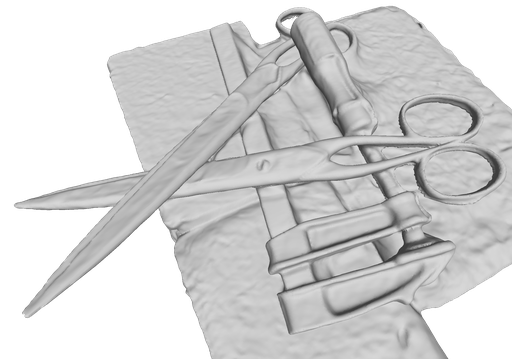} & \dtuimage{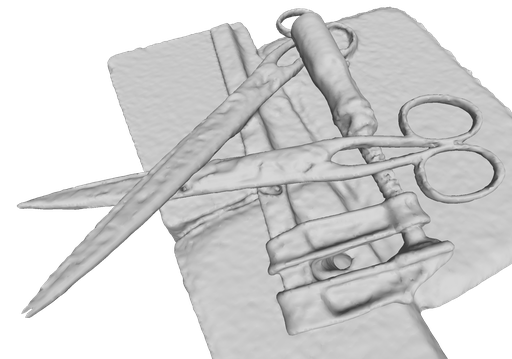} \\
\dtuimage{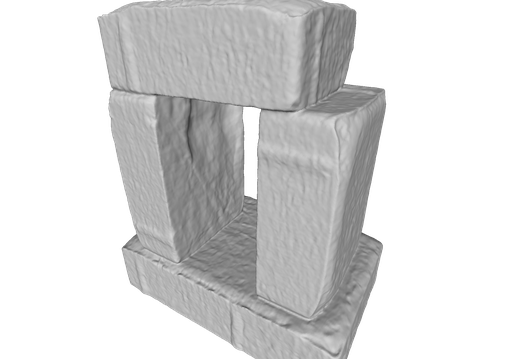} & \dtuimage{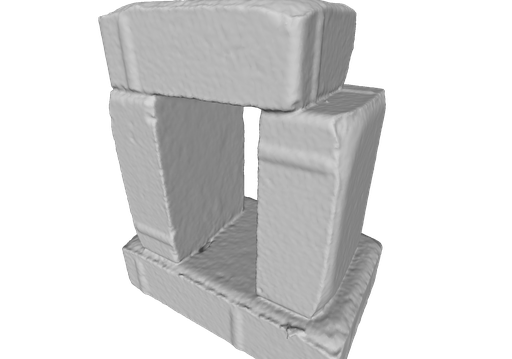} & \dtuimage{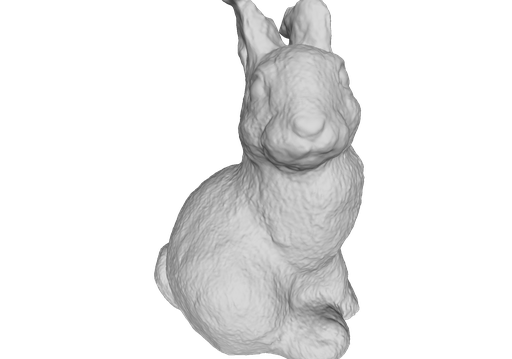} & \dtuimage{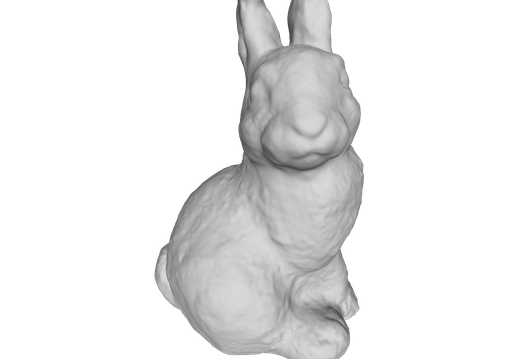} \\
\dtuimage{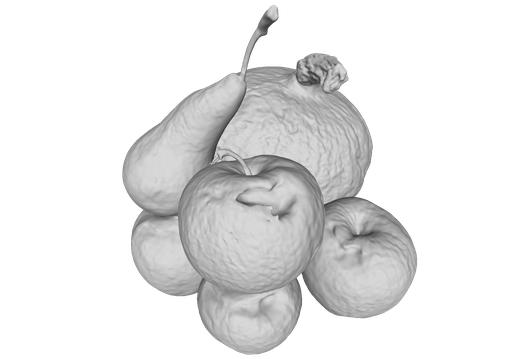} & \dtuimage{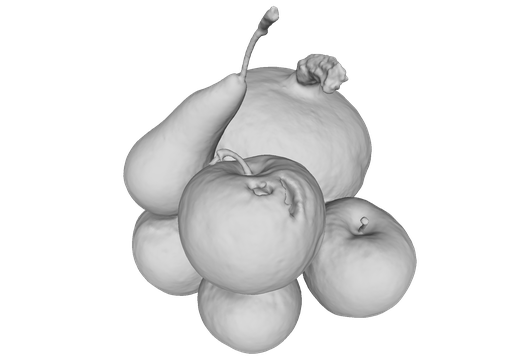} & \dtuimage{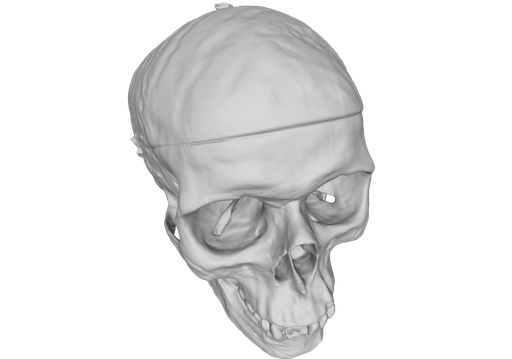} & \dtuimage{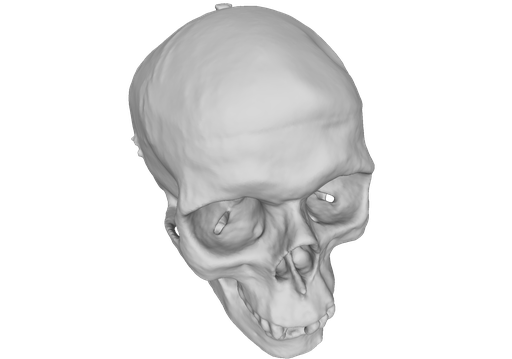} \\
\dtuimage{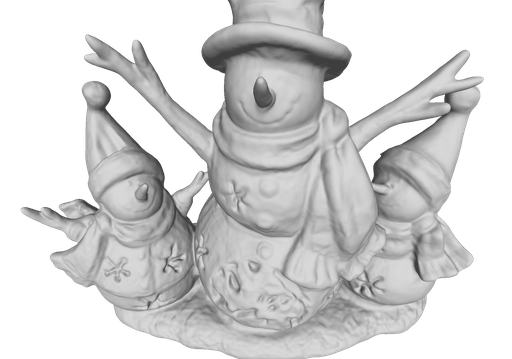} & \dtuimage{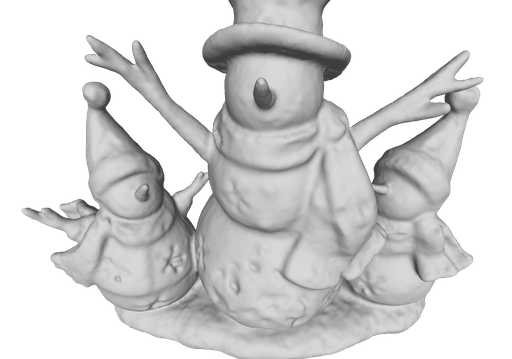} & \dtuimage{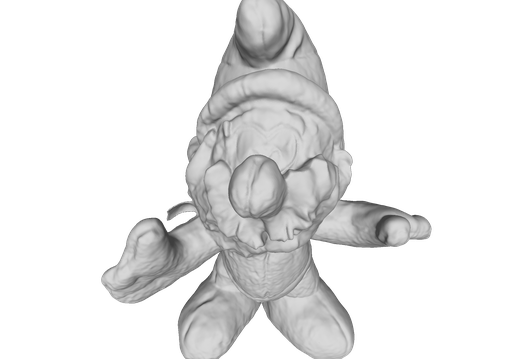} & \dtuimage{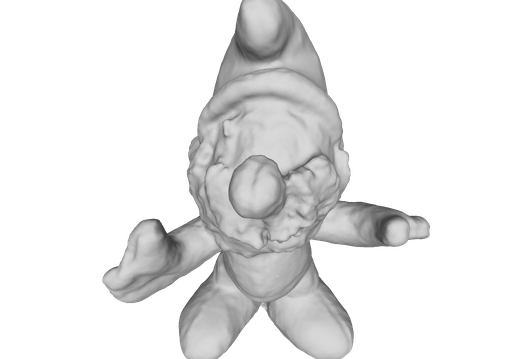} \\
\dtuimage{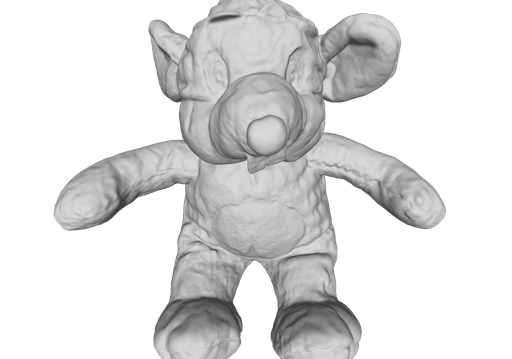} & \dtuimage{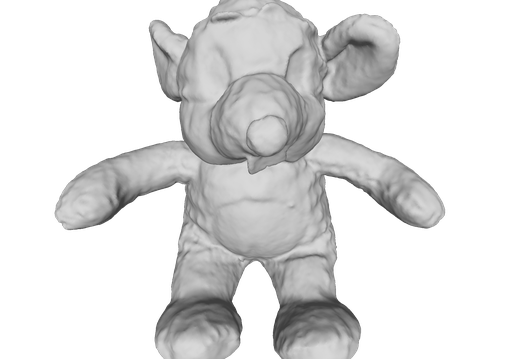} & \dtuimage{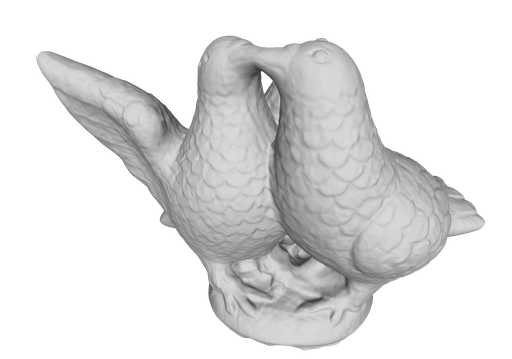} & \dtuimage{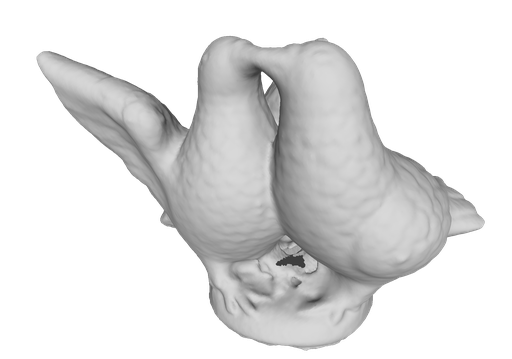} \\
\dtuimage{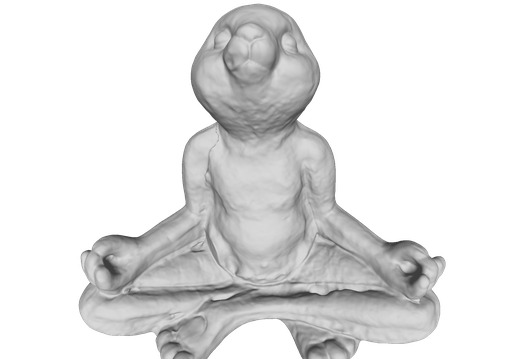} & \dtuimage{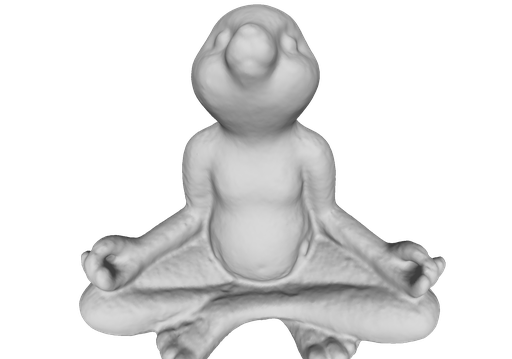} & \dtuimage{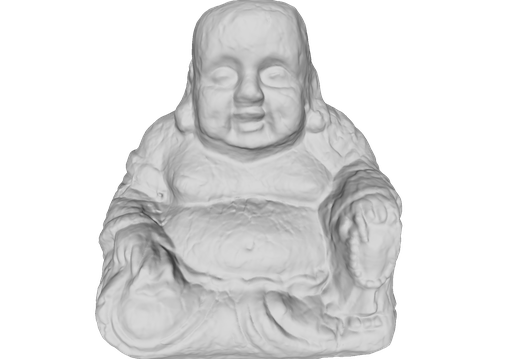} & \dtuimage{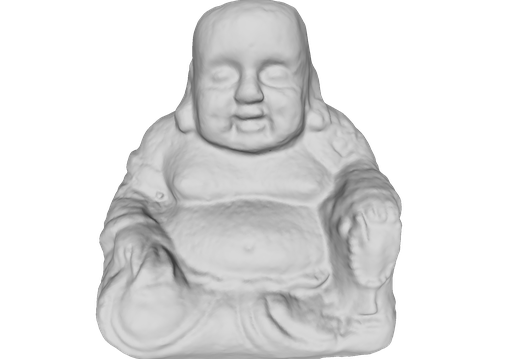} \\
\end{tabular}
\caption{{\bf Quantitative results breakdown and visualization on the DTU MVS dataset~\cite{JensenDVTA14}.}
We use official evaluation script to measure Chamfer distances (in mm).
Our results are typically smoother with some details missing.
We do not apply PBIR shape refinement as DTU exhibits significant lighting variation.
See \cref{fig:our_vis_abla} and the main paper for the experiments about shape refinement.
}
\label{tab:dtu_detail}
\end{table*}

%% file: supp/ours_synthetic_abla.tex
\newcommand{\oursynablaimage}[2]{\includegraphics[#1]{figs/abla_buddha_lion/#2}}

\begin{figure*}
    \centering
    \resizebox{\textwidth}{!}{%
    \begin{tikzpicture}[x=1pt, y=1pt]
        \node[anchor=south west] (img1) at (0pt,0pt) {\oursynablaimage{trim=20 20 20 20,clip,width=100pt}{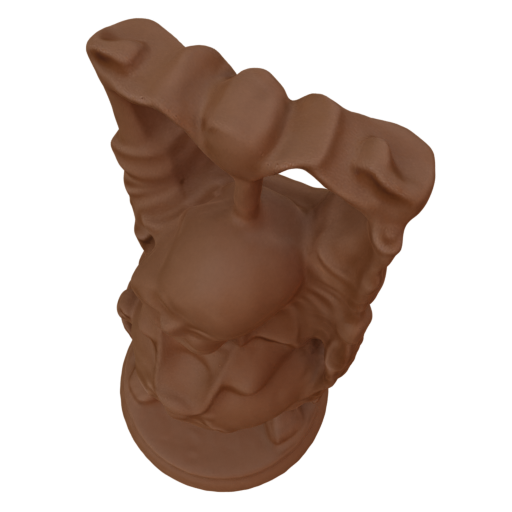}};
        \draw[green, line width=1pt] (37pt,60pt) rectangle +(40pt,40pt);
        \draw[cyan, line width=1pt] (30pt,19pt) rectangle +(20pt,20pt);
        \node[anchor=south west, draw=cyan, inner sep=1pt, line width=1pt] (img1c) at (0pt,-50pt) 
        {\oursynablaimage{trim=144 96 272 320,clip,width=47pt}{buddha_wo.png}};
        \node[anchor=south west, draw=green, inner sep=1pt, line width=1pt] (img1rt) at (50pt,-50pt) 
        {\oursynablaimage{trim=176 288 144 32,clip,width=47pt}{buddha_wo.png}};
        \node[anchor=south west, draw=cyan, inner sep=1pt, line width=1pt] at (0pt,-100pt) 
        {\oursynablaimage{trim=144 96 272 320,clip,width=47pt}{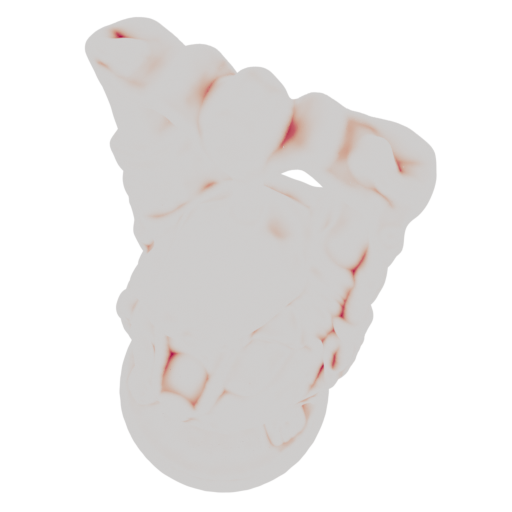}};
        \node[anchor=south west, draw=green, inner sep=1pt, line width=1pt] (img1et) at (50pt,-100pt) 
        {\oursynablaimage{trim=176 288 144 32,clip,width=47pt}{buddha_wo_cd.png}};

        \node[anchor=south west] (img2) at (120pt,0pt) {\oursynablaimage{trim=20 20 20 20,clip,width=100pt}{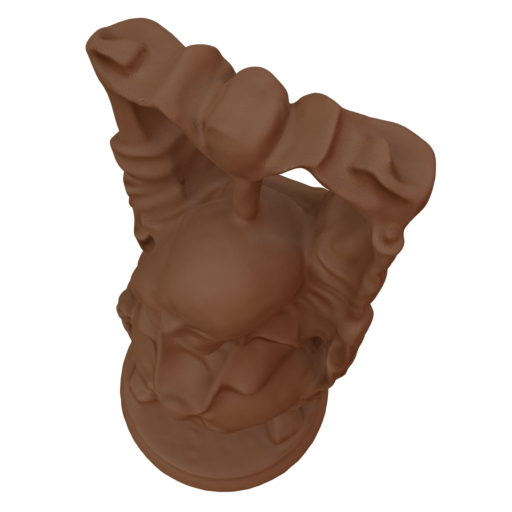}};
        \draw[green, line width=1pt] (157pt,60pt) rectangle +(40pt,40pt);
        \draw[cyan, line width=1pt] (150pt,19pt) rectangle +(20pt,20pt);
        \node[anchor=south west, draw=cyan, inner sep=1pt, line width=1pt] (img2c) at (120pt,-50pt) 
        {\oursynablaimage{trim=144 96 272 320,clip,width=47pt}{buddha_w.png}};
        \node[anchor=south west, draw=green, inner sep=1pt, line width=1pt] (img2rt) at (170pt,-50pt) 
        {\oursynablaimage{trim=176 288 144 32,clip,width=47pt}{buddha_w.png}};
        \node[anchor=south west, draw=cyan, inner sep=1pt, line width=1pt] at (120pt,-100pt) 
        {\oursynablaimage{trim=144 96 272 320,clip,width=47pt}{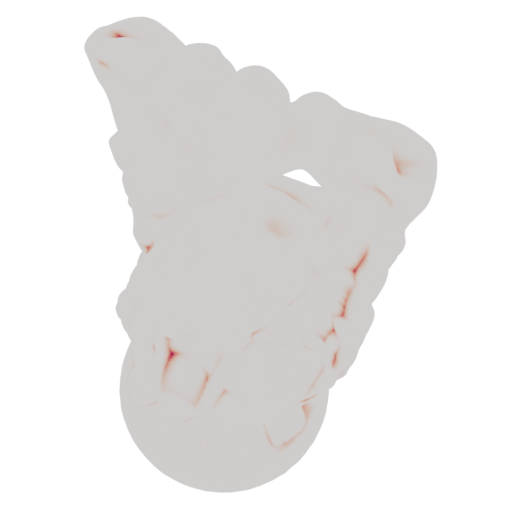}};
        \node[anchor=south west, draw=green, inner sep=1pt, line width=1pt] (img2et) at (170pt,-100pt) 
        {\oursynablaimage{trim=176 288 144 32,clip,width=47pt}{buddha_w_cd.png}};

        \node[anchor=south west] (img3) at (240pt,0pt) {\oursynablaimage{trim=20 20 20 20,clip,width=100pt}{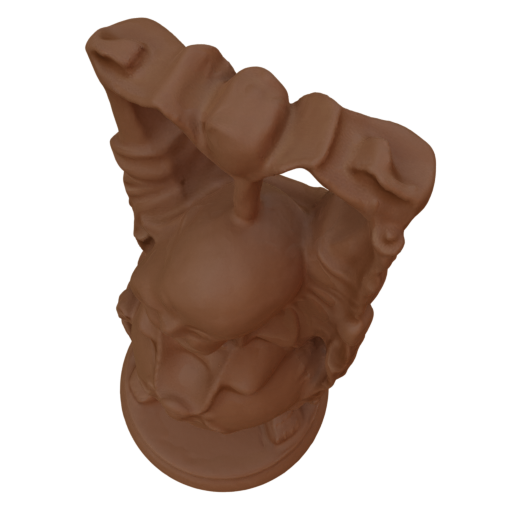}};
        \draw[green, line width=1pt] (277pt,60pt) rectangle +(40pt,40pt);
        \draw[cyan, line width=1pt] (270pt,19pt) rectangle +(20pt,20pt);
        \node[anchor=south west, draw=cyan, inner sep=1pt, line width=1pt] (img2c) at (240pt,-50pt) 
        {\oursynablaimage{trim=144 96 272 320,clip,width=47pt}{buddha_gt.png}};
        \node[anchor=south west, draw=green, inner sep=1pt, line width=1pt] (img2rt) at (290pt,-50pt) 
        {\oursynablaimage{trim=176 288 144 32,clip,width=47pt}{buddha_gt.png}};
        
        \node[below] at (img1.south) [yshift=-100pt]{\footnotesize w/o shape refinement};
        \node[below] at (img1.south) [yshift=-110pt]{\footnotesize(34.78 / 1.98e-5)};
        
        \node[below] at (img2.south) [yshift=-100pt]{\footnotesize w/ shape refinement};
        \node[below] at (img2.south) [yshift=-110pt]{\footnotesize (36.13 / 1.91e-5)};

        \node[below] at (img3.south) [yshift=-100pt]{\footnotesize GT};

    \end{tikzpicture}
    }
    \caption{
        {\bf Usefulness of our shape refinement in the physics-based inverse rendering (PBIR) stage.} To showcase the effectiveness of our shape refinement, we employ the \textit{buddha} scene and present zoom-in renderings along with Chamfer distance visualizations where darker colors indicate higher errors. Additionally, we report the PSNR for relighting and the Chamfer distance, presented at the bottom of our results.
    }
    \label{fig:our_vis_abla}
\end{figure*}